# Decision-Theoretic Planning: Structural Assumptions and Computational Leverage

**Craig Boutilier** CEBLY@CS.UBC.CA
*Department of Computer Science, University of British Columbia*
*Vancouver, BC, V6T 1Z4, Canada*

**Thomas Dean** TLD@CS.BROWN.EDU
*Department of Computer Science, Brown University*
*Box 1910, Providence, RI, 02912, USA*

**Steve Hanks** HANKS@CS.WASHINGTON.EDU
*Department of Computer Science and Engineering, University of Washington*
*Seattle, WA, 98195, USA*

## Abstract

Planning under uncertainty is a central problem in the study of automated sequential decision making, and has been addressed by researchers in many different fields, including AI planning, decision analysis, operations research, control theory and economics. While the assumptions and perspectives adopted in these areas often differ in substantial ways, many planning problems of interest to researchers in these fields can be modeled as *Markov decision processes* (MDPs) and analyzed using the techniques of decision theory.

This paper presents an overview and synthesis of MDP-related methods, showing how they provide a unifying framework for modeling many classes of planning problems studied in AI. It also describes structural properties of MDPs that, when exhibited by particular classes of problems, can be exploited in the construction of optimal or approximately optimal policies or plans. Planning problems commonly possess structure in the reward and value functions used to describe performance criteria, in the functions used to describe state transitions and observations, and in the relationships among features used to describe states, actions, rewards, and observations.

Specialized representations, and algorithms employing these representations, can achieve computational leverage by exploiting these various forms of structure. Certain AI techniques—in particular those based on the use of structured, intensional representations—can be viewed in this way. This paper surveys several types of representations for both classical and decision-theoretic planning problems, and planning algorithms that exploit these representations in a number of different ways to ease the computational burden of constructing policies or plans. It focuses primarily on abstraction, aggregation and decomposition techniques based on AI-style representations.

## 1. Introduction

Planning using decision-theoretic notions to represent domain uncertainty and plan quality has recently drawn considerable attention in artificial intelligence (AI).[1] *Decision-theoretic planning* (DTP) is an attractive extension of the classical AI planning paradigm because it allows one to model problems in which actions have uncertain effects, the decision maker has

1. See, for example, the recent texts (Dean, Allen, & Aloimonos, 1995; Dean & Wellman, 1991; Russell & Norvig, 1995) and the research reported in (Hanks, Russell, & Wellman, 1994).

incomplete information about the world, where factors such as resource consumption lead to solutions of varying quality, and where there may not be an absolute or well-defined "goal" state. Roughly, the aim of DTP is to form courses of action (plans or policies) that have high expected utility rather than plans that are guaranteed to achieve certain goals. When AI planning is viewed as a particular approach to solving sequential decision problems of this type, the connections between DTP and models used in other fields of research—such as decision analysis, economics and operations research (OR)—become more apparent. At a conceptual level, most sequential decision problems can be viewed as instances of *Markov decision processes* (MDPs), and we will use the MDP framework to make the connections explicit.

Much recent research on DTP has explicitly adopted the MDP framework as an underlying model (Barto, Bradtke, & Singh, 1995; Boutilier & Dearden, 1994; Boutilier, Dearden, & Goldszmidt, 1995; Dean, Kaelbling, Kirman, & Nicholson, 1993; Koenig, 1991; Simmons & Koenig, 1995; Tash & Russell, 1994), allowing the adaptation of existing results and algorithms for solving MDPs (e.g., from the field of OR) to be applied to planning problems. In doing so, however, this work has departed from the traditional definition of the "planning problem" in the AI planning community—one goal of this paper is to make explicit the connection between these two lines of work.

Adopting the MDP framework as a model for posing and solving planning problems has illuminated a number of interesting connections among techniques for solving decision problems, drawing on work from AI planning, reasoning under uncertainty, decision analysis and OR. One of the most interesting insights to emerge from this body of work is that many DTP problems exhibit considerable structure, and thus can be solved using special-purpose methods that recognize and exploit that structure. In particular, the use of feature-based representations to describe problems, as is the typical practice in AI, often highlights the problem's special structure and allows it to be exploited computationally with little effort.

There are two general impediments to the more widespread acceptance of MDPs within AI as a general model of planning. The first is the absence of explanations of the MDP model that make the connections to current planning research explicit, at either the conceptual or computational level. This may be due in large part to the fact that MDPs have been developed and studied primarily in OR, where the dominant concerns are, naturally, rather different. One aim of this paper is to make the connections clear: we provide a brief description of MDPs as a conceptual model for planning that emphasizes the connection to AI planning, and explore the relationship between MDP solution algorithms and AI planning algorithms. In particular, we emphasize that most AI planning models can be viewed as special cases of MDPs, and that classical planning algorithms have been designed to exploit the problem characteristics associated with these cases.

The second impediment is skepticism among AI researchers regarding the computational adequacy of MDPs as a planning model: can the techniques scale to solve planning problems of reasonable size? One difficulty with solution techniques for MDPs is the tendency to rely on explicit, state-based problem formulations. This can be problematic in AI planning since state spaces grow exponentially with the number of problem features. State space size and dimensionality are of somewhat lesser concern in OR and decision analysis. In these fields, an operations researcher or decision analyst will often hand-craft a model that ignores certain problem features deemed irrelevant, or will define other features that summarize a wide class of problem states. In AI, the emphasis is on the automatic solution of problems posed by users who lack the expertise of a decision analyst. Thus, assuming a well-crafted, compact state space is often not appropriate.

In this paper we show how specialized representations and algorithms from AI planning and problem solving can be used to design efficient MDP solution techniques. In particular, AI planning methods assume a certain *structure* in the state space, in the actions (or operators), and in the specification of a goal or other success criteria. Representations and algorithms have been designed that make the problem structure explicit and exploit that structure to solve problems effectively. We demonstrate how this same process of identifying structure, making it explicit, and exploiting it algorithmically can be brought to bear in the solution of MDPs.

This paper has several objectives. First, it provides an overview of DTP and MDPs suitable for readers familiar with traditional AI planning methods and makes connections with this work. Second, it describes the types of structure that can be exploited and how AI representations and methods facilitate computationally effective planning with MDPs. As such, it is a suitable introduction to AI methods for those familiar with the classical presentation of MDPs. Finally, it surveys recent work on the use of MDPs in AI and suggests directions for further research in this regard, and should therefore be of interest to researchers in DTP.

### 1.1 General Problem Definition

Roughly speaking, the class of problems we consider are those involving systems whose dynamics can be modeled as *stochastic processes*, where the actions of decision maker, referred to here as the *agent*, can influence the system's behavior. The system's current state and the choice of action jointly determine a probability distribution over the system's possible next states. The agent prefers to be in certain system states (e.g., goal states) over others, and therefore must determine a course of action—also called a "plan" or "policy" in this paper—that is likely to lead to these target states, possibly avoiding undesirable states along the way. The agent may not know the system's state exactly in making its decision on how to act, however—it may have to rely on incomplete and noisy sensors and be forced to base its choice of action on a probabilistic estimate of the state.

To help illustrate the types of problems in which we are interested, consider the following example. Imagine that we have a robot agent designed to help someone (the "user") in an office environment (see Figure 1). There are three activities it might undertake: picking up the user's mail, getting coffee, or tidying up the user's research lab. The robot can move from location to location and perform various actions that tend to achieve certain target states (e.g., bringing coffee to the user on demand, or maintaining a minimal level of tidiness in the lab).

We might associate a certain level of uncertainty with the effects of the robot's actions (e.g., when it tries to move to an adjacent location it might succeed 90% of the time and fail to move at all the other 10% of the time). The robot might have incomplete access to the true state of the system in that its sensors might supply it with incomplete information (it cannot tell whether mail is available for pickup if it is not in the mail room) and incorrect information (even when in the mail room its sensors occasionally fail to detect the presence of mail).

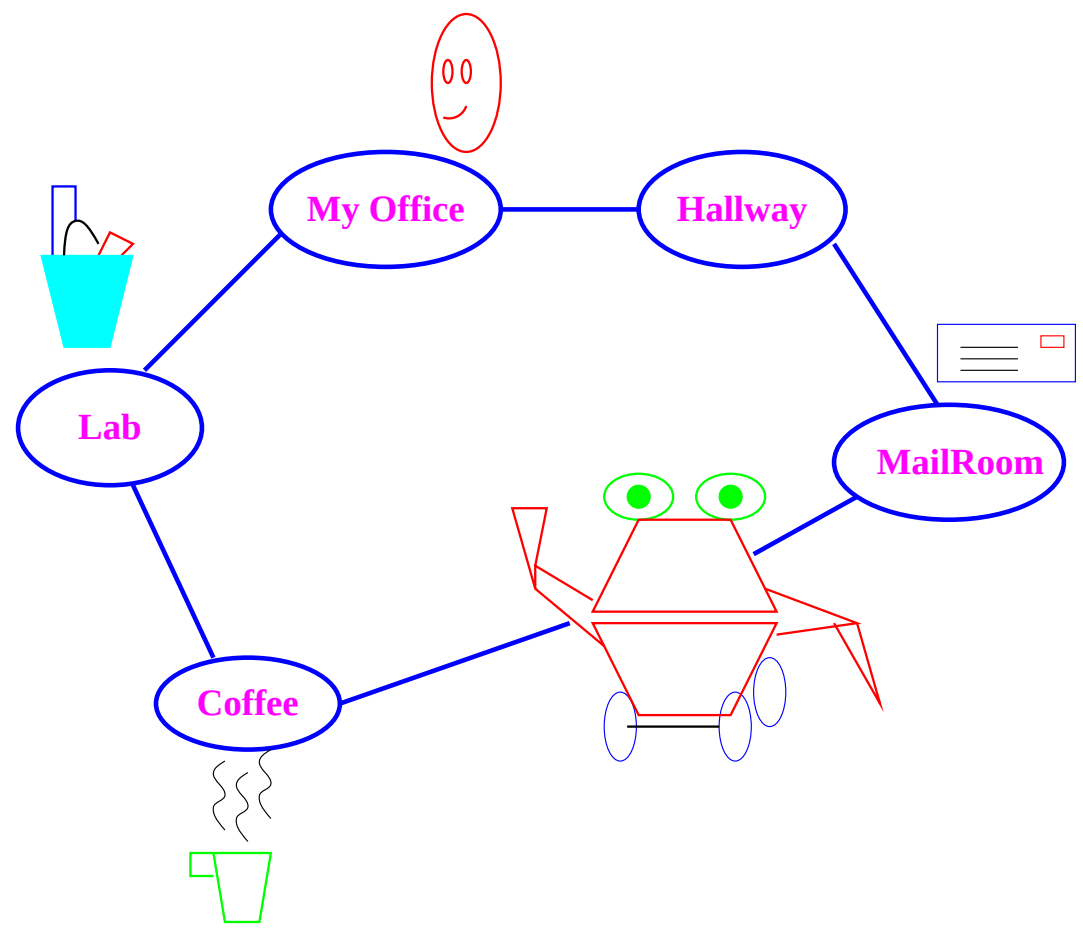

Figure 1: A decision-theoretic planning problem

Finally, the performance of the robot might be measured in various ways: do its actions guarantee that a goal will be achieved? Do they maximize some objective function defined over possible effects of its actions? Do they achieve a goal state with sufficient probability while avoiding "disastrous" states with near certainty? The stipulation of optimal or acceptable behavior is an important part of the problem specification.

The types of problems that can be captured using this general framework include classical (goal-oriented, deterministic, complete knowledge) planning problems and extensions such as conditional and probabilistic planning problems, as well as other more general problem formulations.

The discussion to this point has assumed an *extensional* representation of the system's states—one in which each state is explicitly named. In AI research, *intensional* representations are more common. An intensional representation is one in which states or sets of states are described using sets of multi-valued *features*. The choice of an appropriate set of features is an important part of the problem design. These features might include the current location of the robot, the presence or absence of mail, and so on. The performance metric is also typically expressed intensionally. Figure 2 serves as a reference for our example problem, which we use throughout the paper. It lists the basic features used to describe the states of the system, the actions available to the robot and the exogenous events that might occur, together with an intuitive description of the features, actions and events.

The remainder of the paper is organized as follows. In Section 2, we present the MDP framework in the abstract, introducing basic concepts and terminology and noting the relationship between this abstract model and the classical AI planning problem. Section 3 surveys common solution techniques—algorithms based on dynamic programming for general MDP problems and search algorithms for planning problems—and points out the relationship between problem assumptions and solution techniques. Section 4 turns from algorithms to representations, showing various ways in which the structured representations commonly used by AI algorithms can be used to represent MDPs compactly as well. Section 5 surveys some recent work on abstraction, aggregation and problem decomposition methods, and shows the connection to more traditional AI methods such as goal regression. This last section demonstrates that representational and computational methods from AI planning can be used in the solution of general MDPs. Section 5 also points out additional ways in which this type of computational leverage might be developed in the future.

| Features | Denoted | Description |
|---|---|---|
| Location | $Loc(M)$, etc. | Location of robot. Five possible locations: mailroom (M), coffee room (C), user's office (O), hallway (H), laboratory (L) |
| Tidiness | $T(0)$, etc. | Degree of lab tidiness. Five possible values: from 0 (messiest) to 4 (tidiest) |
| Mail present | $M, \overline{M}$ | Is there mail is user's mail box? True ($M$) or False ($\overline{M}$) |
| Robot has mail | $RHM, \overline{RHM}$ | Does the robot have mail in its possession? |
| Coffee request | $CR, \overline{CR}$ | Is there an outstanding (unfulfilled) request for coffee by the user? |
| Robot has coffee | $RHC, \overline{RHC}$ | Does the robot have coffee in its possession? |

| Actions | Denoted | Description |
|---|---|---|
| Move clockwise | *Clk* | Move to adjacent location (clockwise direction) |
| Counterclockwise | *CClk* | Move to adjacent location (counterclockwise direction) |
| Tidy lab | *Tidy* | If the robot is in the lab, the degree of tidiness is increased by 1 |
| Pickup mail | *PUM* | If the robot is in the mailroom and there is mail present, the robot takes the mail (*RHM* becomes true and $M$ becomes false) |
| Get coffee | *GetC* | If the robot is in the coffee room, it gets coffee (*RHC* becomes true) |
| Deliver mail | *DelM* | If the robot is in the office and has mail, it hands the mail to the user (*RHM* becomes false) |
| Deliver coffee | *DelC* | If the robot is in the office and has coffee, it hands the coffee to the user (*RHC* and *CR* both become false) |

| Events | Denoted | Description |
|---|---|---|
| Mail arrival | *ArrM* | Mail arrives causing $M$ to become true |
| Request coffee | *ReqC* | User issues coffee request causing *CR* to become true |
| Untidy the lab | *Mess* | The lab becomes messier (one degree less tidy) |

Figure 2: Elements of the robot domain.

## 2. Markov Decision Processes: Basic Problem Formulation

In this section we introduce the MDP framework and make explicit the relationship between this model and classical AI planning models. We are interested in controlling a *stochastic dynamical system*: a system that at any point in time can be in one of a number of distinct *states*, and in which the system's state changes over time in response to *events*. An *action* is a particular kind of event instigated by an agent in order to change the system's state. We assume that the agent has control over what actions are taken and when, though the effects of taking an action might not be perfectly predictable. In contrast, *exogenous events* are not under the agent's control, and their occurrence may be only partially predictable. This abstract view of an agent is consistent both with the "AI" view where the agent is an autonomous decision maker and the "control" view where a policy is determined ahead of time, programmed into a device, and executed without further deliberation.

### 2.1 States and State Transitions

We define a *state* to be a description of the system at a particular point in time. How one defines states can vary with particular applications, some notions being more natural than others. However, it is common to assume that the state captures all information relevant to the agent's decision-making process. We assume a finite *state space* $\mathcal{S} = \{s_1, \ldots, s_N\}$ of possible system states.[2] In most cases the agent will not have complete information about the current state; this uncertainty or incomplete information can be captured using a probability distribution over the states in $\mathcal{S}$.

A discrete-time stochastic dynamical system consists of a state space and probability distributions governing possible *state transitions*—how the *next state* of the system depends on past states. These distributions constitute a model of how the system evolves over time in response to actions and exogenous events, reflecting the fact that the effects of actions and events may not be perfectly predictable even if the prevailing state is known.

Although we are generally concerned with how the agent *chooses* an appropriate course of action, for the remainder of this section we assume that the agent's course of action is fixed, concentrating on the problem of predicting the system's state after the occurrence of a predetermined sequence of actions. We discuss the action selection problem in the next section.

We assume the system evolves in *stages*, where the occurrence of an event marks the transition from one stage $t$ to the next stage $t + 1$. Since events define changes in stage, and since events often (but not necessarily) cause state transitions, we often equate stage transitions with state transitions. Of course, it is possible for an event to occur but leave the system in the same state.

The system's progression through stages is roughly analogous to the passage of time. The two are identical if we assume that *some* action (possibly a no-op) is taken at each stage, and that every action takes unit time to complete. We can thus speak loosely as if stages correspond to units of time, and we refer to $\mathcal{T}$ interchangeably as the set of all stages and the set of all time points.[3]

We can model uncertainty by regarding the system's state at some stage $t$ as a random variable $S^t$ that takes values from $\mathcal{S}$. An assumption of "forward causality" requires that the variable $S^t$ does not depend *directly* on the value of future variable $S^k$ $(k > t)$. Roughly, it requires that we *model* our system such that the past history "directly" determines the distribution over current states, whereas knowledge of future states can influence the estimate of the current state only indirectly by providing evidence on what the current state may have been so as to lead to these future states. Figure 3(a) shows a graphical perspective on a discrete-time, stochastic dynamical system. The nodes are random variables denoting the state at a particular time, and the arcs indicate the direct probabilistic dependence of states on previous states. To describe this system completely we must also supply the conditional distributions $\Pr(S^t | S^0, S^1, \cdots S^{t-1})$ for all times $t$.

States should be thought of as descriptions of the system being modeled, so the question arises of how much detail about the system is captured in a state description. More

2. Most of the discussion in this paper also applies to cases where the state space is countably infinite. See (Puterman, 1994) for a discussion of infinite and continuous-state problems.
3. While we do not deal with such topics here, there is a considerable literature in the OR community on continuous-time Markov decision processes (Puterman, 1994).

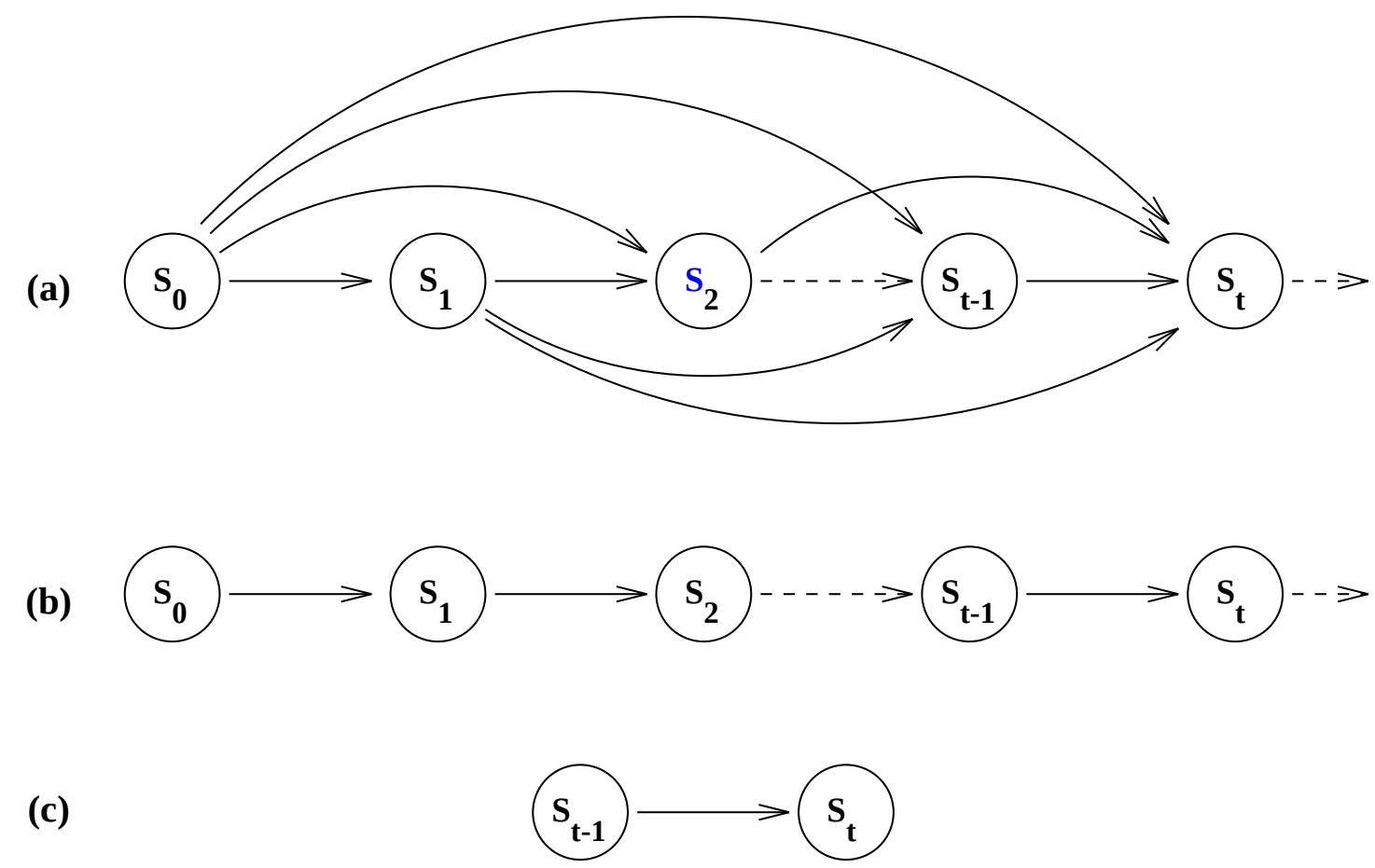

Figure 3: A general stochastic process (a), a Markov chain (b), and a stationary Markov chain (c).

detail implies more information about the system, which in turn often translates into better predictions of future behavior. Of course, more detail also implies a larger set $\mathcal{S}$, which can increase the computational cost of decision making.

It is commonly assumed that a state contains enough information to predict the next state. In other words, any information about the *history* of the system relevant to predicting its future is captured explicitly in the state itself. Formally, this assumption, the *Markov assumption*, says that knowledge of the present state renders information about the past irrelevant to making predictions about the future:

$$\Pr(S^{t+1}|S^t, S^{t-1}, \ldots, S^0) = \Pr(S^{t+1}|S^t)$$

Markovian models can be represented graphically using a structure like that in Figure 3(b), reflecting the fact that the present state is sufficient to predict future state evolution.[4]

Finally, it is common to assume that the effects of an event depend only on the prevailing state, and not the *stage* or time at which the event occurs.[5] If the distribution predicting the next state is the same regardless of stage, the model is said to be *stationary* and can be represented schematically using just two stages, as in Figure 3(c). In this case only a single conditional distribution is required. In this paper we generally restrict our attention to discrete-time, finite-state, stochastic dynamical systems with the Markov property, commonly called *Markov chains*. Furthermore, most of our discussion is restricted to stationary chains.

To complete the model we must provide a probability distribution over initial states, reflecting the probability of being in any state at stage 0. This distribution can be repre-

4. It is worth mentioning that the Markov property applies to the particular *model* and not to the system itself. Indeed, any non-Markovian model of a system (of finite order, i.e., whose dynamics depend on at most the $k$ previous states for some $k$) can be converted to an equivalent though larger Markov model. In control theory, this is called conversion to *state form* (Luenberger, 1979).

5. Of course, this is also a statement about model detail, saying that the state carries enough information to make the stage irrelevant to predicting transitions.

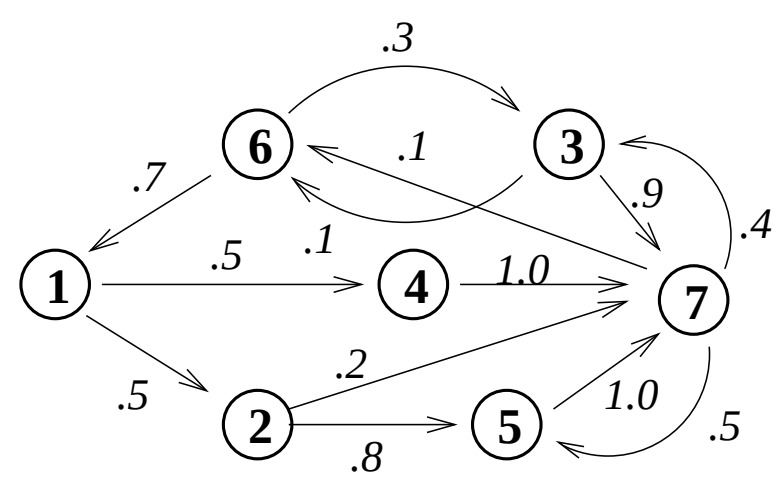

Figure 4: A state-transition diagram.

sented as a real-valued (row) vector of size $N = |S|$ (one entry for each state). We denote this vector $P^0$ and use $p_i^0$ to denote its $i$th entry, that is, the probability of starting in state $s_i$.

We can represent a $T$-stage nonstationary Markov chain with $T$ *transition matrices*, each of size $N \times N$, where matrix $P^t$ captures the transition probabilities governing the system as it moves from stage $t$ to stage $t+1$. Each matrix consists of probabilities $p_{ij}^t$, where $p_{ij}^t = \Pr(S^{t+1} = s_j | S^t = s_i)$. If the process is stationary, the transition matrix is the same at all stages and one matrix (whose entries are denoted $p_{ij}$) will suffice. Given an initial distribution over states $P^0$, the probability distribution over states after $n$ stages is $\prod_{i=n}^{0} P^i$.

A stationary Markov process can also be represented using a *state-transition diagram* as in Figure 4. Here nodes correspond to particular states and the stage is not represented explicitly. Arcs denote possible transitions (those with non-zero probability) and are labeled with the transition probabilities $p_{ij} = \Pr(S^{t+1} = s_j | S^t = s_i)$. The arc from node $i$ to node $j$ is labeled with $p_{ij}$ if $p_{ij} > 0$.[6] The size of such a diagram is at least $O(N)$ and at most $O(N^2)$, depending on the number of arcs. This is a useful representation when the transition graph is relatively sparse, for example, when most states have immediate transitions to only few neighbors.

**Example 2.1** To illustrate these notions, imagine that the robot in Figure 1 is executing the policy of moving counterclockwise repeatedly. We restrict our attention to two variables, location *Loc* and presence of mail $M$, giving a state space of size 10. We suppose that the robot always moves to the adjacent location with probability 1.0. In addition, mail can arrive at the mailroom with probability 0.2 at any time (independent of the robot's location), causing the variable $M$ to become true. Once $M$ becomes true, the robot cannot move to a state where $M$ is false, since the action of moving does not influence the presence of mail. The state-transition diagram for this example is illustrated in Figure 5. The transition matrix is also shown. □

The structure of a Markov chain is occasionally of interest to us in planning. A subset $C \subseteq \mathcal{S}$ is *closed* if $p_{ij} = 0$ for all $i \in C$ and $j \notin C$. It is a *proper closed* set if no proper subset of $C$ enjoys this property. We sometimes refer to proper closed sets as *recurrent classes* of states. If a closed set consists of a single state, then that state is called an *absorbing* state. Once an agent enters a closed set or absorbing state, it remains there

6. It is important to note that the nodes here do not represent random variables as in the earlier figures.

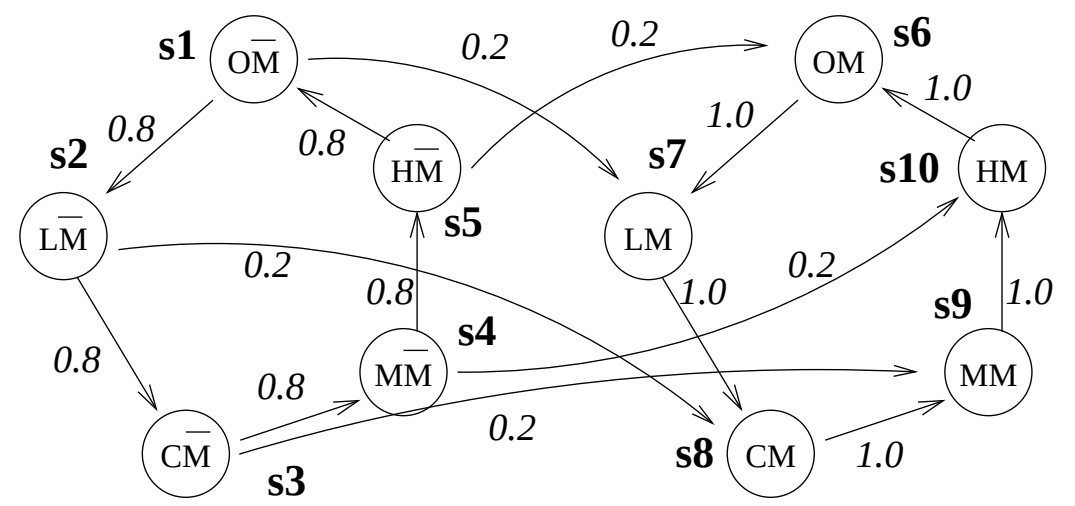

| | $s_1$ | $s_2$ | $s_3$ | $s_4$ | $s_5$ | $s_6$ | $s_7$ | $s_8$ | $s_9$ | $s_{10}$ |
|---|---|---|---|---|---|---|---|---|---|---|
| $s_1$ | 0.0 | 0.8 | 0.0 | 0.0 | 0.0 | 0.0 | 0.2 | 0.0 | 0.0 | 0.0 |
| $s_2$ | 0.0 | 0.0 | 0.8 | 0.0 | 0.0 | 0.0 | 0.0 | 0.2 | 0.0 | 0.0 |
| $s_3$ | 0.0 | 0.0 | 0.0 | 0.8 | 0.0 | 0.0 | 0.0 | 0.0 | 0.2 | 0.0 |
| $s_4$ | 0.0 | 0.0 | 0.0 | 0.0 | 0.8 | 0.0 | 0.0 | 0.0 | 0.0 | 0.2 |
| $s_5$ | 0.8 | 0.0 | 0.0 | 0.0 | 0.0 | 0.2 | 0.0 | 0.0 | 0.0 | 0.0 |
| $s_6$ | 0.0 | 0.0 | 0.0 | 0.0 | 0.0 | 0.0 | 1.0 | 0.0 | 0.0 | 0.0 |
| $s_7$ | 0.0 | 0.0 | 0.0 | 0.0 | 0.0 | 0.0 | 0.0 | 1.0 | 0.0 | 0.0 |
| $s_8$ | 0.0 | 0.0 | 0.0 | 0.0 | 0.0 | 0.0 | 0.0 | 0.0 | 1.0 | 0.0 |
| $s_9$ | 0.0 | 0.0 | 0.0 | 0.0 | 0.0 | 0.0 | 0.0 | 0.0 | 0.0 | 1.0 |
| $s_{10}$ | 0.0 | 0.0 | 0.0 | 0.0 | 0.0 | 1.0 | 0.0 | 0.0 | 0.0 | 0.0 |

Figure 5: The state-transition diagram and transition matrix for a moving robot.

forever with probability 1. In the example above (Figure 5), the set of states where $M$ holds forms a recurrent class. There are no absorbing states in the example, but should we program the robot to stay put whenever it is in the state $\langle M, Loc(O)\rangle$, then this would be an absorbing state in the altered chain. Finally, we say a state is *transient* if it does not belong to a recurrent class. In Figure 5, each state where $\overline{M}$ holds is transient—eventually (with probability 1), the agent leaves the state and never returns, since there is no way to remove mail once it arrives.

### 2.2 Actions

Markov chains can be used to describe the evolution of a stochastic system, but they do not capture the fact that an agent can *choose* to perform actions that alter the state of the system. A key element of MDPs is the set of actions available to the decision maker. When an action is performed in a particular state, the state changes stochastically in response to the action. We assume that the agent takes some action at each stage of the process, and then the system changes state accordingly.

At each stage $t$ of the process and each state $s$, the agent has available a set of actions $\mathcal{A}^t_s$. This is called the *feasible set* for $s$ at stage $t$. To describe the effects of $a \in \mathcal{A}^t_s$, we must supply the state-transition distribution $\Pr(S^{t+1}|S^t = s, A^t = a)$ for all actions $a$, states $s$, and stages $t$. Unlike the case of a Markov chain, the terms $\Pr(S^{t+1}|S^t = s, A^t = a)$ are not true conditional distributions, but rather a family of distributions parameterized by $S^t$ and $A^t$, since the probability of $A^t$ is not part of the model. We retain this notation, however, for its suggestive nature.

We often assume that the feasible set of actions is the same for all stages and states, in which case the set of actions is $\mathcal{A} = \{a_1, \ldots, a_K\}$ and each can be executed at any time. This contrasts with the AI planning practice of assigning *preconditions* to actions defining the states in which they can meaningfully be executed. Our model takes the view that any action can be executed (or "attempted") in any state. If the action has no effect when executed in some state, or its execution leads to disastrous effects, this can be noted in the action's transition matrix. Action preconditions are often a computational convenience rather than a representational necessity: they can make the planning process more efficient by identifying states in which the planner should not even consider selecting that action. Preconditions can be represented in MDPs by relaxing the assumption that the set of feasible actions is the same for all states. To illustrate planning concepts below, however, we sometimes assume actions do have preconditions.

| | $s_1$ | $s_2$ | $s_3$ | $s_4$ | $s_5$ | $s_6$ | $s_7$ | $s_8$ | $s_9$ | $s_{10}$ |
|---|---|---|---|---|---|---|---|---|---|---|
| $s_1$ | 0.0 | 0.8 | 0.0 | 0.0 | 0.0 | 0.0 | 0.2 | 0.0 | 0.0 | 0.2 |
| $s_2$ | 0.0 | 0.0 | 0.8 | 0.0 | 0.0 | 0.0 | 0.0 | 0.2 | 0.0 | 0.0 |
| $s_3$ | 0.0 | 0.8 | 0.0 | 0.0 | 0.0 | 0.0 | 0.2 | 0.0 | 0.0 | 0.0 |
| $s_4$ | 0.0 | 0.0 | 0.8 | 0.0 | 0.0 | 0.0 | 0.0 | 0.2 | 0.0 | 0.0 |
| $s_5$ | 0.8 | 0.0 | 0.0 | 0.0 | 0.0 | 0.2 | 0.0 | 0.0 | 0.0 | 0.0 |
| $s_6$ | 0.0 | 0.0 | 0.0 | 0.0 | 0.0 | 0.0 | 1.0 | 0.0 | 0.0 | 0.0 |
| $s_7$ | 0.0 | 0.0 | 0.0 | 0.0 | 0.0 | 0.0 | 0.0 | 1.0 | 0.0 | 0.0 |
| $s_8$ | 0.0 | 0.0 | 0.0 | 0.0 | 0.0 | 0.0 | 1.0 | 0.0 | 0.0 | 0.0 |
| $s_9$ | 0.0 | 0.0 | 0.0 | 0.0 | 0.0 | 0.0 | 0.0 | 1.0 | 0.0 | 0.0 |
| $s_{10}$ | 0.0 | 0.0 | 0.0 | 0.0 | 0.0 | 1.0 | 0.0 | 0.0 | 0.0 | 0.0 |

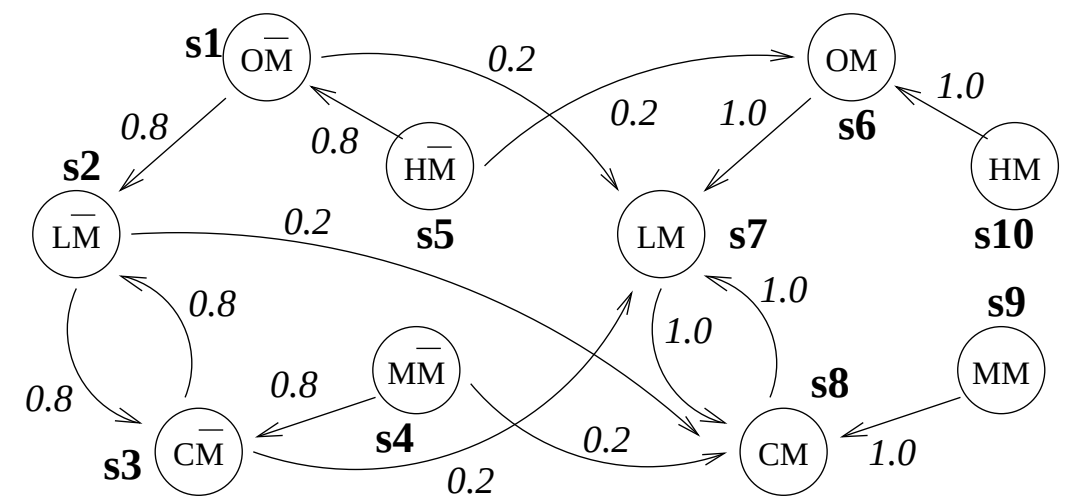

Figure 6: The transition matrix for *Clk* and the induced transition diagram for a two-action policy.

We again restrict our attention to stationary processes, which in this case means that the effects of each action depends only on the state and not on the stage. Our transition matrices thus take the form $p^k_{ij} = \Pr(S^{t+1} = s_j | S^t = s_i, A^t = a_k)$, capturing the probability that the system moves to state $s_j$ when $a_k$ is executed in state $s_i$. In stationary models an action is fully described by a *single* $N \times N$ transition matrix $P^k$. It is important to note that the transition matrix for an action includes not only the direct effects of executing the action but also the effects of any exogenous events that might occur at the same stage.[7]

**Example 2.2** The example in Figure 5 can be extended so the agent has two available actions: moving clockwise and moving counterclockwise. The transition matrix for *CClk* (with the assumption that mail arrives with probability 0.2) is shown in Figure 5. The matrix for *Clk* appears on the left in Figure 6. Suppose the agent fixes its behavior so that it moves clockwise in locations $M$ and $C$ and counterclockwise in locations $H$, $O$ and $L$ (we address below how the agent might come to know its location so that it can actually implement this behavior). This defines the Markov chain illustrated in the transition diagram on the right in Figure 6. □

## 2.3 Exogenous Events

*Exogenous events* are those events that stochastically cause state transitions, much like actions, but beyond the control of the decision maker. These might correspond to the evolution of a natural process or the action of another agent. Notice that the effect of the action *CClk* in Figure 5 "combines" the effects of the robot's action with that of the exogenous event of mail arrival: state-transition probabilities incorporate both the motion of the robot (causing a change in location) and the possible change in mail status due to mail arrival. For the purposes of decision making, it is precisely this combined effect

7. It is possible to assess the effects of actions and exogenous events separately, then combine them into a single transition matrix in certain cases (Boutilier & Puterman, 1995). We discuss this later in this section.

that is important when predicting the distribution over possible states resulting when an action is taken. We call such models of actions *implicit-event models*, since the effects of the exogenous event are folded into the transition probabilities associated with the action. However, it is often natural to view these transitions as comprised of these two separate events, each having its own effect on the state. More generally, we often think of transitions as determined by the effects of the agent's chosen action and those of certain *exogenous events* beyond the agent's control, each of which may occur with a certain probability. When the effects of actions are decomposed in this fashion, we call the action model an *explicit-event model.*

Specifying a transition function for an action and zero or more exogenous events is not generally easy, for actions and events can interact in complex ways. For instance, consider specifying the effect of action $PUM$ (pickup mail) at a state where no mail is present but there is the possibility of "simultaneous" mail arrival (i.e., during the "same unit" of discrete time). If the event $ArrM$ occurs, does the robot obtain the newly arrived mail, or does the mail remain in the mailbox? Intuitively, this depends on whether the mail arrived before or after the pickup was completed (albeit within the same time quantum). The state transition in this case can be viewed as the composition of two transitions where the precise description of the composition depends on the ordering of the agent's action and the exogenous event. If mail arrives first, the transition might be $s \rightarrow s' \rightarrow s''$, where $s'$ is a state where mail is waiting and $s''$ is a state where no mail is waiting and the robot is holding mail; but if the pickup action is completed first, the transition would be $s \rightarrow s \rightarrow s'$ (i.e., $PUM$ has no effect, then mail arrives and remains in the box).

The picture is more complicated if the actions and events can truly occur simultaneously over some interval—in this case the resulting transition need not be a composition of the individual transitions. As an example, if the robot lifts the side of a table on which a glass of water is situated, the water will spill; similarly if an exogenous event causes the other side to be raised. But if the action and event occur simultaneously, the result is qualitatively different (the water is not spilled). Thus, the "interleaving" semantics described above is not always appropriate.

Because of such complications, modeling exogenous events and their combination with actions or other events can be approached in many ways, depending on the modeling assumptions one is willing to make. Generally, we specify three types of information. First, we provide transition probabilities for all actions and events under the assumption that these *occur in isolation*—these are standard transition matrices. The transition matrix in Figure 5 can be decomposed into the two matrices shown in Figure 7, one for *Clk* and one for $ArrM$.[8] Second, for each exogenous event, we must specify its *probability of occurrence.* Since this can vary with the state, we generally require a vector of length $N$ indicating the probability of occurrence at each state. The occurrence vector for $ArrM$ would be

$$[0.2\ 0.2\ 0.2\ 0.2\ 0.2\ 0.0\ 0.0\ 0.0\ 0.0\ 0.0]$$

8. The fact that these individual matrices are deterministic is an artifact of the example. In general, the actions and events will each be represented using genuinely stochastic matrices.

| | $s_1$ | $s_2$ | $s_3$ | $s_4$ | $s_5$ | $s_6$ | $s_7$ | $s_8$ | $s_9$ | $s_{10}$ |
|---|---|---|---|---|---|---|---|---|---|---|
| $s_1$ | 0.0 | 1.0 | 0.0 | 0.0 | 0.0 | 0.0 | 0.0 | 0.0 | 0.0 | 0.0 |
| $s_2$ | 0.0 | 0.0 | 1.0 | 0.0 | 0.0 | 0.0 | 0.0 | 0.0 | 0.0 | 0.0 |
| $s_3$ | 0.0 | 0.0 | 0.0 | 1.0 | 0.0 | 0.0 | 0.0 | 0.0 | 0.0 | 0.0 |
| $s_4$ | 0.0 | 0.0 | 0.0 | 0.0 | 1.0 | 0.0 | 0.0 | 0.0 | 0.0 | 0.0 |
| $s_5$ | 1.0 | 0.0 | 0.0 | 0.0 | 0.0 | 0.0 | 0.0 | 0.0 | 0.0 | 0.0 |
| $s_6$ | 0.0 | 0.0 | 0.0 | 0.0 | 0.0 | 0.0 | 1.0 | 0.0 | 0.0 | 0.0 |
| $s_7$ | 0.0 | 0.0 | 0.0 | 0.0 | 0.0 | 0.0 | 0.0 | 1.0 | 0.0 | 0.0 |
| $s_8$ | 0.0 | 0.0 | 0.0 | 0.0 | 0.0 | 0.0 | 0.0 | 0.0 | 1.0 | 0.0 |
| $s_9$ | 0.0 | 0.0 | 0.0 | 0.0 | 0.0 | 0.0 | 0.0 | 0.0 | 0.0 | 1.0 |
| $s_{10}$ | 0.0 | 0.0 | 0.0 | 0.0 | 0.0 | 1.0 | 0.0 | 0.0 | 0.0 | 0.0 |

Action *Clk*

| | $s_1$ | $s_2$ | $s_3$ | $s_4$ | $s_5$ | $s_6$ | $s_7$ | $s_8$ | $s_9$ | $s_{10}$ |
|---|---|---|---|---|---|---|---|---|---|---|
| $s_1$ | 0.0 | 0.0 | 0.0 | 0.0 | 0.0 | 1.0 | 0.0 | 0.0 | 0.0 | 0.0 |
| $s_2$ | 0.0 | 0.0 | 0.0 | 0.0 | 0.0 | 0.0 | 1.0 | 0.0 | 0.0 | 0.0 |
| $s_3$ | 0.0 | 0.0 | 0.0 | 0.0 | 0.0 | 0.0 | 0.0 | 1.0 | 0.0 | 0.0 |
| $s_4$ | 0.0 | 0.0 | 0.0 | 0.0 | 0.0 | 0.0 | 0.0 | 0.0 | 1.0 | 0.0 |
| $s_5$ | 0.0 | 0.0 | 0.0 | 0.0 | 0.0 | 0.0 | 0.0 | 0.0 | 0.0 | 1.0 |
| $s_6$ | 0.0 | 0.0 | 0.0 | 0.0 | 0.0 | 1.0 | 0.0 | 0.0 | 0.0 | 0.0 |
| $s_7$ | 0.0 | 0.0 | 0.0 | 0.0 | 0.0 | 0.0 | 1.0 | 0.0 | 0.0 | 0.0 |
| $s_8$ | 0.0 | 0.0 | 0.0 | 0.0 | 0.0 | 0.0 | 0.0 | 1.0 | 0.0 | 0.0 |
| $s_9$ | 0.0 | 0.0 | 0.0 | 0.0 | 0.0 | 0.0 | 0.0 | 0.0 | 1.0 | 0.0 |
| $s_{10}$ | 0.0 | 0.0 | 0.0 | 0.0 | 0.0 | 0.0 | 0.0 | 0.0 | 0.0 | 1.0 |

Event *ArrM*

Figure 7: The transition matrices for an action and exogenous event in an explicit-event model.

where we assume, for illustration, that mail arrives only when none is present.[9] The final requirement is a *combination function* that describes how to "compose" the transitions of an action with any subset of event transitions. As indicated above, this can be very complex, sometimes almost unrelated to the individual action and event transitions. However, under certain assumptions combination functions can be specified reasonably concisely.

One way of modeling the composition of transitions is to assume an interleaving semantics of the type alluded to above. In this case, one needs to specify the probability that the action and events that take place occur in a specific order. For instance, one might assume that each event occurs at a time—within the discrete time unit—according to some continuous distribution (e.g., an exponential distribution with a given rate). With this information, the probability of any particular ordering of transitions, given that certain events occur, can be computed, as can the resulting distribution over possible next states. In the example above, the probability of (composed) transitions $s_1 \rightarrow s_2 \rightarrow s_3$ and $s_1 \rightarrow s_1 \rightarrow s_2$ would be given by the probabilities with which mail arrived first or last, respectively.

In certain cases, the probability of this ordering is not needed. To illustrate another combination function, assume that the action always occurs before the exogenous events. Furthermore, assume that events are *commutative*: (a) for any initial state $s$ and any pair of events $e_1$ and $e_2$, the distribution that results from applying event sequence $e_1 \cdot e_2$ to $s$ is identical to that obtained from the sequence $e_2 \cdot e_1$; and (b) the occurrence probabilities at intermediate states are identical. Intuitively, the set of events in our domain, *ArrM*, *ReqC* and *Mess*, has this property. Under these conditions the combined transition distribution for any action $a$ is computed by considering the probability of any subset of events and applying that subset in any order to the distribution associated with $a$.

Generally, we can construct an implicit-event model from the various components of the explicit-event model; thus, the "natural" specification can be converted to the form usually used by MDP solution algorithms. Under the two assumptions above, for instance, we can form an implicit event transition matrix $\Pr(s_i, a, s_j)$ for any action $a$, given the matrix $\widehat{\Pr}_a(s_i, s_j)$ for $a$ (which assumes no event occurrences), the matrices $\Pr_e(s_i, s_j)$ for events $e$, and the occurrence vector $\Pr_e(s_i)$ for each event $e$. The *effective transition matrix* for

9. The probability of different events may be correlated (possibly at particular states). If this is the case, then it is necessary to specify occurrence probabilities for subsets of events. We will treat event occurrence probabilities as independent for ease of exposition.

event $e$ is defined as follows:

$$\widehat{\Pr}_e(s_i, s_j) = \Pr_e(s_i)\Pr_e(s_i, s_j) + \begin{cases} 1 - \Pr_e(s_i) & : \quad i = j \\ 0 & : \quad i \neq j \end{cases}$$

This equation captures the event transition probabilities with the probability of event occurrence factored in. If we let $E, E'$ denote the diagonal matrices with entries $E_{kk} = \Pr_e(s_k)$ and $E'_{kk} = 1 - \Pr_e(s_k)$, then $\widehat{\Pr}_e(s_i, s_j) = E \Pr_e + E'$. Under the assumptions above, the implicit-event matrix $\Pr(s_i, a, s_j)$ for action $a$ is then given by $\Pr = \widehat{\Pr}_{e_1} \cdots \widehat{\Pr}_{e_n} \Pr_a$ for any ordering of the $n$ possible events.

Naturally, different procedures for constructing implicit-event matrices will be required given different assumptions about action and event interaction. Whether such implicit models are constructed or specified directly without explicit mention of the exogenous events, we will always assume unless stated otherwise that action transition matrices take into account the effects of exogenous events as well, and thus represent the agent's best information about what will happen if it takes a particular action.

### 2.4 Observations

Although the *effects* of an action can depend on any aspect of the prevailing state, the *choice* of action can depend only on what the agent can *observe* about the current state and *remember* about its prior observations. We model the agent's observational or sensing capabilities by introducing a finite set of *observations* $\mathcal{O} = \{o_1, \ldots, o_H\}$. The agent receives an observation from this set at each stage prior to choosing its action at that stage. We can model this observation as a random variable $O^t$ whose value is taken from $\mathcal{O}$. The probability that a particular $O^t$ is generated can depend on:

- the state of the system at $t - 1$
- the action taken at $t - 1$
- the state of the system at $t$ after taking the action at $t - 1$ and after the effects of any exogenous events at $t - 1$ are realized, but before the action at $t$ is taken.

We let $\Pr(O^t = o_h | S^{t-1} = s_i, A^{t-1} = a_k, S^t = s_j)$ be the probability that the agent observes $o_h$ at stage $t$ given that it performs $a_k$ in state $s_i$ and ends up in state $s_j$. As with actions, we assume that observational distributions are stationary (independent of the stage), using $p^h_{i,j,k} = \Pr(o_h | s_i, a_k, s_j)$ to denote this quantity. We can view the probabilistic dependencies among state, action and observation variables as a graph in which the time-indexed variables are shown as nodes and one variable is directly probabilistically dependent on another if there is an edge from the latter to the former; see Figure 8.

This model allows a wide variety of assumptions about the agent's sensing capabilities. At one extreme are *fully observable MDPs* (FOMDPs), in which the agent knows exactly what state it is in at each stage $t$. We model this case by letting $\mathcal{O} = \mathcal{S}$ and setting

$$\Pr(o_h | s_i, a_k, s_j) = \begin{cases} 1 & \text{iff } o_h = s_j \\ 0 & \text{otherwise} \end{cases}$$

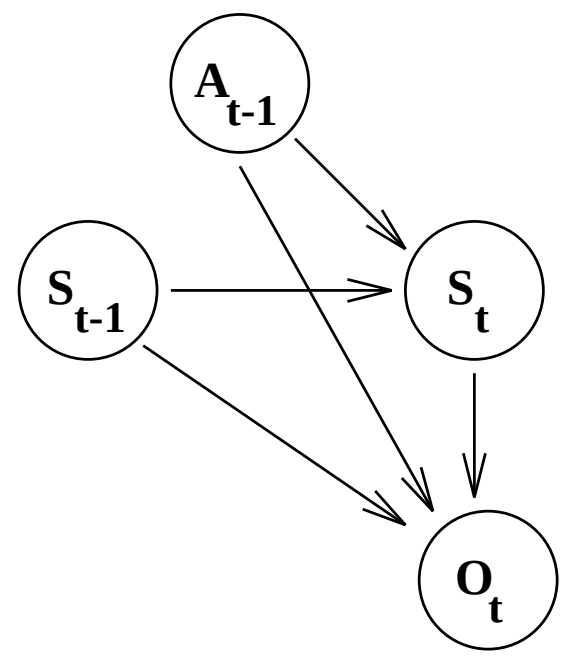

Figure 8: Graph showing the dependency relationships among states, actions and observations at different times.

In the example above, this means the robot always knows its exact location and whether or not mail is waiting in the mailbox, even if it is not in the mailroom when the mail arrives. The agent thus receives perfect feedback about the results of its actions and the effects of exogenous events—it has noisy effectors but complete, noise-free, and "instantaneous" sensors. Most recent AI research that adopts the MDP framework explicitly assumes full observability.

At the other extreme we might consider *non-observable* systems (NOMDPs) in which the agent receives *no* information about the system's state during execution. We can model this case by letting $\mathcal{O} = \{o\}$. Here the same observation is reported at each stage, revealing no information about the state, so that $\Pr(s_j|s_i, a_k, o) = \Pr(s_j|s_i, a_k)$. In these *open-loop systems*, the agent receives no useful feedback about the results of its actions: the agent has noisy effectors and *no* sensors. In this case an agent chooses its actions according to a *plan* consisting of a sequence of actions executed unconditionally. In effect, the agent is relying on its predictive model to determine good action choices before execution time.

Traditionally, AI planning work has implicitly made the assumption of non-observability, often coupled with an *omniscience assumption*—that the agent knows the initial state with certainty, can predict the effects of its actions perfectly, and can precisely predict the occurrence of any exogenous events and their effects. Under these circumstances, the agent can predict the exact outcome of any plan, thus obviating the need for observation. Such an agent can build a straight-line plan—a sequence of actions to be performed without feedback—that is as good as any plan whose execution might depend on information gathered at execution time.

These two extremes are special cases of the general observation model described above, which allows the agent to receive incomplete or noisy information about the system state (i.e., *partially observable* MDPs, or POMDPs). For example, the robot might be able to determine its location exactly, but might not be able to determine whether mail arrives unless it is in the mailroom. Furthermore, its "mail" sensors might occasionally report inaccurately, leading to an incorrect belief as to whether there is mail waiting.

**Example 2.3** Suppose the robot has a "checkmail" action that does not change the system state but generates an observation that is influenced by the presence of mail, provided the robot is in the mailroom at the time the action is performed. If the robot is not in the mailroom, the sensor always reports "no mail." A noisy "checkmail" sensor can be described by a probability distribution like the one shown in Figure 9. We can view these error probabilities as the probability of "false positives" (0.05) and "false negatives" (0.08). □

| | $\Pr(Obs = mail)$ | $\Pr(Obs = nomail)$ |
|---|---|---|
| $Loc(M), M$ | 0.92 | 0.08 |
| $Loc(M), \overline{M}$ | 0.05 | 0.95 |
| $\overline{Loc(M)}, M$ | 0.00 | 1.00 |
| $\overline{Loc(M)}, \overline{M}$ | 0.00 | 1.00 |

Figure 9: Observation probabilities for checking mailbox.

### 2.5 System Trajectories and Observable Histories

We use the terms *trajectory* and *history* interchangeably to describe the system's behavior during the course of a problem-solving episode, or perhaps some initial segment thereof. The *complete system history* is the sequence of states, actions, and observations generated from stage 0 to some time point of interest, and can be of finite or infinite length. Complete histories can be represented by a (possibly infinite) sequence of tuples of the form

$$\langle \langle S^0, O^0, A^0 \rangle, \langle S^1, O^1, A^1 \rangle, \ldots \langle S^T, O^T, A^T \rangle \rangle$$

We can define two alternative notions of history that contain less complete information. For some arbitrary stage $t$ we define the *observable history* as the sequence

$$\langle \langle O^0, A^0 \rangle, \ldots, \langle O^{t-1}, A^{t-1} \rangle \rangle$$

where $O^0$ is the observation of the initial state. The observable history at stage $t$ comprises all information available to the agent about its history when it chooses its action at stage $t$.

A third type of trajectory is the *system trajectory*, which is the sequence

$$\langle \langle S^0, A^0 \rangle, \ldots, \langle S^{t-1}, A^{t-1} \rangle, S^t \rangle$$

describing the system's behavior in "objective" terms, independent of the agent's particular view of the system.

In evaluating an agent's performance, we will generally be interested in the system trajectory. An agent's policy must be defined in terms of the observable history, since the agent does not have access to the system trajectory, except in the fully observable case, when the two are equivalent.

### 2.6 Reward and Value

The problem facing the decision maker is to select an action to be performed at each stage of the decision problem, making this decision on the basis of the observable history. The agent still needs some way to judge the *quality* of a course of action. This is done by defining a *value function* $V(\cdot)$ as a function mapping the set of system histories $\mathcal{H}_S$ into the reals; that is, $V : \mathcal{H}_S \to \mathbf{R}$.[10] The agent *prefers* system history $h$ to $h'$ just in case $V(h) > V(h')$. Thus, the agent judges its behavior to be good or bad depending on its effect on the underlying system trajectory. Generally, the agent cannot predict with certainty which system trajectory will occur, and can at best generate a probability distribution over the possible trajectories caused by its actions. In that case, it computes the *expected value* of each candidate course of action and chooses a policy that maximizes that quantity.

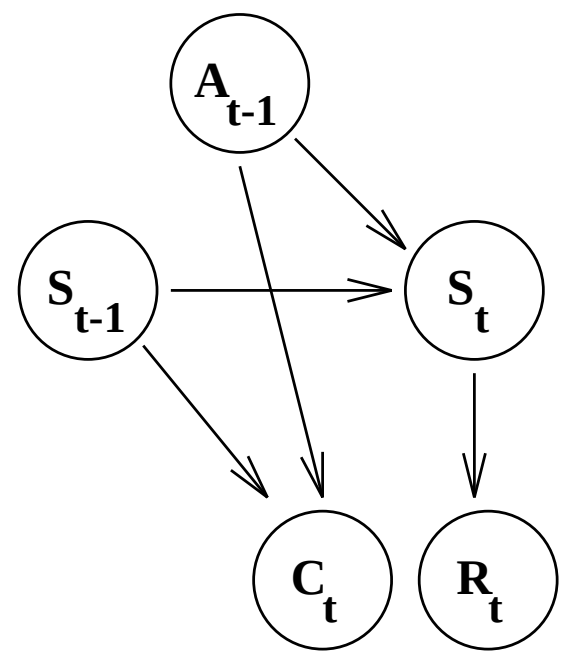

Figure 10: Decision process with rewards and action costs.

Just as with system dynamics, specifying a value function over arbitrary trajectories can be cumbersome and unintuitive. It is therefore important to identify structure in the value function that can lead to a more parsimonious representation.

Two assumptions about value functions commonly made in the MDP literature are *time-separability* and *additivity*. A time-separable value function is defined in terms of more primitive functions that can be applied to component states and actions. The *reward function* $R : \mathcal{S} \to \mathbf{R}$ associates a reward with being in a state $s$. Costs can be assigned to taking actions by defining a *cost function* $C : \mathcal{S} \times \mathcal{A} \to \mathbf{R}$ that associates a cost with performing an action $a$ in state $s$. Rewards are added to the value function, while costs are subtracted.[11]

A value function is *time-separable* if it is a "simple combination" of the rewards and costs accrued at each stage. "Simple combination" means that value is taken to be a function of costs and rewards at each stage, where the costs and rewards can depend on the stage $t$, but the function that combines these must be *independent* of the stage, most commonly a linear combination or a product.[12] A value function is *additive* if the combination function is a *sum* of the reward and cost function values accrued over the history's stages. The addition of rewards and action costs in a system with time-separable value can be viewed graphically as shown in Figure 10.

The assumption of time-separability is restrictive. In our example, there might be certain goals involving temporal deadlines (have the workplace tidy as soon as possible after 9:00 tomorrow morning) and maintenance (do not allow mail to sit in the mailroom

10. Technically, the set of histories of interest also depends on the *horizon* chosen, as described below.
11. The term "reward" is somewhat of a misnomer in that the reward could be negative, in which case "penalty" might be a better word. Likewise, "costs" can be either positive (punitive) or negative (beneficial). Thus, they admit great flexibility in defining value functions.
12. See (Luenberger, 1973) for a more precise definition of time-separability.

undelivered for more than 10 minutes) that require value functions that are non-separable given our current representation of the state. Note, however, that separability—like the Markov property—is a property of a particular representation. We could add additional information to the state in our example: the clock time, the interval of time between 9:00 and the time at which tidiness is achieved, the length of time mail sits in the mail room before the robot picks it up, and so on. With this additional information we could re-establish a time-separable value function, but at the expense of an increase in the number of states and a more *ad hoc* and cumbersome action representation.[13]

### 2.7 Horizons and Success Criteria

In order to evaluate a particular course of action, we need to specify how long (in how many stages) it will be executed. This is known as the problem's *horizon*. In *finite-horizon problems*, the agent's performance is evaluated over a fixed, finite number of stages $T$. Commonly, our aim is to maximize the total expected reward associated with a course of action; we therefore define the (finite-horizon) value of any length $T$ history $h$ as (Bellman, 1957):

$$V(h) = \sum_{t=0}^{T-1} \{R(s^t) - C(s^t, a^t)\} + R(s^T)$$

An *infinite-horizon problem*, on the other hand, requires that the agent's performance be evaluated over an infinite trajectory. In this case the total reward may be unbounded, meaning that any policy could be arbitrarily good or bad if it were executed for long enough. In this case it may be necessary to adopt a different means of evaluating a trajectory. The most common is to introduce a *discount factor*, ensuring that rewards or costs accrued at later stages are counted less than those accrued at earlier stages. The value function for an *expected total discounted reward* problem is defined as follows (Bellman, 1957; Howard, 1960):

$$V(h) = \sum_{t=0}^{\infty} \gamma^t (R(s^t) - C(s^t, a^t))$$

where $\gamma$ is a fixed *discount rate* ($0 \leq \gamma < 1$). This formulation is a particularly simple and elegant way to ensure a bounded measure of value over an infinite horizon, though it is important to verify that discounting is in fact appropriate. Economic justifications are often provided for discounted models—a reward earned sooner is worth more than one earned later provided the reward can somehow be invested. Discounting can also be suitable for modeling a process that terminates with probability $1 - \gamma$ at at any point in time (e.g., a robot that can break down), in which case discounted models correspond to expected total reward over a finite but uncertain horizon. For these reasons, discounting is sometimes used for finite-horizon problems as well.

Another technique for dealing with infinite-horizon problems is to evaluate a trajectory based on the average reward accrued per stage, or *gain*. The gain of a history is defined as

$$g(h) = \lim_{n \to \infty} \frac{1}{n} \sum_{t=0}^{n} \{R(s^t) - C(s^t, a^t)\}$$

13. See (Bacchus, Boutilier, & Grove, 1996, 1997), however, for a systematic approach to handling certain types of history-dependent reward functions.

Refinements of this criterion have also been proposed (Puterman, 1994).

Sometimes the problem itself ensures that total reward over any infinite trajectory is bounded, and thus the expected total reward criterion is well-defined. Consider the case common in AI planners in which the agent's task is to bring the system to a goal state. A positive reward is received only when the goal is reached, all actions incur a non-negative cost, and when a goal is reached the system enters an absorbing state in which no further rewards or costs are accrued. As long as the goal can be reached with certainty, this situation can be formulated as an infinite-horizon problem where total reward is bounded for any desired trajectory (Bertsekas, 1987; Puterman, 1994). In general, such problems cannot be formulated as (fixed) finite-horizon problems unless an *a priori* bound on the number of steps needed to reach the goal can be established. These problems are sometimes called *indefinite-horizon* problems: from a practical point of view, the agent will continue to execute actions for some finite number of stages, but the exact number cannot be determined ahead of time.

### 2.8 Solution Criteria

To complete our definition of the planning problem we need to specify what constitutes a *solution* to the problem. Here again we see a split between explicit MDP formulations and work in the AI planning community. Classical MDP problems are generally stated as *optimization problems*: given a value function, a horizon, and an evaluation metric (e.g., expected total reward, expected total discounted reward, expected average reward per stage) the agent seeks a behavioral policy that *maximizes* the objective function.

Work in AI often seeks *satisficing* solutions to such problems. In the planning literature, it is generally taken that any plan that satisfies the goal is equally preferred to any other plan that satisfies the goal, and that any plan that satisfies the goal is preferable to any plan that does not.[14] In a probabilistic framework, we might seek the plan that satisfies the goal with maximum probability (an optimization), but this can lead to situations in which the optimal plan has infinite length if the system state is not fully observable. The satisficing alternative (Kushmerick, Hanks, & Weld, 1995) is to seek *any* plan that satisfies the goal with a probability exceeding a given threshold.

**Example 2.4** We extend our running example to demonstrate an infinite-horizon, fully observable, discounted reward situation. We begin by adding one new dimension to the state description, the boolean variable *RHM* (does the robot have mail), giving us a system with 20 states. We also provide the agent with two additional actions: $PUM$ (pickup mail) and *DelM* (deliver mail) as described in Figure 2. We can now reward the agent in such a way that mail delivery is encouraged: we associate a reward of 10 with each state in which *RHM* and $M$ are both false and 0 with all other states. If actions have no cost, the agent gets a total reward of 20 for this six-stage system trajectory:

$$\langle Loc(M), \overline{M}, \overline{RHM}\rangle, Stay, \langle Loc(M), M, \overline{RHM}\rangle, PUM, \langle Loc(M), \overline{M}, RHM\rangle,$$
$$Clk, \langle Loc(H), \overline{M}, RHM\rangle, Clk, \langle Loc(O), \overline{M}, RHM\rangle, DelM, \langle Loc(O), \overline{M}, \overline{RHM}\rangle$$

14. Though see (Haddawy & Hanks, 1998; Williamson & Hanks, 1994) for a restatement of planning as an optimization problem.

If we assign an action cost of $-1$ for each action except *Stay* (which has 0 cost), the total reward becomes 16. If we use a discount rate of 0.9 to discount future rewards and costs, this initial segment of an infinite-horizon history would contribute $10 + .9(-1) + .81(-1) + .729(-1) + .6561(-1) + .59054(-1 + 10) = 12.2$ to the total value of the trajectory (as subsequently extended). Furthermore, we can establish a bound on the total expected value of this trajectory. In the best case, all subsequent stages will yield a reward of 10, so the expected total discounted reward is bounded by

$$12.2 + .9^6(10) + .9^7(10) + \ldots = 12.2 + 10 * .9^6 \sum_{i=0}^{\infty} 0.9^i < 66$$

A similar effect on behavior can be achieved by penalizing states (i.e., having negative rewards) in which either $M$ or $RHM$ is true. □

## 2.9 Policies

We have mentioned policies (or courses of action, or plans) informally to this point, and now provide a precise definition. The decision problem facing an agent can be viewed most generally as deciding which action to perform given the current observable history. We define a *policy* $\pi$ to be a mapping from the set of observable histories $\mathcal{H}_O$ to actions, that is, $\pi : \mathcal{H}_O \rightarrow \mathcal{A}$. Intuitively, the agent executes action

$$a^t = \pi(\langle\langle o^0, a^0\rangle, \ldots, \langle o^{t-1}, a^{t-1}\rangle, o^t\rangle)$$

at stage $t$ if it has performed the actions $a^0, \cdots a^{t-1}$ and made observations $o^0, \cdots o^{t-1}$ at earlier stages, and has just made observation $o^t$ at the current stage.

A policy induces a distribution $\Pr(h|\pi)$ over the set of system histories $\mathcal{H}_S$; this probability distribution depends on the initial distribution $P^0$. We define the *expected value of a policy* to be:

$$\mathbf{EV}(\pi) = \sum_{h \in \mathcal{H}_S} \mathrm{V}(h) \Pr(h|\pi)$$

We would like the agent to adopt a policy that either maximizes this expected value or, in a satisficing context, has an acceptably high expected value.

The general form of a policy, depending as it does on an arbitrary observation history, can lead to very complicated policies and policy-construction algorithms. In special cases, however, assumptions about observability and the structure of the value function can result in optimal policies that have a much simpler form.

In the case of a fully observable MDP with a time-separable value function, the optimal action at any stage can be computed using only information about the *current* state and the stage: that is, we can restrict policies to have the simpler form $\pi : \mathcal{S} \times \mathcal{T} \rightarrow \mathcal{A}$ without danger of acting suboptimally. This is due to the fact that full observability allows the state to be observed completely, and the Markov assumption renders prior history irrelevant.

In the non-observable case, the observational history contains only vacuous observations and the agent must choose its actions using only knowledge of its previous actions and the stage; however, since $\pi$ incorporates previous actions, it takes the form $\pi : \mathcal{T} \rightarrow \mathcal{A}$. This form of policy corresponds to a linear, unconditional sequence of actions $\langle a^1, a^2, \ldots, a^T \rangle$, or a *straight-line plan* in AI nomenclature.[15]

## 2.10 Model Summary: Assumptions, Problems, and Computational Complexity

This concludes our exposition of the MDP model for planning under uncertainty. Its generality allows us to capture a wide variety of the problem classes that are currently being studied in the literature. In this section we review the basic components of the model, describe problems commonly studied in the DTP literature with respect to this model, and summarize known complexity results for each. In Section 3, we describe some of the specialized computational techniques used to solve problems in each of these problem classes.

### 2.10.1 Model Summary and Assumptions

The MDP model consists of the following components:

- The state space $\mathcal{S}$, a finite or countable set of states. We generally make the Markov assumption, which requires that each state convey all information necessary to predict the effects of all actions and events independent of any further information about system history.
- The set of actions $\mathcal{A}$. Each action $a_k$ is represented by a transition matrix of size $|S| \times |S|$ representing the probability $p_{ij}^k$ that performing action $a_k$ in state $s_i$ will move the system into state $s_j$. We assume throughout that the action model is *stationary*, meaning that transition probabilities do not vary with time. The transition matrix for an action is generally assumed to account for any exogenous events that might occur at the stage at which the action is executed.
- The set of observation variables $\mathcal{O}$. This is the set of "messages" sent to the agent after an action is performed, that provide execution-time information about the current system state. With each action $a_k$ and pair of states $s_i$, $s_j$, such that $p_{ij}^k > 0$, we associate a distribution over possible observations: $p_{ij}^{km}$ denotes the probability of obtaining observation $o_m$ given that action $a_k$ was taken in $s_i$ and resulted in a transition to state $s_j$.
- The value function $V$. The value function maps a state history into a real number such that $V(h_1) \geq V(h_2)$ just in case the agent considers history $h_1$ at least as good as $h_2$. A state history records the progression of states the system assumes along with the actions performed. Assumptions such as time-separability and additivity are common for $V$. In particular, we generally use a reward function $R$ and cost function $C$ when defining value.
- The horizon $T$. This is the number of stages over which the state histories should be evaluated using $V$.

15. Many algorithms in the AI literature produce a *partially ordered* sequence of actions. These plans do not, however, involve conditional or nondeterministic execution. Rather, they represent the fact that *any* linear sequence consistent with the partial order will solve the problem. Thus, a partially ordered plan is a concise representation for a particular *set* of straight-line plans.

- An optimality criterion. This provides a criterion for evaluating potential solutions to planning problems.

### 2.10.2 Common Planning Problems

We can use this general framework to classify various problems commonly studied in the planning and decision-making literature. In each case below, we note the modeling assumptions that define the problem class.

**Planning Problems in the OR/Decision Sciences Tradition**

- **Fully Observable Markov Decision Processes (FOMDPs)** — There is an extremely large body of research studying FOMDPs, and we present the basic algorithmic techniques in some detail in the next section. The most commonly used formulation of FOMDPs assumes full observability and stationarity, and uses as its optimality criterion the maximization of expected total reward over a finite horizon, maximization of expected total discounted reward over an infinite horizon, or minimization of the expected cost to a goal state.

  FOMDPs were introduced by Bellman (1957) and have been studied in depth in the fields of decision analysis and OR, including the seminal work of Howard (1960). Recent texts on FOMDPs include (Bertsekas, 1987) and (Puterman, 1994). Average reward optimality has also received attention in this literature (Blackwell, 1962; Howard, 1960; Puterman, 1994). In the AI literature, discounted or total reward models have been most popular as well (Barto et al., 1995; Dearden & Boutilier, 1997; Dean, Kaelbling, Kirman, & Nicholson, 1995; Koenig, 1991), though the average-reward criterion has been proposed as more suitable for modeling AI planning problems (Boutilier & Puterman, 1995; Mahadevan, 1994; Schwartz, 1993).

- **Partially Observable Markov Decision Processes (POMDPs)** — POMDPs are closer than FOMDPs to the general model of decision processes we have described. POMDPs have generally been studied with the assumption of stationarity and optimality criteria identical to those of FOMDPs, though the average-reward criterion has not been widely considered. As we discuss below, a POMDP can be viewed as a FOMDP with a state space consisting of the set of *probability distributions* over $S$. These probability distributions represent states of belief: the agent can "observe" its state of belief about the system although it does not have exact knowledge of the system state itself.

  POMDPs have been widely studied in OR and control theory (Aström, 1965; Lovejoy, 1991b; Smallwood & Sondik, 1973; Sondik, 1978), and have drawn increasing attention in AI circles (Cassandra, Kaelbling, & Littman, 1994; Hauskrecht, 1998; Littman, 1996; Parr & Russell, 1995; Simmons & Koenig, 1995; Thrun, Fox, & Burgard, 1998; Zhang & Liu, 1997). Influence diagrams (Howard & Matheson, 1984; Shachter, 1986) are a popular model for decision making in AI and are, in fact, a structured representational method for POMDPs (see Section 4.3).

**Planning Problems in the AI Tradition**

- **Classical Deterministic Planning** — The classical AI planning model assumes deterministic actions: any action $a_k$ taken at any state $s_i$ has at most one successor $s_j$. The other important assumptions are non-observability and that value is determined by reaching a *goal state*: any plan that leads to a goal state is preferred to any that does not. Often there is a preference for shorter plans; this can be represented by using a discount factor to "encourage" faster goal achievement or by assigning a cost to actions. Reward is associated only with transitions to goal states, which are absorbing. Action costs are typically ignored, except as noted above.

  In classical models it is usually assumed that the initial state is known with certainty. This contrasts with the general specification of MDPs above, which does not assume knowledge of or even distributional information about the initial state. Policies are defined to be applicable no matter what state (or distribution over states) one finds oneself in—action choices are defined for every possible state or history. Knowledge of the initial state and determinism allow optimal straight-line plans to be constructed, with no loss in value associated with non-observability, but unpredictable exogenous events and uncertain action effects cannot be modeled consistently if these assumptions are adopted.

  For an overview of early classical planning research and the variety of approaches adopted, see (Allen, Hendler, & Tate, 1990) as well as Yang's (1998) recent text.

- **Optimal Deterministic Planning** — A separate body of work retains the classical assumptions of complete information and determinism, but tries to recast the planning problem as an optimization that relaxes the implicit assumption of "achieve the goal at all costs." At the same time, these methods use the same sort of representations and algorithms applied to satisficing planning.

  Haddawy and Hanks (1998) present a multi-attribute utility model for planners that keeps the explicit information about the initial state and goals, but allows preferences to be stated about the partial satisfaction of the goals as well as the cost of the resources consumed in satisfying them. The model also allows the expression of preferences over phenomena like temporal deadlines and maintenance intervals that are difficult to capture using a time-separable additive value function. Williamson (1996) (see also Williamson & Hanks, 1994). implements this model by extending a classical planning algorithm to solve the resulting optimization problem. Haddawy and Suwandi (1994) also implement this model in a complete decision-theoretic framework. Their model of planning, *refinement planning*, differs somewhat from the generative model discussed in this paper. In their model the set of all possible plans is pre-stored in an abstraction hierarchy, and the problem solver's job is to find in the hierarchy the optimal choice of concrete actions for a particular problem.

  Perez and Carbonell's (1994) work also incorporates cost information into the classical planning framework, but maintains the split between a classical satisficing planner and additional cost information provided in the utility model. The cost information is used to learn search-control rules that allow the classical planner to generate low-cost goal-satisfying plans.

- **Conditional Deterministic Planning** — The classical planning assumption of omniscience can be relaxed somewhat by allowing the state of some aspects of the world to be *unknown*. The agent is thus in a situation where it is certain that the system is one of a particular set of states, but does not know which one. Unknown truth values can be included in the initial state specification, and taking actions can cause a proposition to become unknown as well.

  Actions can provide the agent with information while the plan is being executed: conditional planners introduce the idea of actions providing *runtime information* about the prevailing state, distinguishing between an action that makes proposition $P$ true and an action that will tell the agent *whether* $P$ is true when the action is executed. An action can have both causal and informational effects, simultaneously changing the world and reporting on the value of one or more propositions. This second sort of information is not useful at planning time except that it allows steps in the plan to be executed conditionally, depending on the runtime information provided by prior information-producing steps. The value of such actions lies in the fact that different courses of action may be appropriate under different conditions—these informational effects allow runtime selection of actions based on the observations produced, much like the general POMDP model.

  Examples of conditional planners in the classical framework include early work by Warren (1976) and the more recent CNLP (Peot & Smith, 1992), Cassandra (Pryor & Collins, 1993), Plynth (Goldman & Boddy, 1994), and UWL (Etzioni, Hanks, Weld, Draper, Lesh, & Williamson, 1992) systems.

- **Probabilistic Planning Without Feedback** — A direct probabilistic extension of the classical planning problem can be stated as follows (Kushmerick et al., 1995): take as input (a) a *probability distribution* over initial states, (b) stochastic actions (explicit or implicit transition matrices), (c) a set of goal states, and (d) a probability success threshold $\tau$. The objective is to produce a plan that reaches any goal state with probability at least $\tau$, given the initial state distribution. No provision is made for execution-time observation, thus straight-line plans are the only form of policy possible. This is a restricted case of the infinite-horizon NOMDP problem, one in which actions incur no cost and goal states offer positive reward and are absorbing. It is also a special case in that the objective is to find a *satisficing* policy rather than an optimal one.

- **Probabilistic Planning With Feedback** — Draper et al. (1994a) have proposed an extension of the probabilistic planning problem in which actions provide feedback, using exactly the observation model described in Section 2.4. Again, the problem is posed as that of building a plan that leaves the system in a goal state with sufficient probability. But a plan is no longer a simple sequence of actions—it can contain conditionals and loops whose execution depends on the observations generated by sensing actions. This problem is a restricted case of the general POMDP problem: absorbing goal states and cost-free actions are used, and the objective is to find any policy (conditional plan) that leaves the system in a goal state with sufficient probability.

**Comparing the Frameworks: Task-oriented Versus Process-oriented Problems** It is useful at this point to pause and contrast the types of problems considered in the classical planning literature with those typically studied within the MDP framework. Although problems in the AI planning literature have emphasized a goal-pursuit or "one-shot" view of problem solving, in some cases viewing the problem as an infinite-horizon decision problem results in a more satisfying formulation. Consider our running example involving the office robot. It is simply not possible to model the problem of responding to coffee requests, mail arrival and keeping the lab tidy as a strict *goal-satisfaction problem* while capturing the possible nuances of intuitively optimal behavior.

The primary difficulty is that no explicit and persistent goal states exist. If we were simply to require that the robot attain a state where the lab is tidy, no mail awaits, and no unfilled coffee requests exist, no "successful" plan could anticipate possible system behavior after a goal state was reached. The possible occurrence of exogenous events *after* goal achievement requires that the robot bias its methods for achieving its goals in a way that best suits the expected course of subsequent events. For instance, if coffee requests are very likely at any point in time and unmet requests are highly penalized, the robot should situate itself in the coffee room in order to satisfy an anticipated *future* request quickly. Most realistic decision scenarios involve both *task-oriented* and *process-oriented* behavior, and problem formulations that take both into account will provide more satisfying models for a wider range of situations.

### 2.10.3 The Complexity of Policy Construction

We have now defined the planning problem in several different ways, each having a different set of assumptions about the state space, system dynamics and actions (deterministic or stochastic), observability (full, partial, or none), value function (time-separable, goal only, goal rewards and action costs, partially satisfiable goals with temporal deadlines), planning horizon (finite, infinite, or indefinite), and optimality criterion (optimal or satisficing solutions). Each set of assumptions puts the corresponding problem in a particular complexity class, which defines worst-case time and space bounds on *any* representation and algorithm for solving that problem. Here we summarize known complexity results for each of the problem classes defined above.

**Fully Observable Markov Decision Processes** Fully observable MDPs (FOMDPs) with time-separable, additive value functions can be solved in time polynomial in the size of the state space, the number of actions, and the size of the inputs.[16] The most common algorithms for solving FOMDPs are *value iteration* and *policy iteration*, which are described in the next section. Both finite-horizon and discounted infinite-horizon problems require a polynomial amount of computation per iteration—$O(|S|^2|A|)$ and $O(|S|^2|A|+|S|^3)$, respectively—and converge in a polynomial number of iterations (with factor $\frac{1}{1-\gamma}$ in the discounted case). On the other hand, these problems have been shown to be P-complete (Papadimitriou & Tsitsiklis, 1987), which means that an efficient parallel solution algorithm is unlikely.[17] The space required to store the policy for an $n$-stage finite-horizon problem

16. More precisely, the maximum number of bits required to represent any of the transition probabilities or costs.
17. See (Littman, Dean, & Kaelbling, 1995) for a summary of these complexity results.

is $O(|S|n)$. For most interesting classes of infinite-horizon problems, specifically those involving discounted models with time-separable additive reward, the optimal policy can be shown to be *stationary*, and the policy can be stored in $O(|S|)$ space.

Bear in mind that these are worst-case bounds. In many cases, better time bounds and more compact representations can be found. Sections 4 and 5 explore ways to represent and solve these problems more efficiently.

**Partially Observable Markov Decision Processes** POMDPs are notorious for their computational difficulty. As mentioned above, a POMDP can be viewed as a FOMDP with an infinite state space consisting of *probability distributions* over $S$, each distribution representing the agent's state of belief at a point in time (Aström, 1965; Smallwood & Sondik, 1973). The problem of finding an optimal policy for a POMDP with the objective of maximizing expected total reward or expected total discounted reward over a finite horizon $T$ has been shown to be exponentially hard both in $|S|$ and in $T$ (Papadimitriou & Tsitsiklis, 1987). The problem of finding a policy that maximizes or approximately maximizes the expected discounted total reward over an infinite horizon is shown to be undecidable (Madani, Condon, & Hanks, 1999).

Even restricted cases of the POMDP problem are computationally difficult in the worst case. Littman (1996) considers the special case of boolean rewards: determining whether there is an infinite-horizon policy with nonzero total reward given that the rewards associated with all states are non-negative. He shows that the problem is EXPTIME-complete if the transitions are stochastic, and PSPACE-hard if the transitions are deterministic.

**Deterministic Planning** Recall that the classical planning problem is defined quite differently from the MDP problems above: the agent has no ability to observe the state but has perfect predictive powers, knowing the initial state and the effects of all actions with certainty. In addition, rewards come only from reaching a goal state, and any plan that achieves the goal suffices.

Planning problems are typically defined in terms of a set $\mathcal{P}$ of boolean features or propositions: a complete assignment of truth values to features describes exactly one state, and a partial assignment of truth values describes a set of states. A set of propositions $\mathcal{P}$ induces a state space of size $2^{|\mathcal{P}|}$. Thus, the space required to represent a planning problem using a feature-based representation can be exponentially smaller than that required by a flat representation for the same problem (see Section 4).

The ability to represent planning problems compactly has a dramatic impact on worst-case complexity. Bylander (1994) shows that the deterministic planning problem without observation is PSPACE-complete. Roughly speaking, this means that at worst planning time will increase exponentially with $\mathcal{P}$ and $\mathcal{A}$, and further, that the size of a solution plan can grow exponentially with the problem size. These results hold even when the action space $\mathcal{A}$ is severely restricted. For example, the planning problem is NP-complete even in cases where each action is restricted to one precondition feature and one postcondition feature. Conditional and optimal planning are PSPACE-complete as well. These results are for inputs that are generally more compact (generally exponentially so) than those in terms of which the complexity of the FOMDP and POMDP problems are phrased.

**Probabilistic Planning** In probabilistic goal-oriented planning, as for POMDPs, we typically search for a solution in a space of *probability distributions* over states (or over formulas that describe states). Even the simplest problem in probabilistic planning—one that admits *no* observability—is undecidable at worst (Madani et al., 1999). The intuition is that even though the set of states is finite, the set of distributions over those states is not, and at worst the agent may have to search an infinite number of plans before being able to determine whether or not a solution exists. An algorithm can be guaranteed to find a solution plan eventually if one exists, but cannot be guaranteed to terminate in finite time if there is no solution plan. Conditional probabilistic planning is a generalization of the non-observable probabilistic planning problem, and thus is undecidable as well.

It is interesting to note a connection between conditional probabilistic planning and POMDPs. The actions and observations of the two problems have equivalent expressive power, but the reward structure of the conditional probabilistic planning problem is quite restrictive: goal states have positive rewards, all other states have no reward, and goal states are absorbing. Since we cannot put an *a priori* bound on the length of a solution plan, conditional probabilistic planning must be viewed as an infinite-horizon problem where the objective is to maximize total expected *undiscounted* reward. Note, however, that since goal states are absorbing, we can guarantee that total expected reward will be non-negative and bounded, even over an infinite horizon. Technically this means that the conditional probabilistic planning problem is a restricted case of an infinite-horizon *positive-bounded* problem (Puterman, 1994, Section 7.2). We can therefore conclude that the problem of solving an arbitrary infinite-horizon undiscounted positive-bounded POMDP is also undecidable. The more commonly studied problem is the infinite-horizon POMDP with a criterion of maximizing expected *discounted* total reward; finding optimal or near-optimal solutions to that problem is also undecidable, as noted above.

#### 2.10.4 Conclusion

We end this section by noting again that these results are algorithm-independent and describe worst-case behavior. In effect, they indicate how badly any algorithm can be made to perform on an "arbitrarily unfortunate" problem instance. The more interesting question is whether we can build representations, techniques, and algorithms that typically perform well on problem instances that typically arise in practice. This concern leads us to examine the problem characteristics with an eye toward exploiting the restrictions placed on the states and actions, on observability, and on the value function and optimality criterion. We begin with algorithmic techniques that focus on the value function—particularly those that take advantage of time-separability and goal orientation. Then in the following section we explore complementary techniques for building compact problem representations.

## 3. Solution Algorithms: Dynamic Programming and Search

In this section we review standard algorithms for solving the problems described above in terms of the "unstructured" or "flat" problem representations. As noted in the analysis above, fully observable Markov decision processes (FOMDPs) are by far the most widely studied models in this general class of stochastic sequential decision problems. We begin by describing techniques for solving FOMDPs, focusing on techniques that exploit structure in the value function like time-separability and additivity.

### 3.1 Dynamic Programming Approaches

Suppose we are given a fully-observable MDP with a time-separable, additive value function. In other words, we are given the state space $\mathcal{S}$, action space $\mathcal{A}$, a transition matrix $\Pr(s'|s,a)$ for each action $a$, a reward function $R$, and a cost function $C$. We start with the problem of finding the policy that maximizes expected total reward for some fixed, finite-horizon $T$. Suppose we are given a policy $\pi$ such that $\pi(s,t)$ is the action to be performed by the agent in state $s$ with $t$ stages remaining to act (for $0 \leq t \leq T$).[18] Bellman (1957) shows that the expected value of such a policy at any state can be computed using the set of *$t$-stage-to-go value functions* $V_t^\pi$. We define $V_0^\pi(s)$ to be $R(s)$, then define:

$$V_t^\pi(s) = R(s) + C(\pi(s,t)) + \sum_{s' \in \mathcal{S}} \{\Pr(s'|\pi(s,t),s) V_{t-1}^\pi(s')\} \tag{1}$$

This definition of the value function for $\pi$ makes its dependence on the initial state clear.

We say a policy $\pi$ is *optimal* if $V_T^\pi(s) \geq V_T^{\pi'}(s)$ for all policies $\pi'$ and all $s \in \mathcal{S}$. The optimal $T$-stage-to-go value function, denoted $V_T^*$, is simply the value function of any optimal $T$-horizon policy. Bellman's *principle of optimality* (Bellman, 1957) forms the basis of the stochastic dynamic programming algorithms used to solve MDPs, establishing the following relationship between the optimal value function at $t^{th}$ stage and the optimal value function at the previous stage:

$$V_t^*(s) = R(s) + \max_{a \in \mathcal{A}} \{C(a) + \sum_{s' \in \mathcal{S}} \Pr(s'|a,s) V_{t-1}^*(s')\} \tag{2}$$

#### 3.1.1 Value Iteration

Equation 2 forms the basis of the *value iteration algorithm* for finite-horizon problems. Value iteration begins with the value function $V_0^* = R$, and uses Equation 2 to compute in sequence the value functions for longer time intervals, up to the horizon $T$. Any action that maximizes the right-hand side of Equation 2 can be chosen as the policy element $\pi(s,t)$. The resulting policy is optimal for the $T$-stage, fully observable MDP, and indeed for any shorter horizon $t < T$.

It is important to note that a policy describes what should be done at every stage and for every state of the system, even if the agent cannot reach certain states given the system's initial configuration and its available actions. We return to this point below.

**Example 3.1** Consider a simplified version of the robot example, in which we have four state variables $M$, $CR$, $RHC$, and $RHM$ (movement to various locations is ignored), and four actions *GetC*, *PUM*, *DelC*, and *DelM*. Actions *GetC* and *PUM* make *RHC* and *RHM*, respectively, true with certainty. Action *DelM*, when *RHM* holds, makes both $M$ and *RHM* false with probability 1.0; *DelC* makes both *CR* and *RHC* false with probability 0.3, leaving the state unchanged with probability 0.7. A reward of 3 is associated with $\overline{CR}$ and a reward of 1 is associated with $\overline{M}$. The reward for any state is the sum of the rewards for each objective satisfied in that state. Figure 11 shows the optimal 0-stage, 1-stage and 2-stage value functions for various states, along with

18. Recall that for FOMDPs other aspects of history are not relevant.

| State | $V_0^*$ | $V_1^*$ | $\pi(1)$ | $V_2^*$ | $\pi(2)$ |
|---|---|---|---|---|---|
| $s_0 = \langle M, \overline{RHM}, CR, \overline{RHC} \rangle$ | 0 | 0 | any | 1 | *PUM* |
| $s_1 = \langle M, RHM, CR, \overline{RHC} \rangle$ | 0 | 1 | *DelM* | 2 | *DelM* |
| $s_2 = \langle M, \overline{RHM}, CR, RHC \rangle$ | 0 | 0.9 | *DelC* | 2.43 | *DelC* |
| $s_3 = \langle M, RHM, CR, RHC \rangle$ | 0 | 1 | *DelM* | 2.9 | *DelM* |
| $s_4 = \langle \overline{M}, CR, RHC \rangle$ | 1 | 2.9 | *DelC* | 5.43 | *DelC* |
| $s_5 = \langle \overline{M}, CR, \overline{RHC} \rangle$ | 1 | 2 | any | 3.9 | *GetC* |
| $s_6 = \langle M, RHM, \overline{CR} \rangle$ | 3 | 7 | *DelM* | 11 | *DelM* |
| $s_7 = \langle M, \overline{RHM}, \overline{CR} \rangle$ | 3 | 6 | any | 10 | *PUM* |
| $s_8 = \langle \overline{M}, \overline{CR} \rangle$ | 4 | 8 | any | 12 | any |

Figure 11: Finite-horizon optimal value and policy.

the optimal choice of action at each state-stage pairing (the values for any "state" with missing variables hold for all instantiations of those variables). Note that $V_0^*(s)$ is simply $R(s)$ for each state $s$.

To illustrate the application of Equation 2, first consider the calculation of $V_1^*(s_3)$. The robot has the choice of delivering coffee or delivering mail, and the expected value of each option, with one stage remaining, is given by:

$$\begin{aligned} \mathbf{EV}_1(s_3, \mathit{DelC}) &= 0.3V_0^*(s_6) + 0.7V_0^*(s_3) &= 0.9 \\ \mathbf{EV}_1(s_3, \mathit{DelM}) &= 1.0V_0^*(s_4) &= 1.0 \end{aligned}$$

Thus $\pi^*(s_3, 1) = \mathit{DelM}$ and $V_1^*(s_3)$ is the value of this maximizing choice. Notice that the robot with one action to perform will aim for the "lesser" objective $\overline{M}$ due to the risk of failure inherent in delivering coffee. With two stages remaining at the same state, the robot will again deliver mail, but with certainty will move to $s_4$ with one stage to go, where it will attempt to deliver coffee ($\pi^*(s_4, 1) = \mathit{DelC}$).

To illustrate the effects a fixed finite horizon can have on policy choice, note that $\pi^*(s_0, 2) = \mathit{PUM}$. With two stages remaining and the choice of getting mail or coffee, the robot will get mail because subsequent delivery (in the last stage) is guaranteed to succeed, whereas subsequent coffee delivery may fail. However, if we compute $\pi^*(s_0, 3)$, we see:

$$\begin{aligned} \mathbf{EV}_3(s_0, \mathit{GetC}) &= 1.0V_2^*(s_2) &= 2.43 \\ \mathbf{EV}_3(s_0, \mathit{PUM}) &= 1.0V_2^*(s_1) &= 2.0 \end{aligned}$$

With *three* stages to go, the robot will instead retrieve coffee at $s_0$. Once it has coffee, it has two chances at successful delivery. The expected value of this course of action is greater than that of (guaranteed) mail delivery. Note that three stages does not allow sufficient time to try to achieve both objectives at $s_0$. In fact, the larger reward associated with coffee delivery ensures that with any greater number of stages remaining, the robot should focus first on coffee retrieval and delivery, and then attempt mail retrieval and delivery once coffee delivery is successfully completed. □

Often we are faced with tasks that do not have a fixed finite horizon. For example, we may want our robot to perform the tasks of keeping the lab tidy, picking up mail whenever it arrives, responding to coffee requests, and so on. There is no fixed time horizon associated with these tasks—they are to be performed as the need arises. Such problems are best modeled as infinite-horizon problems.

We consider here the problem of building a policy that maximizes the discounted sum of expected rewards over an infinite horizon.[19] Howard (1960) showed that there always exists an optimal *stationary* policy for such problems. Intuitively, this is the case because no matter what stage the process is in, there are still an infinite number of stages remaining; so the optimal action at any state is independent of the stage. We can therefore restrict our attention to policies that choose the same action for a state regardless of the stage of the process. Under this restriction, the policy will have the same size $|\mathcal{S}|$ regardless of the number of stages over which the policy is executed—the policy $\pi$ has the form $\pi : \mathcal{S} \rightarrow \mathcal{A}$. In contrast, optimal policies for finite-horizon problems are generally nonstationary, as illustrated in Example 3.1.

Howard also shows that the value of policy $\pi$ satisfies the following recurrence:

$$V^{\pi}(s) = R(s) + \{C(\pi(s)) + \gamma \sum_{s' \in \mathcal{S}} \Pr(s'|\pi(s), s)V^{\pi}(s')\} \tag{3}$$

and that the optimal value function $V^*$ satisfies:

$$V^*(s) = R(s) + \max_{a \in \mathcal{A}} \{C(a) + \gamma \sum_{s' \in \mathcal{S}} \Pr(s'|a, s)V^*(s')\} \tag{4}$$

The value of a fixed policy $\pi$ can be evaluated using the method of *successive approximation*, which is almost identical to the procedure described in Equation 1 above. We begin with an arbitrary assignment of values to $V_0^{\pi}(s)$, then define:

$$V_t^{\pi}(s) = R(s) + C(\pi(s,t)) + \gamma \sum_{s' \in \mathcal{S}} \{\Pr(s'|\pi(s,t), s)V_{t-1}^{\pi}(s')\} \tag{5}$$

The sequence of functions $V_t^{\pi}$ converges linearly to the true value function $V^{\pi}$.

One can also alter the value-iteration algorithm slightly so it builds optimal policies for infinite-horizon discounted problems. The algorithm starts with a value function $V_0$ that assigns an arbitrary value to each $s \in S$. Given value estimate $V_t(s)$ for each state $s$, $V_{t+1}(s)$ is calculated as:

$$V_{t+1}(s) = R(s) + \max_{a \in \mathcal{A}} \{C(a) + \gamma \sum_{s' \in S} \Pr(s'|a, s) \cdot V_t(s')\} \tag{6}$$

The sequence of functions $V_t$ converges linearly to the optimal value function $V^*(s)$. After some finite number of iterations $n$, the choice of maximizing action for each $s$ forms an optimal policy $\pi$ and $V_n$ approximates its value.[20]

19. This is by far the most commonly studied problem in the literature, though it is argued in (Boutilier & Puterman, 1995; Mahadevan, 1994; Schwartz, 1993) that such problems are often best modeled using average reward per stage as the optimality criterion. For a discussion of average reward optimality and its many variants and refinements, see (Puterman, 1994).

20. The number of iterations $n$ is based on a *stopping criterion* that generally involves measuring the difference between $V_t$ and $V_{t+1}$. For a discussion of stopping criteria and convergence of the algorithm, see (Puterman, 1994).

#### 3.1.2 Policy Iteration

Howard's (1960) *policy-iteration algorithm* is an alternative to value iteration for infinite-horizon problems. Rather than iteratively improving the estimated value function, it instead modifies the policies directly. It begins with an arbitrary policy $\pi_0$, then iterates, computing $\pi_{i+1}$ from $\pi_i$.

Each iteration of the algorithm comprises two steps, *policy evaluation* and *policy improvement*:

1. (Policy evaluation) For each $s \in \mathcal{S}$, compute the value function $V^{\pi_i}(s)$ based on the current policy $\pi_i$.

2. (Policy improvement) For each $s \in \mathcal{S}$, find the action $a^*$ that maximizes

$$Q_{i+1}(a, s) = R(s) + C(a) + \gamma \sum_{s' \in S} \Pr(s'|a, s) \cdot V^{\pi_i}(s') \tag{7}$$

   If $Q_{i+1}(a^*, s) > V^{\pi_i}(s)$, let $\pi_{i+1} = a^*$; otherwise $\pi_{i+1}(s) = \pi_i(s)$.[21]

The algorithm iterates until $\pi_{i+1}(s) = \pi_i(s)$ for all states $s$. Step 1 evaluates the current policy by solving the $N \times N$ linear system represented by Equation 3 (one equation for each $s \in \mathcal{S}$), and can be computationally expensive. However, the algorithm converges to an optimal policy at least linearly and under certain conditions converges superlinearly or quadratically (Puterman, 1994). In practice, policy iteration tends to converge in many fewer iterations than does value iteration. Policy iteration thus spends more computational time at each individual stage, with the result that fewer stages need be computed.[22]

*Modified policy iteration* (Puterman & Shin, 1978) provides a middle ground between policy iteration and value iteration. The structure of the algorithm is exactly the same as that of policy iteration, alternating evaluation and improvement phases. The key insight is that one need not evaluate a policy *exactly* in order to improve it. Therefore, the evaluation phase involves some (usually small) number of iterations of successive approximation (i.e., setting $V^\pi = V^\pi_t$ for some small $t$, using Equation 6). With some tuning of the value of $t$ used at each iteration, modified policy iteration can work extremely well in practice (Puterman, 1994). Both value iteration and policy iteration are special cases of modified policy iteration, corresponding to setting $t = 0$ and $t = \infty$, respectively.

A number of other variants of both value and policy iteration have been proposed. For instance, *asynchronous versions* of these algorithms do not require that the value function be constructed, or policy improved, at each state in lockstep. In the case of value iteration for infinite-horizon problems, as long as each state is updated sufficiently often, convergence can be assured. Similar guarantees can be provided for asynchronous forms of policy iteration. Such variants are important tools for understanding various online approaches to solving MDPs (Bertsekas & Tsitsiklis, 1996). For a nice discussion of asynchronous dynamic programming, see (Bertsekas, 1987; Bertsekas & Tsitsiklis, 1996).

21. The *Q-function* defined by Equation 7, and so called because of its use in *Q-learning* (Watkins & Dayan, 1992), gives the value of performing action $a$ at state $s$ assuming the value function $V^\pi$ accurately reflects future value.
22. See (Littman et al., 1995) for a discussion of the complexity of the algorithm.

### 3.1.3 Undiscounted Infinite-Horizon Problems

The difficulty with finding optimal solutions to infinite-horizon problems is that total reward can grow without limit over time. Thus, the problem definition must provide some way to ensure that the value metric is bounded over arbitrarily long horizons. The use of expected total discounted reward as the optimality criterion offers a particularly elegant way to guarantee a bound, since the infinite sum of discounted rewards is finite. However, although discounting is appropriate for certain classes of problems (e.g., economic problems, or those where the system may terminate at any point with a certain probability), for many realistic AI domains it is difficult to justify counting future rewards less than present rewards, and the discounted-reward criterion is not appropriate.

There are a variety of ways to bound total reward in undiscounted problems. In some cases the problem itself is structured so that reward is bounded. In planning problems, for example, the goal reward can be collected at most once, and all actions incur a cost. In that case total reward is bounded from above and the problem can legitimately be posed in terms of maximizing total expected undiscounted reward in many cases (e.g., if the goal can be reached with certainty).

In cases where discounting is inappropriate and total reward is unbounded, different success criteria can be employed. For example, the problem can instead be posed as one in which we wish to maximize expected *average reward per stage*, or *gain*. Computational techniques for constructing gain-optimal policies are similar to the dynamic-programming algorithms described above, but are generally more complicated, and the convergence rate tends to be quite sensitive to the communicating structure and periodicity of the MDP.

Refinements to gain optimality have also been studied. For example, *bias optimality* can be used to distinguish two gain-optimal polices by giving preference to the policy whose total reward over some initial segment of policy execution is larger. Again, while the algorithms are more complicated than those for discounted problems, they are variants of standard policy or value iteration. See (Puterman, 1994) for details.

### 3.1.4 Dynamic Programming and POMDPs

Dynamic programming techniques can be applied in partially observable settings as well (Smallwood & Sondik, 1973). The main difficulty in building policies for situations in which the state is not fully observable is that, since past observations can provide information about the system's current state, decisions must be based on information gleaned in the past. As a result, the optimal policy can depend on all observations the agent has made since the beginning of execution. These history-dependent policies can grow in size exponential in the length of the horizon. While history-dependence precludes dynamic programming, the observable history can be summarized adequately with a probability distribution over $\mathcal{S}$ (Aström, 1965), and policies can be computed as a function of these distributions, or *belief states*.

A key observation of Sondik (Smallwood & Sondik, 1973; Sondik, 1978) is that when one views a POMDP with a time-separable value function by taking the state space to be the set of probability distributions over $\mathcal{S}$, one obtains a fully observable MDP that can be solved by dynamic programming. The (computational) problem with this approach is that the state space for this FOMDP is an $N$-dimensional continuous space,[23] and special techniques must be used to solve it (Smallwood & Sondik, 1973; Sondik, 1978).

We do not explore these techniques here, but note that they are currently practical only for very small problems (Cassandra et al., 1994; Cassandra, Littman, & Zhang, 1997; Littman, 1996; Lovejoy, 1991b). A number of approximation methods, developed both in OR (Lovejoy, 1991a; White III & Scherer, 1989) and AI (Brafman, 1997; Hauskrecht, 1997; Parr & Russell, 1995; Zhang & Liu, 1997), can be used to increase the range of solvable problems, but even these techniques are presently of limited practical value.

POMDPs play a key role in reinforcement learning as well, where the "natural state space" consisting of agent observations provides incomplete information about the underlying system state (see, e.g., McCallum, 1995).

### 3.2 AI Planning and State-Based Search

We noted in Section 2.7 that the classical AI planning problem can be formulated as an infinite-horizon MDP and can therefore be solved using an algorithm like value iteration. Recall that two assumptions in classical planning specialize the general MDP model, namely determinism of actions and the use of goal states instead of a more general reward function. A third assumption—that we want to construct an optimal course of action starting from a known initial state—does not have a counterpart in the FOMDP model as presented above, since the policy dictates the optimal action from *any* state at *any* stage of the plan. As we will see below, the interest in *online* algorithms within AI has led to revised formulations of FOMDPs that do take initial and *current* states into account.

Though we defined the classical planning problem earlier as a non-observable process (NOMDP), it can be solved as if it were fully observable. We let $G$ be the set of goal states and $s_{init}$ be the initial state. Applying value iteration to this type of problem is equivalent to determining the *reachability* of goal states from all system states. For instance, if we make goal states absorbing, assign a reward of 1 to all transitions from any $s \in S - G$ to some $g \in G$ and 0 to all others, then the set of all states where $V_k^*(s) > 0$ is exactly the set of states that can lead to a goal state.[24] In particular, if $V_k^*(s_{init}) > 0$, then a successful plan can be constructed by extracting actions from the $k$-stage (finite-horizon) policy produced by value iteration. The determinism assumption means that the agent can *predict* the state perfectly at every stage of execution; the fact that it cannot *observe* the state is unimportant.

The assumptions commonly made in classical planning can be exploited computationally in value iteration. First, we can terminate the process at the first iteration $k$ where $V_k^*(s_{init}) > 0$, because we are interested only in plans that begin at $s_{init}$, not in acting optimally from every possible start state. Second, we can terminate value iteration after $|S|$ iterations: if $V_{|S|}^*(s_{init}) = 0$ at that point, the algorithm will have searched every possible state and can guarantee that no solution plan exists. Therefore, we can view classical planning as a finite-horizon decision problem with a horizon of $|S|$. This use of value iteration

23. More accurately, it is an $N$-dimensional simplex, or $(N-1)$-dimensional space.

24. Specifically, $V_k^*(s)$ indicates the probability with which one reaches the goal region under the optimal policy from $s \in S - G$ in stochastic settings. In the deterministic case being discussed, this value must be 1 or 0.

is equivalent to using the Floyd-Warshall algorithm to find a minimum-cost path through a weighted graph (Floyd, 1962).

### 3.2.1 Planning and Search

While value iteration can, in theory, be used for classical planning, it does not take advantage of the fact that the goal and initial states are known. In particular, it computes the value and policy assignment for all states at all stages. This can be very wasteful since optimal actions will be computed for states that cannot be reached from $s_{init}$ or that cannot possibly lead to any state $g \in G$. It is also problematic when $|S|$ is large, since each iteration of value iteration requires $O(|S||A|)$ computations. For this reason dynamic programming approaches have not been used extensively in AI planning.

The restricted form of the value function, especially the fact that initial and goal states are given, makes it more advantageous to view planning as a *graph-search* problem. Unlike general FOMDPs, where it is generally not known *a priori* which states are most desirable with respect to (long-term) *value*, the well-defined set of target states in a classical planning problem makes search-based algorithms appropriate. This is the approach taken by most AI planning algorithms.

One way to formulate the problem as a graph search is to make each node of the graph correspond to a state in $\mathcal{S}$. The initial state and goal states can then be identified, and the search can proceed either *forward* or *backward* through the graph, or in both directions simultaneously.

In forward search, the initial state is the root of the search tree. A node is then chosen from the tree's fringe (the set of all leaf nodes), and all feasible actions are applied. Each action application extends the plan by one step (or one stage) and generates a unique new successor state, which is a new leaf node in the tree. This node can be pruned if the state it defines is already in the tree. The search ends when a state is identified as a member of the goal set (in which case a solution plan can be extracted from the tree), or when all branches have been pruned (in which case no solution plan exists). Forward search attempts to build a plan from beginning to end, adding actions to the end of the current sequence of actions. Forward search never considers states that cannot be reached from the $s_{init}$.

Backward search can be viewed in several different ways. We could arbitrarily select some $g \in G$ as the root of the search tree, and expand the search tree at the fringe by selecting a state on the fringe and adding to the tree all states such that some action would *cause* the system to enter the chosen state. In general, an action can give rise to more than one *predecessor* vertex, even if actions are deterministic. A state can again be pruned if it appears in the search tree already. The search terminates when the initial state is added to the tree, and a solution plan can again be extracted from the tree. This search is similar to dynamic-programming-based algorithms for finding the shortest path through a graph. The difference is that backward search considers only those states at a depth $k$ in the search tree that can reach the chosen goal state within $k$ steps. Dynamic programming algorithms, in contrast, visit every state at every stage of the search.

One difficulty with the backward approach as described above is the commitment to a particular goal state. Of course, this assumption can be relaxed, as an algorithm could "simultaneously" search for paths to all goal states by adding at the first level of the search tree any vertex that can reach *some* $g \in G$. We will see in Section 5 that goal regression can be viewed as doing this, at least implicitly.

It is generally thought that *regression* (or backward) techniques are more effective in practice than *progression* (or forward) methods. The reasoning is that the branching factor in the forward graph, which is the number of actions that can feasibly be applied in a given state, is substantially larger than the branching factor in the reverse graph, which is the number of operators that could bring the system into a given state.[25] This is especially true when goal sets are represented by a small set of propositional literals (Section 5). The two approaches are not mutually exclusive, however: one can mix forward and backward expansions of the underlying problem graph and terminate when a forward path and backward path meet.

The important thing to observe about these algorithms is that they restrict their attention to the *relevant* and *reachable* states. In forward search, only those states that can be reached from $s_{init}$ are ever considered: this can provide benefit over dynamic programming methods if few states are reachable, since unreachable states cannot play a role in constructing a successful plan. In backward approaches, similarly, only states lying on some path to the goal region $G$ are considered, and this can have significant advantages over dynamic programming if only a fraction of the state space is connected to the goal region.

In contrast, dynamic programming methods (with the exception of asynchronous methods) must examine the entire state space at every iteration. Of course, the ability to ignore parts of the state space comes from planning's stringent definition of what is relevant: states in $G$ have positive reward, no other states matter except to the extent they move the agent closer to the goal, and the choice of action at states unreachable from $s_{init}$ is not of interest.

While state-based search techniques use knowledge of a specific initial state and a specific goal set to constrain the search process, forward search does not exploit knowledge of the goal set, nor does backward search exploit knowledge of the initial state. The GraphPlan algorithm (Blum & Furst, 1995) can be viewed as a planning method that integrates the propagation of forward reachability constraints with backward goal-informed search. We describe this approach in Section 5. Furthermore, work on *partial order planning* (POP) can be viewed as a slightly different approach to this form of search. It too is described in Section 5, after we discuss feature-based or *intensional* representations for MDPs and planning problems.

### 3.2.2 Decision Trees and Real-time Dynamic Programming

State-based search techniques are not limited to deterministic, goal-oriented domains. Knowledge of the initial state can be exploited in more general MDPs as well, forming the basis of *decision tree search algorithms*. Assume we have been given a finite-horizon FOMDP with horizon $T$ and initial state $s_{init}$. A *decision tree rooted at* $s_{init}$ is constructed in much the same way as a search tree for a deterministic planning problem (French, 1986). Each action applicable at $s_{init}$ forms level 1 of the tree. The states $s'$ that result with positive probability when any of those actions occur are applied at $s_{init}$ are placed at level 2, with an arc

25. See Bacchus et al. (1995, 1998) for some recent work that makes the case for progression with good search control, and Bonet et al. (1997) who argue that progression in deterministic planning is useful when integrating planning and execution.

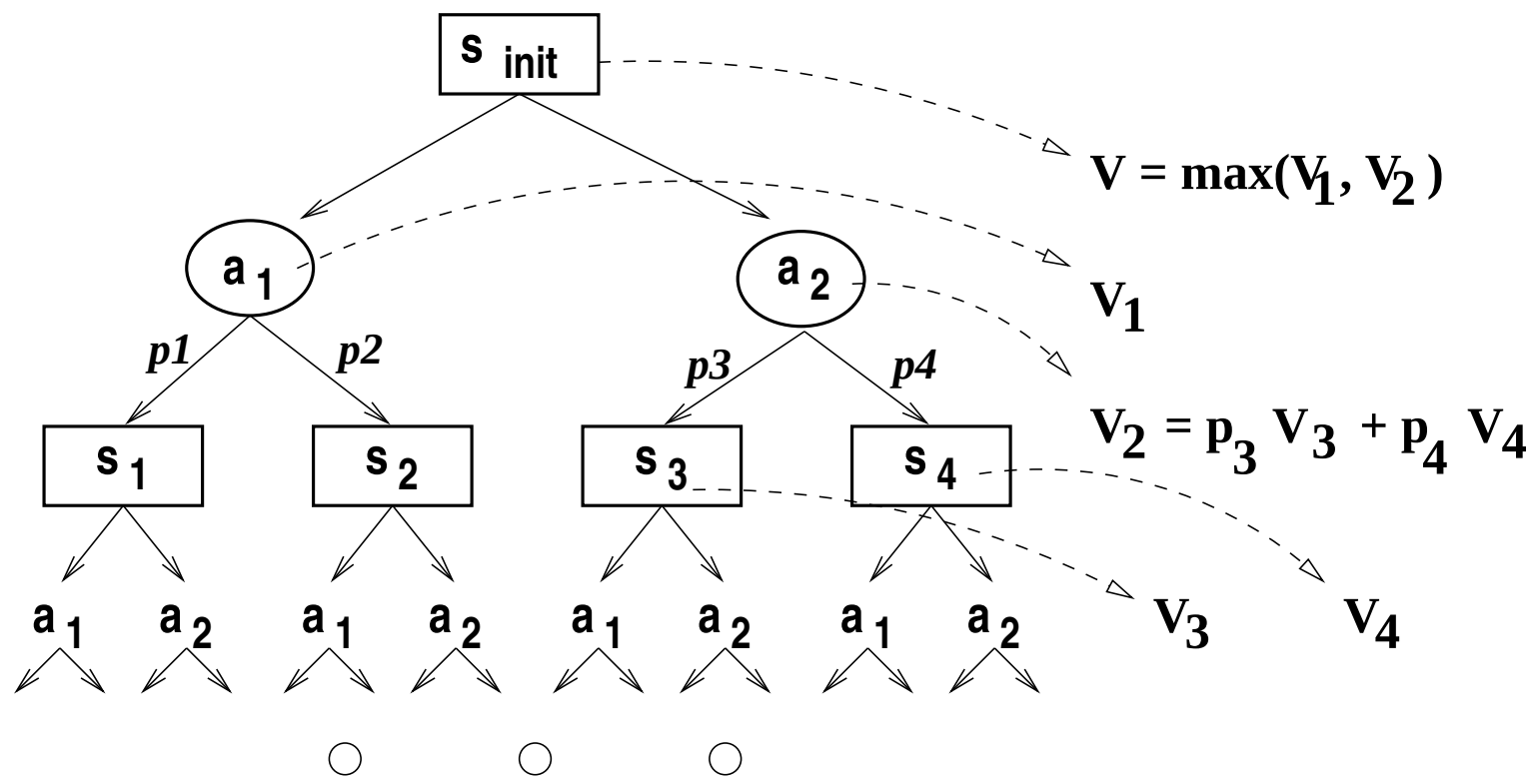

Figure 12: The initial stages of a decision tree for evaluating action choices at $s_{init}$. The value of an action is the expected value of its successor states, while the value of a state is the maximum of the values of its successor actions (as indicated by dashed arrows at selected nodes).

labeled with probability $\Pr(s'|a, s_{init})$ relating $s'$ with $a$. Level 3 has the actions applicable at the states at level 2, and so on, until the tree is grown to depth $2T$, at which point each branch of the tree is a path consisting of a positive-probability length-$T$ trajectory rooted at $s_{init}$ (see Figure 12).

The *relevant* part of the optimal $T$-stage value function and the optimal policy can easily be computed using this tree. We say that value of any node in the tree labeled with an action is the *expected value* of its successor states in the tree (using the probabilities labeling the arcs), while the value of any node in the tree labeled with state $s$ is the sum of $R(s)$ and the *maximum value* of its successor actions.[26] The *rollback procedure*, whereby value at the leaves of the tree are first computed and then values at successively higher levels of the tree are determined using the preceding values, is, in fact, a form of value iteration. The value of any state $s$ at level $2t$ is precisely $V^*_{T-t}(s)$ and the maximizing actions form the optimal finite-horizon policy. This form of value iteration is *directed*: $(T - t)$-stage-to-go values are computed only for states that are reachable from $s_{init}$ within $t$ steps. Infinite-horizon problems can be solved in an analogous fashion if one can determine *a priori* the depth required (i.e., the number of iterations of value iteration needed) to ensure convergence to an optimal policy.

Unfortunately, the branching factor for stochastic problems is generally much greater than that for deterministic problems. More troublesome still is the fact that one must construct the entire decision tree to be sure that the proper values are computed, and hence the optimal policy constructed. This stands in contrast with classical planning search, where attention can be focused on a single branch: if a goal state is reached, the path constructed determines a satisfactory plan. While worst-case behavior for planning *may* require searching the whole tree, decision-tree evaluation is especially problematic because

26. States at level $2T$ are given value $R(s)$.

the entire tree *must* be generated in general to ensure optimal behavior. Furthermore, infinite-horizon problems pose the difficulty of determining a sufficiently deep tree.

One way around this difficulty is the use of *real time search* (Korf, 1990). In particular, *real-time dynamic programming*, or RTDP, has been proposed in (Barto et al., 1995) as a way of approximately solving large MDPs in an online fashion. One can interleave search with execution of an approximately optimal policy using a form of RTDP similar to decision-tree evaluation as follows. Imagine the agent finds itself in a particular state $s_{init}$. It can then build a partial search tree to some depth, perhaps uniformly or perhaps with some branches expanded more deeply than others. Partial tree construction may be halted due to time pressure or due to an assessment by the agent that further expansion of the tree may not be fruitful. When a decision to act must be made, the rollback procedure is applied to this partial, possibly unevenly expanded decision tree.

Reward values can be used to evaluate the leaves of the tree, but this may offer an inaccurate picture of the value of nodes higher in the tree. Heuristic information can be used to estimate the long-term value of states labeling leaves. As with value iteration, the deeper the tree, the more accurate the estimated value at the root (generally speaking) for a fixed heuristic. We will see in Section 5 that structured representations of MDPs can provide a means to construct such heuristics (Dearden & Boutilier, 1994, 1997). Specifically, with admissible heuristics or upper and lower bounds on the true values of leaf nodes in the tree, methods such as A* or branch-and-bound search can be used.

A key advantage of integrating search with execution is that the actual outcome of the action taken can be used to prune from the tree the branches rooted at the unrealized outcomes. The subtree rooted at the realized state can then be expanded further to make the next action choice. The algorithm of Hansen and Zilberstein (1998) can be viewed as a variant of these methods in which stationary policies (i.e., state-action mappings) can be extracted during the search process.

RTDP is formulated by Barto et al. (1995) more generally as a form of online, asynchronous value iteration. Specifically, the values "rolled backed" can be cached and used as improved heuristic estimates of the value function at the states in question. This technique is also investigated in (Bonet et al., 1997; Dearden & Boutilier, 1994, 1997; Koenig & Simmons, 1995), and is closely tied to Korf's (1990) LRTA* algorithm. These value updates also need not proceed strictly using a decision tree to determine the states; the key requirement of RTDP is simply that the actual state $s_{init}$ be one of the states whose value is updated at each decision-action iteration.

A second way to avoid some of the computational difficulties that arise in large search spaces is to use *sampling methods*. These methods sample from the space of possible trajectories and use this sampled information to provide estimates of the values of specific courses of action. This approach is quite common in reinforcement learning (Sutton & Barto, 1998), where simulation models are often used to generate experience from which a value function can be learned. In the present context, Kearns, Mansour and Ng (Kearns, Mansour, & Ng, 1999) have investigated search methods for infinite-horizon MDPs where the successor states of any specific action are randomly sampled according to the transition distribution. Thus, rather than expand all successor states, only sampled states are searched. Though this method is exponential in the "effective" horizon (or mixing rate) of the MDP and is required to expand all actions, the number of states expanded can be less than that required by full search, even if the underlying transition graph is not sparse. They are able to provide polynomial bounds (ignoring action branching and horizon effects) on the number of trajectories that need to be sampled in order to generate approximately optimal behavior with high probability.

### 3.3 Summary

We have seen that dynamic programming methods and state-based search methods can both be used for fully observable and non-observable MDPs, with forward search methods interpretable as "directed" forms of value iteration. Dynamic programming algorithms generally require explicit enumeration of the state space at each iteration, while search techniques enumerate only reachable states; but the branching factor may require that, at sufficient depth in the search tree, search methods enumerate individual states multiple times, whereas they are considered only once per stage in dynamic programming. Overcoming this difficulty in search requires the use of cycle-checking and multiple-path-checking methods.

We note that search techniques can be applied to partially observable problems as well. One way to do this is to search through the space of belief states (just as dynamic programming can be applied to the belief space MDP—see Section 2.10.2). Specifically, belief states play the role of system states and the stochastic effects of actions on belief states are induced by specific observation probabilities, since each observation has a distinct, but fixed effect on any belief state. This type of approach has been pursued in (Bonet & Geffner, 1998; Koenig & Simmons, 1995).

## 4. Factored Representations

To this point our discussion of MDPs has used an explicit or *extensional* representation for the set of states (and actions) in which states are enumerated directly. In many cases it is advantageous, from both the representational and computational point of view, to talk about *properties* of states or sets of states: the set of possible initial states, the set of states in which action $a$ can be executed, and so on. It is generally more convenient and compact to *describe* sets of states based on certain properties or features than to enumerate them explicitly. Representations in which descriptions of objects substitute for the objects themselves are called *intensional* and are the technique of choice in AI systems.

An intensional representation for planning systems is often built by defining a set of *features* that are sufficient to describe the state of the dynamic system of interest. In the example in Figure 2, the state was described by a set of six features: the robot's location, the lab's tidiness, whether or not mail is present, whether or not the robot has mail, whether or not there is a pending coffee request, and whether or not the robot has coffee. The first and second features can each take one of five values, and the last four can each take one of two values (true or false). An assignment of values to the six features completely defines a state; the state space thus comprises all possible combinations of feature values, with $|\mathcal{S}| = 400$. Each feature, or *factor*, is typically assigned a unique symbolic name, as indicated in the second column of Figure 2. The fundamental tradeoff between extensional and intensional representations becomes clear in this example. An extensional representation of the coffee example views the space of possible states as a single variable that takes on 400 possible values, whereas the intensional or *factored* representation views a state as the cross product of six variables, each of which takes on substantially fewer values. Generally, the state space grows exponentially in the number of features required to describe a system.

The fact that the state of a system can be described using a set of features allows one to adopt *factored representations* of actions, rewards and other components of an MDP. In a factored action representation, for instance, one generally describes the effect of an action on specific state features rather than on entire states. This often provides considerable representational economy. For instance, in the Strips action representation (Fikes & Nilsson, 1971), the state transitions induced by actions are represented implicitly by describing the effects of actions on only those features that *change* value when the action is executed. Factored representations can be very compact when individual actions affect relatively few features, or when their effects exhibit certain regularities. Similar remarks apply to the representation of reward functions, observation models, and so on. The regularities that make factored representations suitable for many planning problems can often be exploited by planning and decision-making algorithms.

While factored representations have long been used in classical AI planning, similar representations have also been adopted in the recent use of MDP models within AI. In this section (Section 4), we focus on the economy of representation afforded by exploiting the structure inherent in many planning domains. In the following section (Section 5), we describe how this structure—when made explicit by the factored representations—can be exploited computationally in plan and policy construction.

### 4.1 Factored State Spaces and Markov Chains

We begin by examining *structured states*, or systems whose state can be described using a finite set of *state variables* whose values change over time.[27] To simplify our illustration of the potential space savings, we assume that these state variables are boolean. If there are $M$ such variables, then the size of the state space is $|\mathcal{S}| = N = 2^M$. For large $M$, specifying or representing the dynamics explicitly using state-transition diagrams or $N \times N$ matrices is impractical. Furthermore, representing a reward function as an $N$-vector, and specifying the observational probabilities, is similarly infeasible. In Section 4.2, we define a class of problems in which the dynamics can be represented in $O(M)$ space in many cases. We begin by considering how to represent Markov chains compactly and then consider incorporating actions, observations and rewards.

We let a state variable $X$ take on a finite number of values and let $\Omega_X$ stand for the set of possible values. We assume that $\Omega_X$ is finite, though much of what follows can be applied to countable state and action spaces as well. We say the state space is *flat* if it is specified using one state variable (this variable is denoted $S$ as in the general model, taking values from $\mathcal{S}$). The state space is *factored* if there is more than one state variable. A state is any possible assignment of values to these variables. Letting $X_i$ represent the $i$th state variable, the state space is the cross product of the value spaces for the individual state variables; that is, $\mathcal{S} = \times_{i=1}^{M} \Omega_{X_i}$. Just as $S^t$ denotes the state of the process at stage $t$, we let $X_i^t$ be the random variable representing the value of the $i$th state variable at stage $t$.

27. These variables are often called *fluents* in the AI literature (McCarthy & Hayes, 1969). In classical planning, these are the atomic propositions used to describe the domain.

A *Bayesian network* (Pearl, 1988) is a representational framework for compactly representing a probability distribution in factored form. Although these networks have most typically been used to represent atemporal problem domains, we can apply the same techniques to represent Markov chains, encoding the chain's transition probabilities in the network structure (Dean & Kanazawa, 1989).

Formally, a Bayes net is a directed acyclic graph in which vertices correspond to random variables and an edge between two variables indicates a direct probabilistic dependency between them. A network so constructed also reflects implicit *independencies* among the variables. The network must be *quantified* by specifying a probability for each variable (vertex) conditioned on all possible values of its immediate parents in the graph. In addition, the network must include a marginal distribution: an unconditional probability for each vertex that has no parents. This quantification is captured by associating a *conditional probability table (CPT)* with each variable in the network. Together with the independence assumptions defined by the graph, this quantification defines a unique joint distribution over the variables in the network. The probability of any event over this space can then be computed using algorithms that exploit the independencies represented within the graphical structure. We refer to Pearl (1988) for details.

Figures 3(a)-(c) (page 7) are special cases of Bayes nets called "temporal" Bayesian networks. In these networks, vertices in the graph represent the system's state at different time points and arcs represent dependencies across time points. In these temporal networks, each vertex's parent is its temporal predecessor, the conditional distributions are transition probability distributions, and the marginal distributions are distributions over initial states.

The networks in Figure 3 reflect an extensional representation scheme in which states are explicitly enumerated, but techniques for building and performing inference in probabilistic temporal networks are designed especially for application to factored representations. Figure 13 illustrates a *two-stage temporal Bayes net* (2TBN) describing the state-transition probabilities associated with the Markov chain induced by the fixed policy of executing the action *CClk* (repeatedly moving counterclockwise). In a 2TBN, the set of variables is partitioned into those corresponding to state variables at a given time (or stage) $t$ and those corresponding to state variables at time $t+1$. Directed arcs indicate probabilistic dependencies between those variables in the Markov chain. *Diachronic arcs* are those directed from time $t$ variables to time $t+1$ variables, while *synchronic arcs* are directed between variables at time $t+1$. Figure 13 contains only diachronic arcs; synchronic arcs will be discussed later in this section.

Given any state at time $t$, the network induces a unique distribution over states at $t+1$. The quantification of the network describes how the state of any particular variable changes as a function of certain state variables. The lack of a direct arc (or more generally a directed path if there are synchronic arcs among the $t+1$ variables) from a variable $X_t$ to another variable $Y_{t+1}$ means that knowledge of $X_t$ is irrelevant to the prediction of the (immediate, or one-stage) evolution of variable $Y$ in the Markov process.

Figure 13 shows how compact this representation can be in the best of circumstances, as many of the potential links between one stage and the next can be omitted. The graphical representation makes explicit the fact that the policy (i.e., the action *CClk*) can affect only the state variable *Loc*, and the exogenous events *ArrM*, *ReqC*, and *Mess* can affect only the variables $M$, *CR*, and *Tidy*, respectively.[28] Furthermore, the dynamics of *Loc* (and the other variables) can be described using only knowledge of the state of their parent variables; for instance, the distribution over *Loc* at $t+1$ depends only on the value of *Loc* at the previous stage (e.g., if $Loc^t = O$, then $Loc^{t+1} = M$ with probability 0.9 and $Loc^{t+1} = O$ with probability 0.1). Similarly, *CR* can become true with probability 0.2 (due to a *ReqC* event), but once true, cannot become false (under this simple policy); and *RHC* remains true (or false) with certainty if it was true (or false) at the previous stage. Finally, the effects on the relevant variables are independent. For any instantiation of the variables at time $t$, the distribution over next states can be computed by multiplying the conditional probabilities of relevant $t+1$ variables.

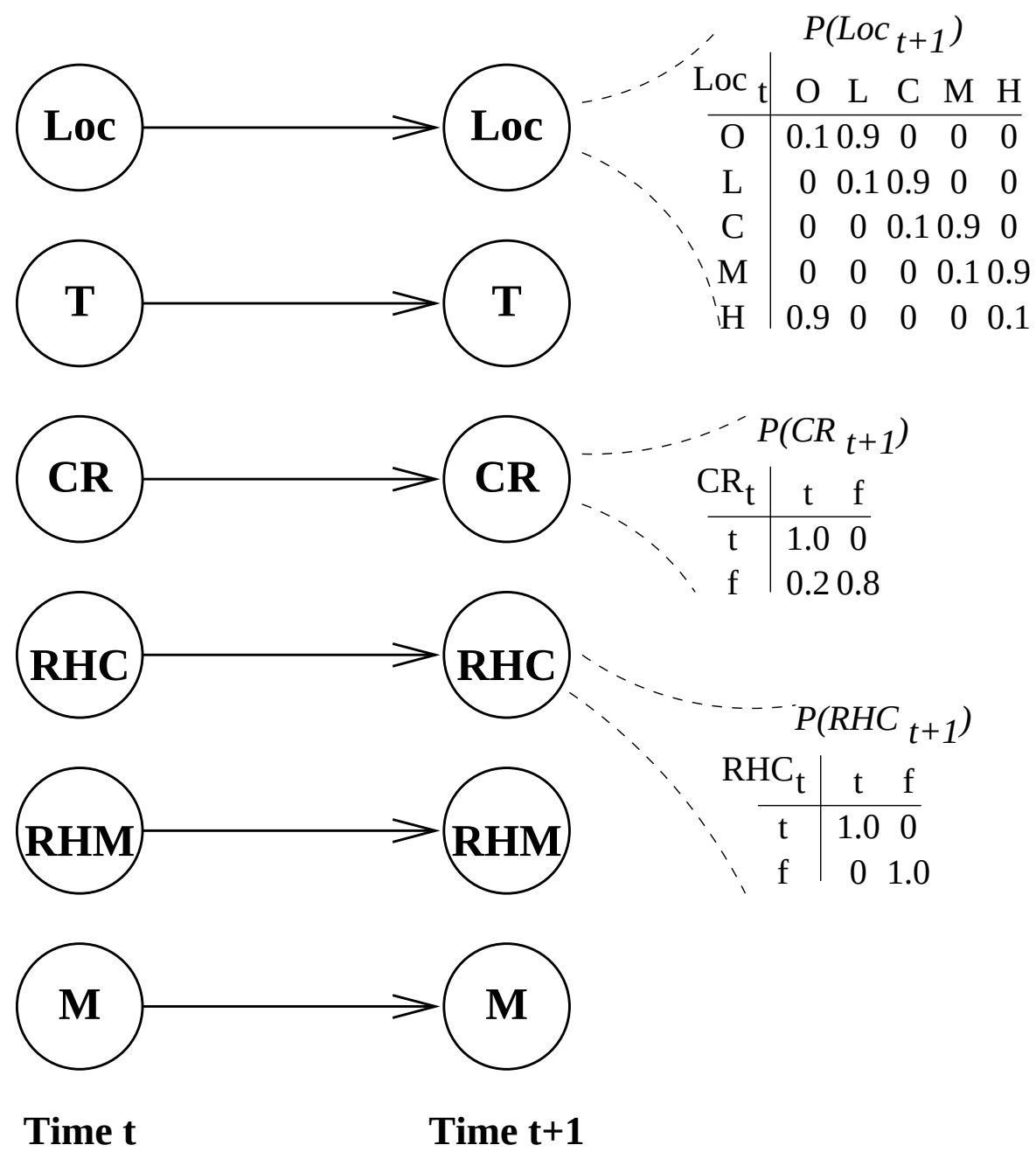

Figure 13: A factored 2TBN for the Markov chain induced by moving counterclockwise (with selected CPTs shown).

The ability to omit arcs from the graph based on the locality and independence of action effects has a strong effect on the number of parameters that must be supplied to complete the model. Although the full transition matrix for *CClk* would be of size $400^2 = 160000$, the transition model in Figure 13 requires only 66 parameters.[29]

The example above shows how 2TBNs exploit independence to represent Markov chains compactly, but the example is extreme in that there is effectively no relationship between the variables—the chain can be viewed as the product of six independently evolving processes.

28. We show only some of the CPTs for brevity.

29. In fact, we can exploit the fact that probabilities sum to one and leave one entry unspecified per row of any CPT or explicit transition matrix. In this case, the 2TBN requires only 48 explicit parameters, while the transition matrix requires $400 \cdot 300 = 159,600$ entries. We generally ignore this fact when comparing the sizes of representations.

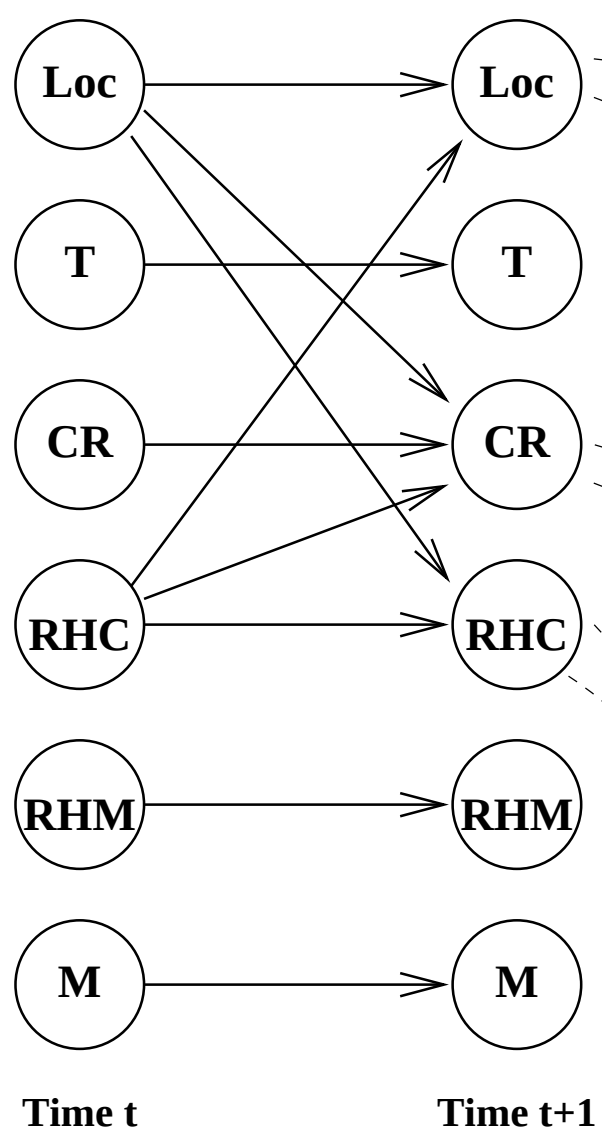

*P(Loc t+1)*

| $Loc_t$ | $RHC_t$ | O | L | C | M | H |
|---|---|---|---|---|---|---|
| O | t | 1.0 | 0 | 0 | 0 | 0 |
| L | t | 0 | 0.1 | 0.9 | 0 | 0 |
| C | t | 0 | 0 | 0.1 | 0.9 | 0 |
| M | t | 0 | 0 | 0 | 0.1 | 0.9 |
| H | t | 0.9 | 0 | 0 | 0 | 0.1 |
| O | f | 0.1 | 0.9 | 0 | 0 | 0 |
| L | f | 0 | 0.1 | 0.9 | 0 | 0 |
| C | f | 0 | 0 | 0.1 | 0.9 | 0 |
| M | f | 0 | 0 | 0 | 0.1 | 0.9 |
| H | f | 0.9 | 0 | 0 | 0 | 0.1 |

*P(CR t+1)*

| $Loc_t$ | $RHC_t$ | $CR_t$ | t | f |
|---|---|---|---|---|
| O | t | t | .05 | .95 |
| O | t | f | 0.2 | 0.8 |
| O | f | t | 1.0 | 0.0 |
| O | f | f | 0.2 | 0.8 |
| L | t | t | 1.0 | 0.0 |
| L | t | f | 0.2 | 0.8 |
| L | f | t | 1.0 | 0.0 |
| L | f | f | 0.2 | 0.8 |
| etc. | | | etc. | |

*P(RHC t+1)*

| $Loc_t$ | $RHC_t$ | t | f |
|---|---|---|---|
| O | t | 0.0 | 1.0 |
| O | f | 0.0 | 1.0 |
| L | t | 1.0 | 0.0 |
| L | f | 0.0 | 1.0 |
| C | t | 1.0 | 0.0 |
| C | f | 0.0 | 1.0 |
| etc. | | etc. | |

Figure 14: A 2TBN for the Markov chain induced by moving counterclockwise and delivering coffee.

In general, these "subprocesses" will interact, but still exhibit certain independencies and regularities that can be exploited by a 2TBN representation. We consider two distinct Markov chains that exhibit different types of dependencies.

Figure 14 illustrates a 2TBN representing the Markov chain induced by the following policy: the robot consistently moves counterclockwise unless it is in the office and has coffee, in which case it delivers coffee to the user. Notice that different variables are now dependent: for instance, predicting the value of $RHC$ at $t+1$ requires knowing the values of $Loc$ and $RHC$ at $t$. The CPT for $RHC$ shows that the robot retains coffee at stage $t+1$ with certainty, if it has it at stage $t$, in all locations except $O$ (where it executes $DelC$, thus losing the coffee). The variable $Loc$ also depends on the value of $RHC$. The location will change as in Figure 13 with one exception: if the robot is in the office with coffee, the location remains the same (since the robot does not move, but executes $DelC$). The effect on the variable $CR$ is explained as follows: if the robot is in the office and delivers coffee in its possession, it will fulfill any outstanding coffee request. However, the 0.05 chance of $CR$ remaining true under these conditions indicates a 5% chance of spilling the coffee.

Even though there are more dependencies (i.e., additional diachronic arcs) in this 2TBN, it still requires only 118 parameters. Again, the distribution over resulting states is determined by multiplying the conditional distributions for the individual variables. Even though the variables are "related," when the state $S^t$ is known, the variables at time $t+1$ ($Loc^{t+1}$, $RHC^{t+1}$, etc.) are independent. In other words,

$$\begin{aligned}\Pr(&Loc^{t+1}, T^{t+1}, CR^{t+1}, RHC^{t+1}, RHM^{t+1}, M^{t+1} | S^t) = \\ &\Pr(Loc^{t+1}|S^t)\Pr(T^{t+1}|S^t)\Pr(CR^{t+1}|S^t)\Pr(RHC^{t+1}|S^t)\Pr(RHM^{t+1}|S^t)\Pr(M^{t+1}|S^t)\end{aligned} \quad (8)$$

Figure 15 illustrates a 2TBN representing the Markov chain induced by the same policy as above, but where we assume that the act of moving counterclockwise has a slightly different effect. We now suppose that, when the robot moves from the hallway into some adjacent location, it has a 0.3 chance of spilling any coffee it has in its possession: the fragment of the CPT for $RHC$ in Figure 15 illustrates this possibility. Furthermore, should the robot be carrying mail whenever it loses coffee (whether "accidentally" or "intentionally" via the $DelC$ action), there is a 0.5 chance it will lose the mail. Notice that the effects of this policy on the variables $RHC$ and $RHM$ are *correlated*: one cannot accurately predict the probability of $RHM^{t+1}$ without determining the probability of $RHC^{t+1}$. This correlation is modeled by the synchronic arc between $RHC$ and $RHM$ at the $t+1$ slice of the network.

The independence of the $t+1$ variables given $S^t$ does not hold in 2TBNs with synchronic arcs. Determining the probability of a resulting state requires some simple probabilistic reasoning, for example, application of the chain rule. In this example, we can write

$$\Pr(RHC^{t+1}, RHM^{t+1}|S^t) = \Pr(RHM^{t+1}|RHC^{t+1}, S^t)\Pr(RHC^{t+1}|S^t)$$

The joint distribution over $t+1$ variables given $S^t$ can then be computed as in Equation 8 above, with this term replacing the $\Pr(RHC^{t+1}|S^t)\Pr(RHM^{t+1}|S^t)$—while these two variables are correlated, the remaining variables are independent.

We refer to 2TBNs with no synchronic arcs, like the one in Figure 14, as *simple 2TBNs*. *General 2TBNs* allow synchronic as well as diachronic arcs, as in Figure 15.

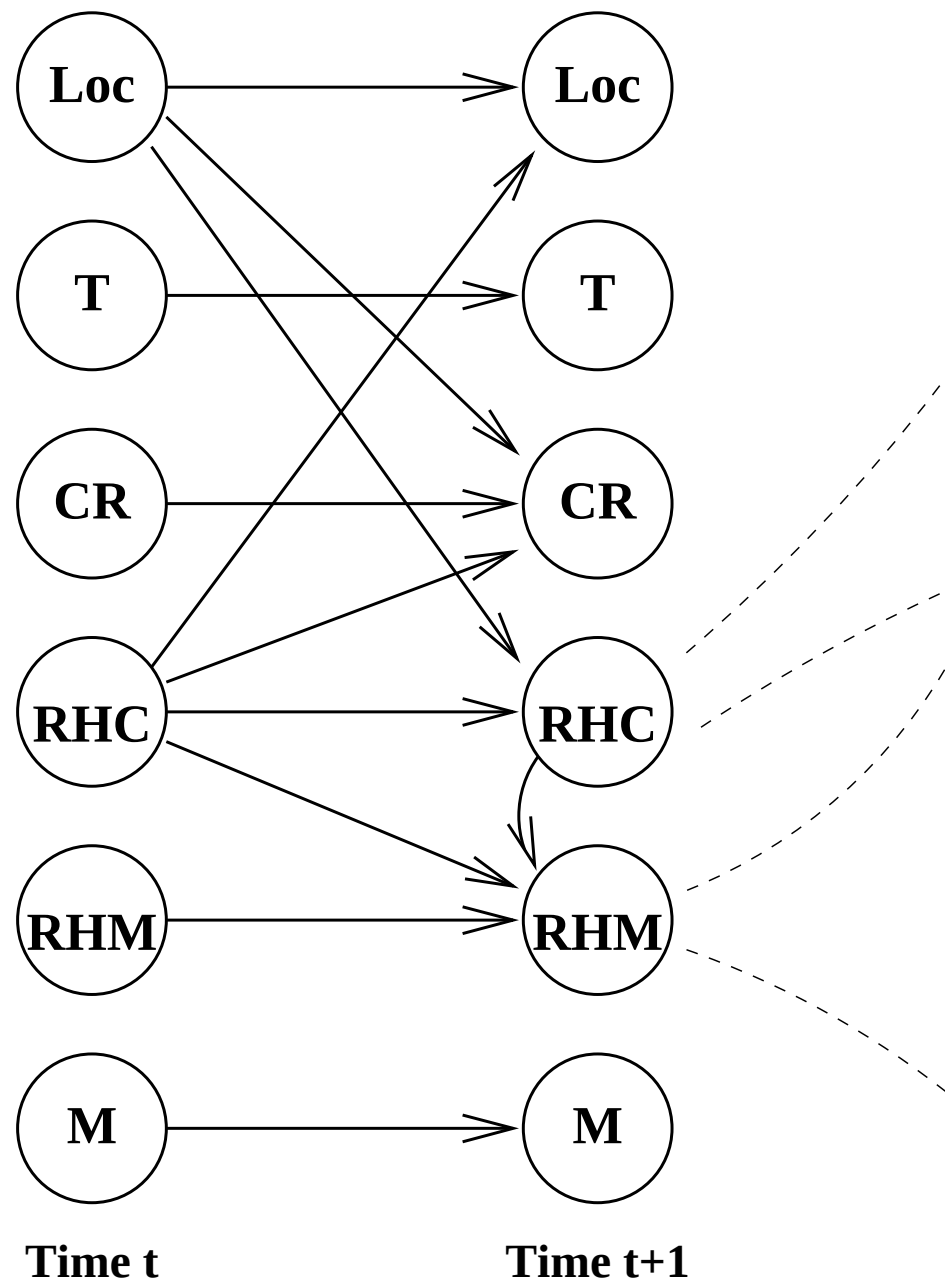

*Pr(RHC_{t+1})*

| $Loc_t$ | $RHC_t$ | t | f |
|---|---|---|---|
| O | t | 1.0 | 1.0 |
| O | f | 0.0 | 1.0 |
| H | t | 0.7 | 0.3 |
| H | f | 0.0 | 1.0 |
| C | t | 1.0 | 0.0 |
| C | f | 0.0 | 1.0 |
| etc. | | etc. | |

*Pr(RHM_{t+1})*

| $RHC_t$ | $RHC_{t+1}$ | $RHM_t$ | t | f |
|---|---|---|---|---|
| t | t | t | 1.0 | 0.0 |
| t | t | f | 0.0 | 1.0 |
| t | f | t | 0.5 | 0.5 |
| t | f | f | 0.0 | 1.0 |
| f | t | t | 1.0 | 0.0 |
| f | t | f | 0.0 | 1.0 |
| f | f | t | 1.0 | 0.0 |
| f | f | f | 0.0 | 1.0 |

Figure 15: A 2TBN for the Markov chain induced by moving counterclockwise and delivering coffee with correlated effects.

### 4.2 Factored Action Representations

Just as we extended Markov chains to account for different actions, we must extend the 2TBN representation to account for the fact that the state transitions are influenced by the agent's choice of action. We discuss a variety of techniques for specifying the transition matrices that exploit the factored state representation to produce representations that are more natural and compact than explicit transition matrices.

#### 4.2.1 Implicit-Event Models

We begin with the implicit-event model from Section 2.3 in which the effects of actions and exogenous events are combined in a single transition matrix. We will consider explicit-event models in Section 4.2.4. As we saw in the previous section, algorithms such as value and policy iteration require the use of transition models that reflect the *ultimate* transition probabilities, including the effects of any exogenous events.

One way to model the dynamics of a fully observable MDP is to represent each action by a separate 2TBN. The 2TBN shown above in Figure 13 can be seen as a representation of the action *CClk* (since the policy inducing the Markov chain in that example consists of the repeated application of that action alone). The network fragment in Figure 16(a) illustrates the interesting aspects of the 2TBN for the *DelC* action including the effects of exogenous events. As above, the robot satisfies an outstanding coffee request if it delivers coffee while it is in the office and has coffee (with a 0.05 chance of spillage), as shown in the conditional probability table for *CR*. The effect on *RHC* can be explained as follows: the robot loses the coffee (to the user or to spillage) if it delivers it in the office; if it attempts delivery elsewhere, there is a 0.7 chance that a random passerby will take the coffee from the robot.

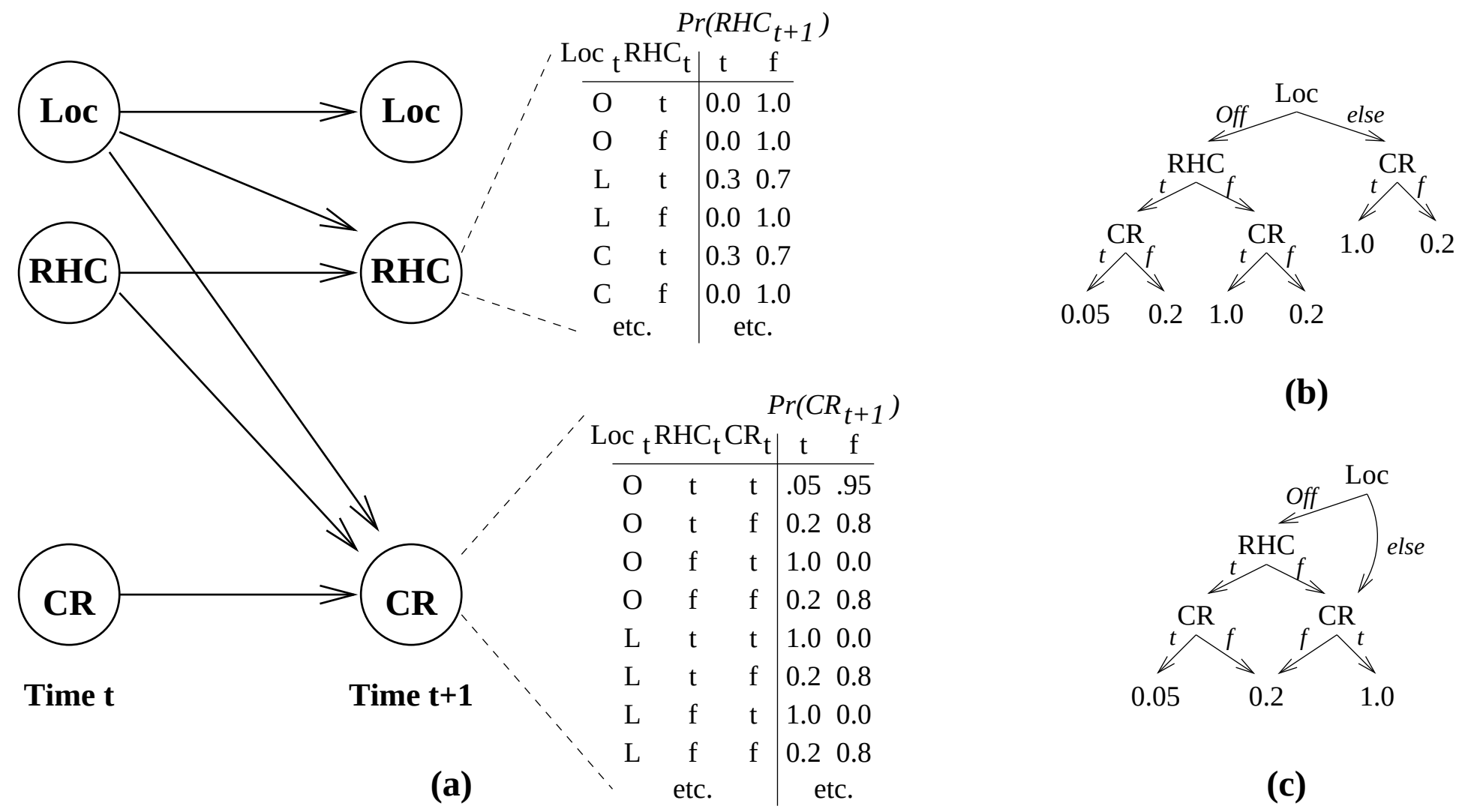

Figure 16: A factored 2TBN for action *DelC* (a) and structured CPT representations (b,c).

As in the case of Markov chains, the effects of actions on different variables can be correlated, in which case we must introduce synchronic arcs. Such correlations can be thought of as *ramifications* (Baker, 1991; Finger, 1986; Lin & Reiter, 1994).

#### 4.2.2 Structured CPTs

The conditional probability table (CPT) for the node *CR* in Figure 16(a) has 20 rows, one for each assignment to its parents. However, the CPT contains a number of regularities. Intuitively, this reflects the fact that the coffee request will be met successfully (i.e., the variable becomes false) 95% of the time when *DelC* is executed, *if* the robot has coffee *and* is in the right location (the user's office). Otherwise, *CR* remains true if it was true and becomes true with probability 0.2 if it was not. In other words, there are three distinct cases to be considered, corresponding to three "rules" governing the (stochastic) effect of *DelC* on *CR*. This can be represented more compactly by using a decision tree representation (with "else" branches to summarize groups of cases involving multivalued variables such as *Loc*) like that shown in Figure 16(b), or more compactly still using a decision graph (Figure 16(c)). In tree- and graph-based representations of CPTs, interior nodes are labeled by parent variables, edges by values of the variables, and leaves or terminals by distributions over the child variable's values.[30]

Decision-tree and decision-graph representations are used to represent actions in fully observable MDPs in (Boutilier et al., 1995; Hoey, St-Aubin, Hu, & Boutilier, 1999) and

30. When the child is boolean, we label the leaves with only the probability of that variable being true (the probability of the variable being false is one minus this value).

are described in detail in (Poole, 1995; Boutilier & Goldszmidt, 1996).[31] Intuitively, trees and graphs embody the rule-like structure present in the family of conditional distributions represented by the CPT, and in the settings we consider often yield considerable representational compactness. Rule-based representations have been used directly by Poole (1995, 1997a) in the context of decision processes and can often be more compact than trees (Poole, 1997b). We generically refer to representations of this type as 2TBNs with *structured CPTs*.

#### 4.2.3 Probabilistic STRIPS Operators

The 2TBN representation can be viewed as oriented toward describing the effects of actions on distinct variables. The CPT for each variable expresses how it (stochastically) changes (or persists), perhaps as a function of the state of certain other variables. However, it has long been noted in AI research on planning and reasoning about action that most actions change the state in limited ways; that is, they affect a relatively small number of variables. One difficulty with *variable-oriented* representations such as 2TBNs is that one must explicitly assert that variables unaffected by a specific action persist in value (e.g., see the CPT for *RHC* in Figure 13)—this is an instance of the infamous *frame problem* (McCarthy & Hayes, 1969).

Another form of representation for actions might be called an *outcome-oriented* representation: one explicitly describes the possible *outcomes* of an action or the possible joint effects over all variables. This was the idea underlying the Strips representation from classical planning (Fikes & Nilsson, 1971).

A classical Strips operator is described by a *precondition* and a set of *effects*. The former identifies the set of states in which the action can be executed, and the latter describes how the input state changes as a result of taking the action. A probabilistic Strips operator (PSO) (Hanks, 1990; Hanks & McDermott, 1994; Kushmerick et al., 1995) extends the Strips representation in two ways. First, it allows actions to have different effects depending on context, and second, it recognizes that the effects of actions are not always known with certainty.[32]

Formally, a PSO consists of a set of mutually exclusive and exhaustive logical formulae, called *contexts*, and a *stochastic effect* associated with each context. Intuitively, a context discriminates situations under which an action can have differing stochastic effects. A stochastic effect is itself a set of *change sets*—a simple list of variable values—with a probability attached to each change set, with the requirement that these probabilities sum to one. The semantics of a stochastic effect can be described as follows: when the stochastic effect of an action is applied at state $s$, the possible resulting states are determined by the change sets, each occurring with the corresponding probability; the resulting state associated with a change set is constructed by changing variable values at state $s$ to match those in the change set, while all unmentioned variables persist in value. Note that since only one

31. The fact that certain direct dependencies among variables in a Bayes net are rendered irrelevant under specific variable assignments has been studied more generally in the guise of *context-specific independence* (Boutilier, Friedman, Goldszmidt, & Koller, 1996); see (Geiger & Heckerman, 1991; Shimony, 1993) for related notions.

32. The *conditional* nature of effects is also a feature of a deterministic extension of Strips known as ADL (Pednault, 1989).

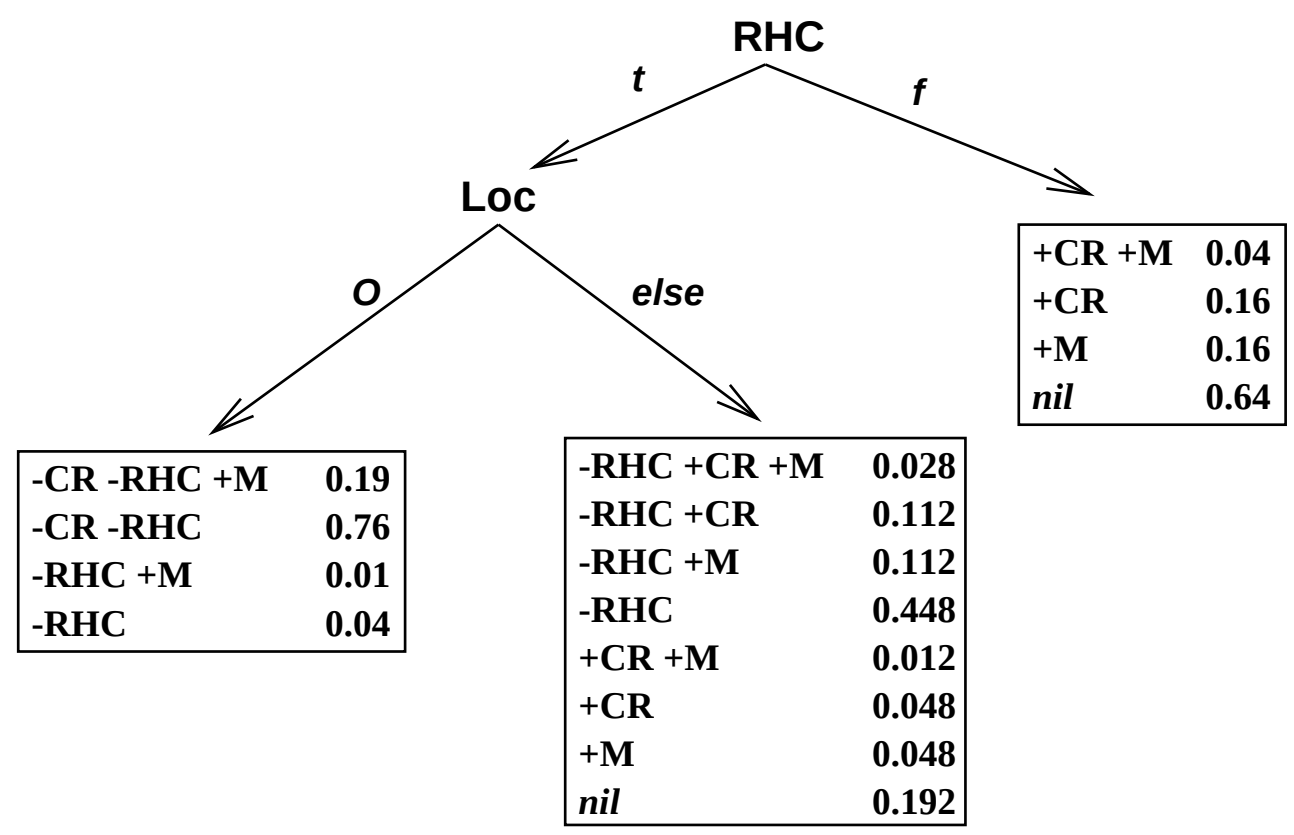

Figure 17: A PSO representation for the *DelC* action.

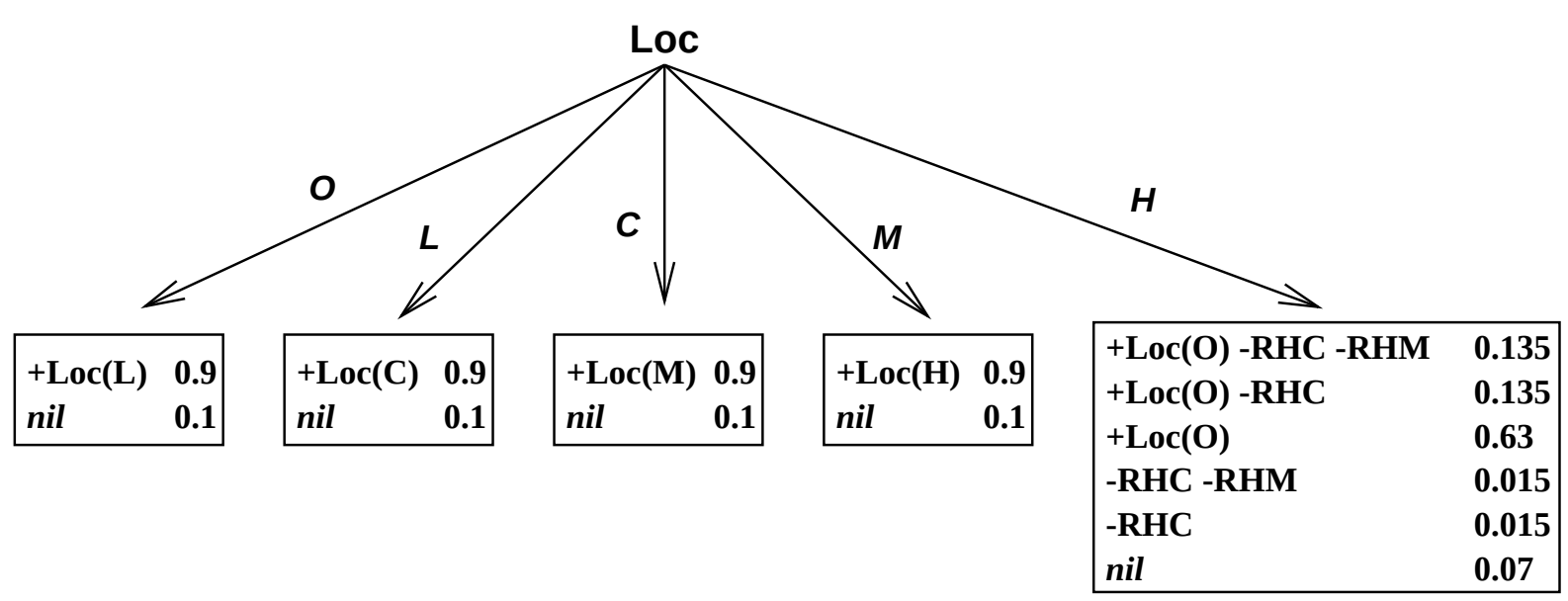

Figure 18: A PSO representation of a simplified *CClk* action.

context can hold in any state $s$, the transition distribution for the action at any state $s$ is easily determined.

Figure 17 gives a graphical depiction of the PSO for the *DelC* action (shown as a 2TBN in Figure 16). The three contexts $\neg RHC$, $RHC \wedge Loc(O)$ and $RHC \wedge \neg Loc(O)$ are represented using a decision tree. At the leaf of each branch in the decision tree is the stochastic effect (set of change sets and associated probabilities) determined by the corresponding context. For example, when $RHC \wedge Loc(O)$ holds, the action has four possible effects: the robot loses the coffee; it may or may not satisfy the coffee request (due to the 0.05 chance of spillage); and mail may or may not arrive. Notice that each outcome is spelled out completely. The number of outcomes in the other two contexts is rather large due to possible exogenous events (we discuss this further in Section 4.2.4).[33]

A key difference between PSOs and 2TBNs lies in their treatment of persistence. All variables that are unaffected by an action must be given CPTs in the 2TBN model, while such variables are not mentioned at all in the PSO model (e.g., compare the variable *Loc* in both representations of *DelC*). In this way, PSOs can be said to "solve" the frame problem, since unaffected variables need not be mentioned in an action's description.[34]

33. To keep Figure 17 manageable, we ignore the effect of the exogenous event *Mess* on variable $T$.
34. For a discussion of the frame problem in 2TBNs, see (Boutilier & Goldszmidt, 1996).

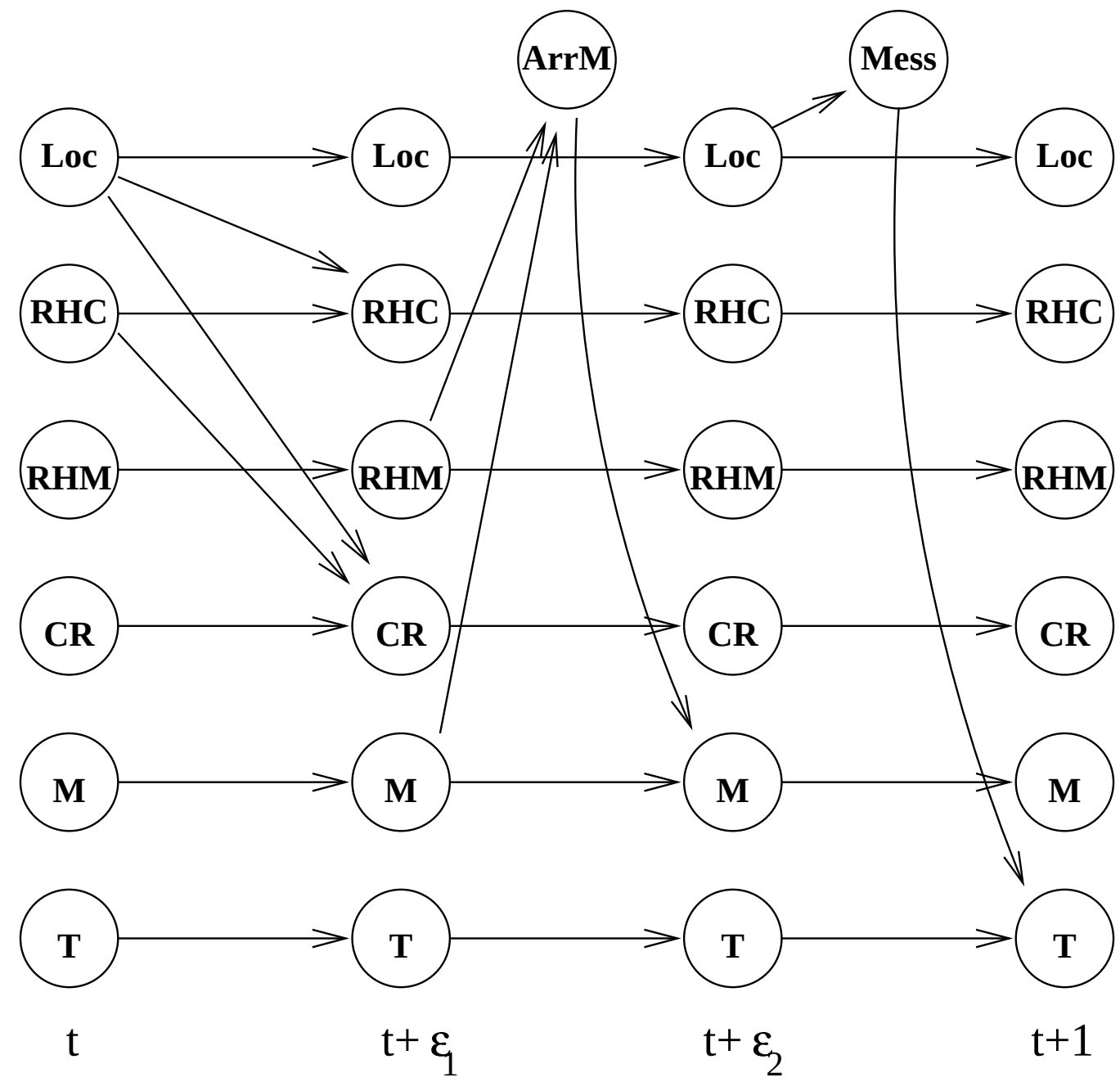

Figure 19: An simplified explicit-event model for *DelC*.

PSOs can provide an effective means for representing actions with correlated effects. Recall the description of the *CClk* action captured in Figure 15, where the robot may drop its coffee as it moves from the hallway, and may drop its mail only if it drops the coffee. In the 2TBN representation of *CClk*, one must have both $RHC^t$ and $RHC^{t+1}$ as parents of $RHM^{t+1}$: we must model the dependence of *RHM* on a *change* in value in the variable *RHC*. Figure 18 shows the *CClk* action in PSO format (for simplicity, we ignore the occurrence of exogenous events). The PSO representation can offer an economical representation of correlated effects such as this since the possible outcomes of moving in the hallway are spelled out explicitly. Specifically, the (possible) simultaneous change in values of the variables in question is made clear.

#### 4.2.4 Explicit-Event Models

Explicit-event models can also be represented using 2TBNs in a somewhat different form. As in our discussion in Section 2.3, the form taken by explicit-event models depends crucially on one's assumptions about the interplay between the effects of the action itself and exogenous events. However, under certain assumptions even explicit-event models can be rather concise.

To illustrate, Figure 19 shows the deliver-coffee action represented as a 2TBN with exogenous events explicitly represented. The first "slice" of the network shows the effects of the action *DelC* without the presence of exogenous events. The subsequent slices describe the effects of the events *ArrM* and *Mess* (we use only two events for illustration). Notice the presence of the extra random variables representing the *occurrence* of the events in question. The CPTs for these nodes reflect the occurrence probabilities for the events under various conditions, while the directed arcs from the event variables to state variables indicate the effects of these events. These probabilities do not depend on all state variables in general; thus, this 2TBN represents the occurrence vectors (see Section 2.3) in a compact form. Also notice that, in contrast to the event occurrence variables, we do not explicitly represent the action occurrence as a variable in the network, since we are modeling the effect on the system *given* that the action was taken.[35]

This example reflects the assumptions described in Section 2.3, namely, that events occur after the action takes place and that event effects are commutative, and for this reason the ordering of the events *ArrM* and *Mess* in the network is irrelevant. Under this model, the system actually passes through two intermediate though not necessarily distinct states as it goes from stage $t$ to stage $t+1$; we use subscripts $\varepsilon_1$ and $\varepsilon_2$ to suggest this process. Of course, as described earlier, not all actions and events can be combined in such a decomposable way; more complex combination functions can also be modeled using 2TBNs (for one example, see Boutilier & Puterman, 1995).

#### 4.2.5 Equivalence of Representations

An obvious question one might ask concerns the extent to which certain representations are inherently more concise than others. Here we focus on the standard implicit-event models, describing some of the domain features that make the different representations more or less suitable.

Both 2TBN and PSO representations are oriented toward representing the changes in the values of the state variables induced by an action; a key distinction lies in the fact that 2TBNs model the influence on each variable separately, while the PSO model explicitly represents complete outcomes. A simple 2TBN—a network with no synchronic arcs—can be used to represent an action in cases where there are no correlations among the action's effect on different state variables. In the worst case, when the effect on each variable differs at each state, each time $t+1$ variable must have all time $t$ variables as parents. If there are no regularities that can be exploited in structured CPT representations, then such an action requires the specification of $O(n2^n)$ parameters (assuming boolean variables), compared with the $2^{2n}$ entries required by an explicit transition matrix. When the number of parents of any variable is bounded by $k$, then we need specify no more than $n2^k$ conditional probabilities. This can be further reduced if the CPTs exhibit structure (e.g., can be represented concisely in a decision tree). For instance, if the CPT can be captured by the representation of choice with no more than $f(k)$ entries, where $f$ is a polynomial function of the number of parents of a variable, then the representation size, $O(n \cdot f(k))$, is polynomial in the number of state variables. This is often the case, for instance, in actions where one of its (stochastic) effects on a variable requires that some number of (pre-) conditions hold; if any of them do not, a different effect comes into play.

A PSO representation may not be as concise as a 2TBN when an action has multiple independent stochastic effects. A PSO requires that each possible change list be enumerated with its corresponding probability of occurrence. The number of such changes grows exponentially with the number of variables affected by the action. This fact is evident in

35. Sections 4.2.7 and 4.3 discuss representations that model the choice of action explicitly as a variable in the network.

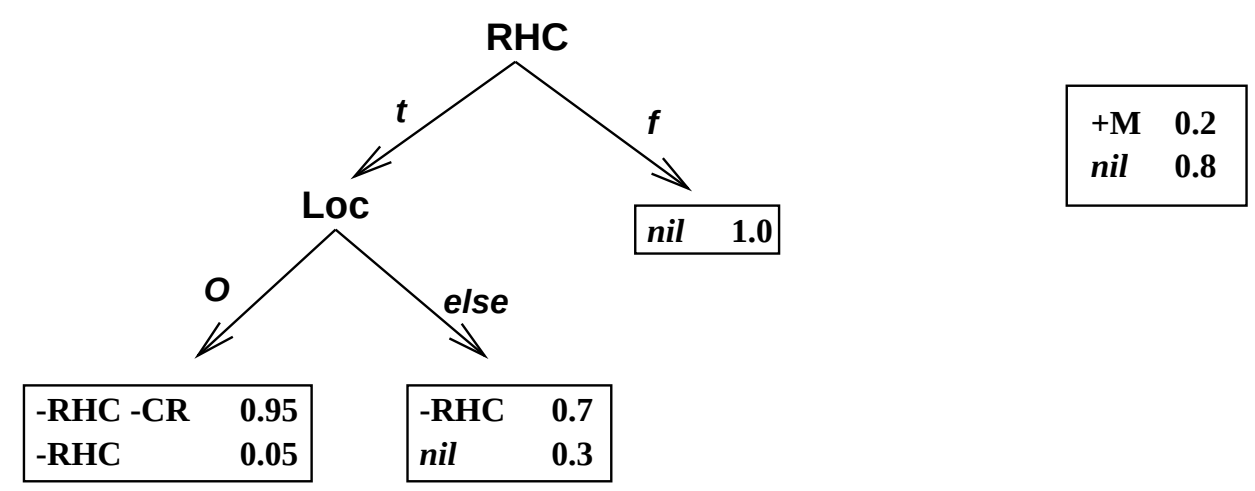

Figure 20: A "factored" PSO representation for the *DelC* action.

Figure 17, where the impact of exogenous events affects a number of variables stochastically and independently. The problem can arise with respect to "direct" action effects, as well. Consider an action in which a set of 10 unpainted parts is spray painted; each part is successfully painted with probability 0.9, and these successes are uncorrelated. Ignoring the complexity of representing different conditions under which the action could take place, a simple 2TBN can represent such an action with 10 parameters (one success probability per part). In contrast, a PSO representation might require one to list all $2^{10}$ distinct change lists and their associated probabilities. Thus, a PSO representation can be exponentially larger (in the number of affected variables) than a simple 2TBN representation.

Fortunately, if certain variables are affected deterministically, these do not cause the PSO representation to blow up. Furthermore, PSO representations can also be modified to exploit the independence of an action's effects on different state variables (Boutilier & Dearden, 1994; Dearden & Boutilier, 1997), thus escaping this combinatorial difficulty. For instance, we might represent the *DelC* action shown in Figure 17 in the more "factored form" illustrated in Figure 20 (for simplicity, we show only the effect of the action and the exogenous event *ArrM*). Much like a 2TBN, we can determine an overall effect by combining the change sets (in the appropriate contexts) and multiplying the corresponding probabilities.

Simple 2TBNs defined over the original set of state variables are not sufficient to represent all actions.[36] Correlated action effects require the presence of synchronic arcs. In the worst case, this means that time $t+1$ variables can have up to $2n-1$ parents. In fact, the acyclicity condition assures that in the worst case, the total number of parents is $\sum_{k=1}^{n} 2k-1$; thus, we end up specifying $O(2^{2n})$ entries, the same as required by an explicit transition matrix. However, if the number of parents (whether occurring within the time slice $t$ or $t+1$) can be bounded, or if regularities in the CPTs allow a compact representation, then 2TBNs can still be profitably used.

PSO representations compare more favorably to 2TBNs in cases in which most of an action's effects on different variables are correlated. In this case, PSOs can provide a somewhat more economical representation of action effects, primarily because one needn't worry about frame conditions. The main advantage of PSOs is that one need not enlist the aid of probabilistic reasoning procedures to determine the transitions induced by actions with correlated effects. Contrast the explicit specification of outcomes in PSOs with the type of reasoning required to determine the joint effects of an action represented in 2TBN

36. However, Section 4.2.6 discusses certain problem transformations that do render simple 2TBNs sufficient for any MDP.

form with synchronic arcs, as described in Section 4.1. Essentially, correlated effects are "compiled" into explicit outcomes in PSOs.

Recent results by Littman (1997) have shown that simple 2TBNs and PSOs can both be used to represent any action represented as a 2TBN without an exponential blowup in representation size. This is effected by a clever problem transformation in which new sets of actions and propositional variables are introduced (using either a simple 2TBN or PSO representation). The structure of the original 2TBN is reflected in the new planning problem, incurring no more than a polynomial increase in the size of the input action descriptions and the description of any policy. Though the resulting policy consists of actions that do not exist in the underlying domain, extracting the true policy is not difficult. It should be noted, however, that while such a representation can automatically be constructed from a general 2TBN specification, it is unlikely that it could be provided directly, since the actions and variables in the transformed problem have no "physical" meaning in the original MDP.

#### 4.2.6 Transformations to Eliminate Synchronic Constraints

The discussion above has assumed that the variables or propositions used in the 2TBN or PSO action descriptions are the original state variables. However, certain problem transformations can be used to ensure that one can represent any action using simple 2TBNs, as long as one does not require the original state variables to be used. One such transformation simply clusters all variables on which some action has a correlated effect. A new *compound variable*—which takes as values assignments to the clustered variables—can then be used in the 2TBN, removing the need for synchronic arcs. Of course, this variable will have a domain size exponential in the number of clustered variables.

Some of the intuitions underlying PSOs can be used to convert general 2TBN action descriptions to simple 2TBN descriptions with explicit "events" dictating the precise outcome of the action. Intuitively, this event can occur in $k$ different forms, each corresponding to a different change list induced by the action (or a change list with respect to the variables in question). As an example, we can convert the "action" description for *CClk* in Figure 15 into the explicit-event model shown in Figure 21.[37] Notice that the "event" takes on values corresponding to the possible effects on the correlated variables *RHC* and *RHM*. Specifically, $a$ denotes the event of the robot escaping the hallway successfully without losing its cargo, $b$ denotes the event of the robot losing only its coffee, and $c$ denotes the event of losing both the coffee and the mail. In effect, the event space represents all possible "combined" effects, obviating the need for synchronic arcs in the network.

#### 4.2.7 Actions as Explicit Nodes in the Network

One difficulty with the 2TBN and PSO approach to action description is that each action is represented separately, offering no opportunity to exploit patterns across actions. For instance, the fact that location persists in all actions except moving clockwise or counterclockwise means that the "frame axiom" is duplicated in the 2TBN for all other actions (this is not the case for PSOs, of course). In addition, *ramifications* (or correlated action

37. While Figure 15 describes a Markov chain induced by a policy, the representation of *CClk* can easily be extracted from it.

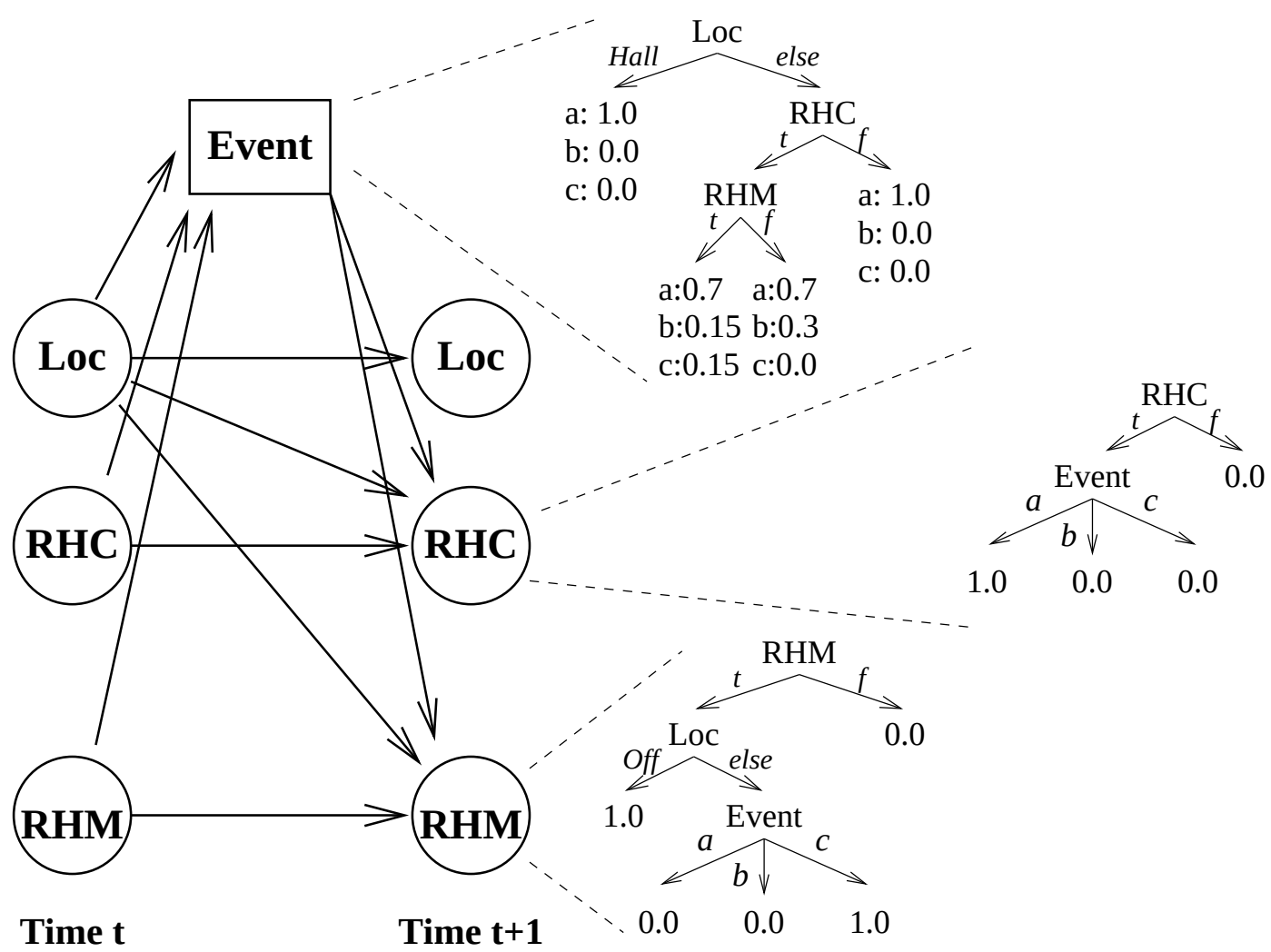

Figure 21: An explicit-event model that removes correlations.

effects) are duplicated across actions as well. For instance, if a coffee request occurs (with probability 0.2) only when the robot *ends up* in the office, then this correlation is duplicated across all actions. A more compelling example might be one in which the robot can move a briefcase to a new location in one of a number of ways. We'd like to capture the fact (or ramification) that the contents of the briefcase move to the same location as the briefcase regardless of the action that moves the briefcase.

To circumvent this difficulty, we can introduce the choice of action as a "random variable" in the network, conditioning the distribution over state variable transitions on the value of this variable. Unlike state variables (or event variables in explicit event models), we do not generally require a distribution over this action variable—the intent is simply to model schematically the conditional state-transition distributions *given* any particular choice of action. This is because the choice of action will be dictated by the decision maker once a policy is determined. For this reason, anticipating terminology used for influence diagrams (see Section 4.3), we call these nodes *decision nodes* and depict them in our network diagrams with boxes. Such a variable can take as its value any action available to the agent.

A 2TBN with an explicit decision node is shown in Figure 22. In this restricted example, we might imagine the decision node can take one of two values, *Clk* or *CClk*. The fact that the issuance of a coffee request at $t+1$ depends on whether the robot successfully moved from (or remained in) the office is now represented "once" by the arc between $Loc^{t+1}$ and $CR^{t+1}$, rather than repeated across multiple action networks. Furthermore, the noisy persistence of $M$ under both actions is also represented only once (adding the action $PUM$, however, undercuts this advantage as we will see when we try to combine actions).

One difficulty with this straightforward use of decision nodes (which is the standard representation in the influence diagram literature) is that adding candidate actions can cause an explosion in the network's dependency structure. For example, consider the two action networks shown in Figure 23(a) and (b). Action $a1$ makes $Y$ true with probability 0.9 if $X$ is true (having no effect otherwise), while $a2$ makes $Y$ true if $Z$ is true.

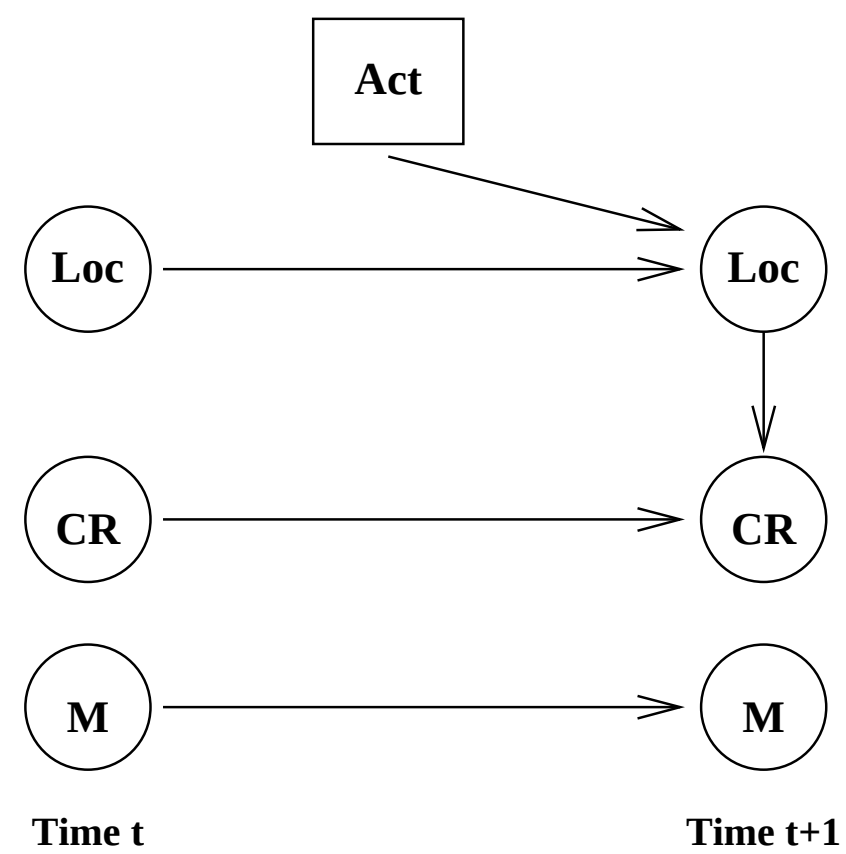

Figure 22: An influence diagram for a restricted process.

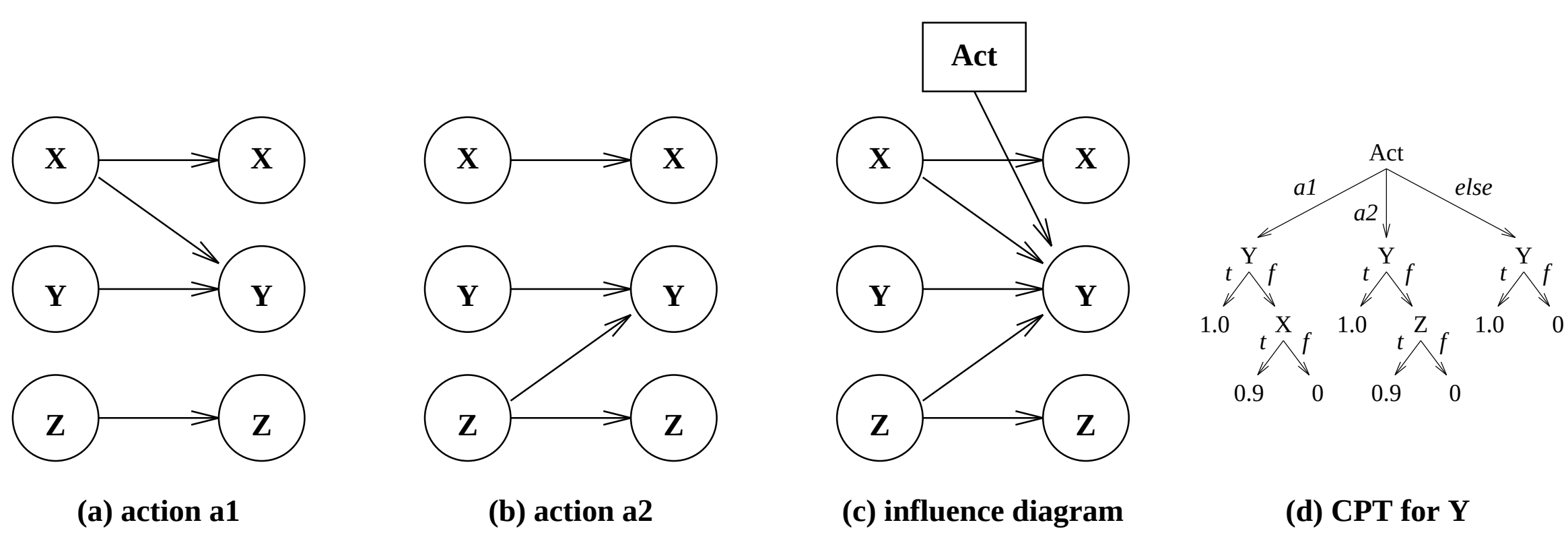

Figure 23: Unwanted dependencies in influence diagrams.

Combining these actions in a single network in the obvious way produces the influence diagram shown in Figure 23(c). Notice that $Y$ now has four parent nodes, inheriting the union of all its parents in the individual networks (plus the action node) and requiring a CPT with 16 entries for actions $a1$ and $a2$ together with eight additional entries for *each* action that does not affect $Y$. The individual networks reflect the fact that $Y$ depends on $X$ only when $a1$ is performed and on $Z$ only when $a2$ is performed. This fact is lost in the naively constructed influence diagram. However, structured CPTs can be used to recapture this independence and compactness of representation: the tree of Figure 23(d) captures the distribution much more concisely, requiring only eight entries. This structured representation also allows us concisely to express that $Y$ persists under all other actions. In large domains, we expect variables to generally be unaffected by a substantial number of (perhaps most) actions, thus requiring representations such as this for influence diagrams. See (Boutilier & Goldszmidt, 1996) for a deeper discussion of this issue and its relationship to the frame problem.

While we provide no distributional information over the action choice, it is not hard to see that a 2TBN with an explicit decision node can be used to represent the Markov chain induced by a particular policy in a very natural way. Specifically, by adding arcs from state variables at time $t$ to the decision node, the value of the decision node (i.e., the choice of action at that point) can be dictated by the prevailing state.[38]

### 4.3 Influence Diagrams

*Influence diagrams* (Howard & Matheson, 1984; Shachter, 1986) extend Bayesian networks to include special *decision nodes* to represent action choices, and *value nodes* to represent the effect of action choice on a value function. The presence of decision nodes means that action choice is treated as a variable under the decision maker's control. Value nodes treat reward as a variable influenced (usually deterministically) by certain state variables.

Influence diagrams have not typically been associated with the schematic representation of stationary systems, instead being used as a tool for decision analysts where the sequential decision problem is carefully handcrafted. This more generic use of influence diagrams has been discussed by Tatman and Shachter (1990). In any event, there is no theory of plan *construction* associated with influence diagrams: the choice of all possible actions at each stage must be explicitly encoded in the model. Influence diagrams are, therefore, usually used to model finite-horizon decision problems by explicitly describing the evolution of the process at each stage in terms of state variables.

As in Section 4.2.7, decision nodes take as values specific actions, though the set of possible actions can be tailored to the particular stage. In addition, an analyst will generally include at each stage only state variables that are thought relevant to the decision at that or subsequent stages. Value nodes are also a key feature of influence diagrams and are discussed Section 4.5. Usually, a single value node is specified, with arcs indicating the

38. More generally, a randomized policy can be represented by specifying a distribution over possible actions conditioned on the state.

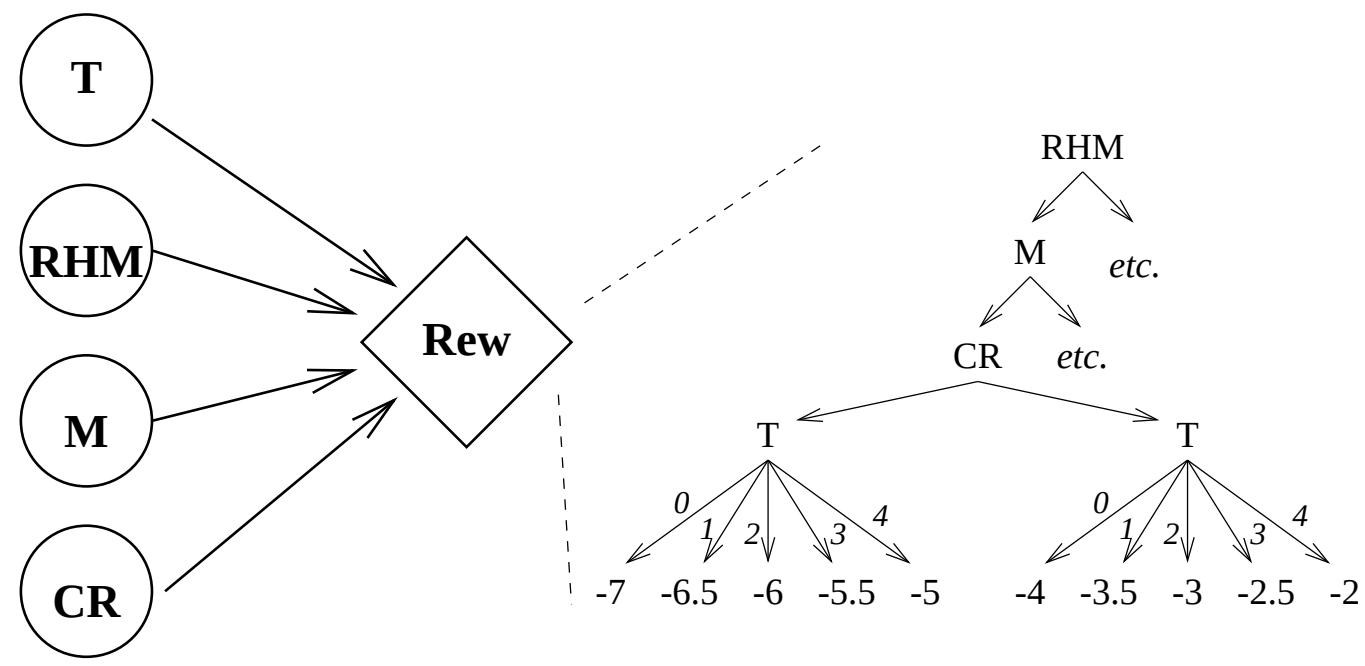

Figure 24: The representation of a reward function in an influence diagram.

influence of particular state and decision variables (often over multiple stages) on the overall value function.

Influence diagrams are typically used to model partially observable problems. An arc from a state variable to a decision node reflects the fact that the value of that state variable is available to the decision maker at the time the action is to be chosen. In other words, this variable's value forms part of the observation made at time $t$ prior to the action being selected at time $t+1$, and the policy constructed can refer to this variable. Once again, this allows a compact specification of the observation probabilities associated with a system. The fact that the probability of a given observation depends directly only on certain variables and not on others can mean that far fewer model parameters are required.

### 4.4 Factored Reward Representation

We have already noted that it is very common in formulating MDP problems to adopt a simplified value function: assigning rewards to states and costs to actions, and evaluating histories by combining these factors according to some simple function like addition. This simplification alone allows a representation for the value function significantly more parsimonious than one based on a more complex comparison of complete histories. Even this representation requires an explicit enumeration of the state and action space, however, motivating the need for more compact representations for these parameters. Factored representations for rewards and action costs can often obviate the need to enumerate state and action parameters explicitly.

Like an action's effect on a particular variable, the reward associated with a state often depends only on the values of certain features of the state. For example, in our robot domain, we can associate rewards or penalties with undelivered mail, with unfulfilled coffee requests and with untidiness in the lab. This reward or penalty is independent of other variables, and individual rewards can be associated with the *groups* of states that differ on the values of the relevant variables. The relationship between rewards and state variables is represented in *value nodes* in influence diagrams, represented by the diamond in Figure 24. The *conditional reward table* (CRT) for such a node is a table that associates a reward with every combination of values for its parents in the graph. This table, not shown in Figure 24, is locally exponential in the number of relevant variables. Although Figure 24 shows the case of a stationary Markovian reward function, influence diagrams can be used to represent nonstationary or history-dependent rewards and are often used to represent value functions for finite-horizon problems.

Although in the worst case the CRT will take exponential space to store, in many cases the reward function exhibits structure, allowing it to be represented compactly using decision trees or graphs (Boutilier et al., 1995), Strips-like tables (Boutilier & Dearden, 1994), or logical rules (Poole, 1995, 1997a). Figure 24 shows a fragment of one possible decision-tree representation for the reward function used in the running example.

The independence assumptions studied in multiattribute utility theory (Keeney & Raiffa, 1976) provide yet another way in which reward functions can be represented compactly. If we assume that the component attributes of the reward function make independent contributions to a state's total reward, the individual contributions can be combined functionally. For instance, we might imagine penalizing states where $CR$ holds with a (partial) reward of $-3$, penalizing situations where there is undelivered mail ($M \vee RHM$) with $-2$, and penalizing untidiness $T(i)$ with $i - 4$ (i.e., in proportion to how untidy things are). The reward for any state can then be determined simply by adding the individual penalties associated with each feature. The individual component rewards along with the combination function constitute a compact representation of the reward function. The tree fragment in Figure 24, which reflects the additive independent structure just described, is considerably more complex than a representation that defines the (independent) rewards for individual propositions separately. The use of additive reward functions for MDPs is considered in (Boutilier, Brafman, & Geib, 1997; Meuleau, Hauskrecht, Kim, Peshkin, Kaelbling, Dean, & Boutilier, 1998; Singh & Cohn, 1998).

Another example of structured rewards is the *goal* structure studied in classical planning. Goals are generally specified by a single proposition (or a set of literals) to be achieved. As such, they can generally be represented very compactly. Haddawy and Hanks (1998) explore generalizations of goal-oriented models that permit extensions such as partial goal satisfaction, yet still admit compact representations.

### 4.5 Factored Policy and Value Function Representation

The techniques studied so far have been concerned with the input specification of the MDP: the states, actions, and reward function. The components of a problem's *solution*—the policy and optimal value function—are also candidates for compact structured representation.

In the simplest case, that of a stationary policy for a fully observable problem, a policy must associate an action with every state, nominally requiring a representation of size $O(|\mathcal{S}|)$. The problem is exacerbated for nonstationary policies and POMDPs. For example, the policy for a finite-horizon FOMDP with $T$ stages generates a policy of size $O(T|\mathcal{S}|)$. For a finite-horizon POMDP, each possible observable history of length $t < T$ might require a different action choice; as many as $\sum_{k=1}^{T} b^k$ such histories can be generated by a fixed policy, where $b$ is the maximum number of possible observations one can make following an action.[39]

The fact that policies require too much space motivates the need to find compact functional representations, and standard techniques like the tree structures discussed above for

39. Other methods of dealing with POMDPs, by conversion to FOMDPs over belief space (see Section 2.10.2), are more complex still.

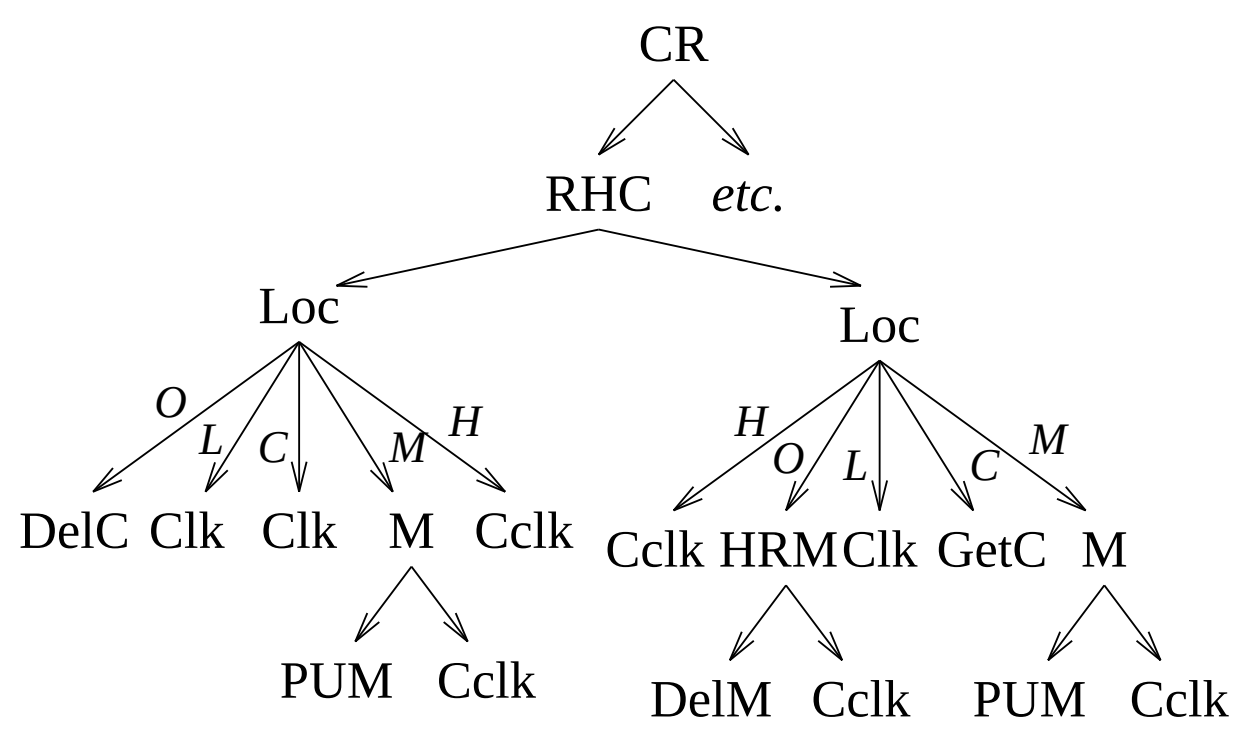

Figure 25: A tree representation of a policy.

actions and reward functions can be used to represent policies and value functions as well. Here we focus on stationary policies and value functions for FOMDPs, for which any logical function representation may be used. For example, Schoppers (1987) uses a Strips-style representation for *universal plans*, which are deterministic, plan-like policies. Decision trees have also been used for policies and value functions (Boutilier et al., 1995; Chapman & Kaelbling, 1991). An example policy for the robot domain specified with a decision tree is given in Figure 25. This policy dictates that, for instance, if *CR* and *RHC* are true: (a) the robot deliver the coffee to the user if it is in the office, and (b) it move toward the office if it is not in the office, unless (c) there is mail and it is in the mailroom, in which case it should pickup the mail on its way.

### 4.6 Summary

In this section we discussed a number of compact factored representations for components of an MDP. We began by discussing intensional state representations, then temporal Bayesian networks as a device for representing the system dynamics. Tree-structured conditional probability tables (CPTs) and probabilistic Strips operators (PSOs) were introduced as an alternative to transition matrices. Similar tree structures and other logical representations were introduced for representing reward functions, value functions, and policies.

While these representations can often be used to *describe* a problem compactly, by themselves they offer no guarantee that the problem can be *solved* effectively. In the next section we explore algorithms that use these factored representations to avoid iterating explicitly over the entire set of states and actions.

## 5. Abstraction, Aggregation, and Decomposition Methods

The greatest challenge in using MDPs as the basis for DTP lies in discovering computationally feasible methods for the construction of optimal, approximately optimal or satisficing policies. Of course, arbitrary decision problems are intractable—even producing satisficing or approximately optimal policies is generally infeasible. However, the previous sections suggest that many realistic application domains may exhibit considerable structure, and furthermore that the structure can be modeled explicitly and exploited so that typical problems can be solved effectively. For instance, structure of this type can lead to compact factored representations of both input data and output policies, often polynomial-sized with respect to the number of variables and actions describing the problem. This suggests that for these compact problem representations, policy construction techniques can be developed that exploit this structure and are tractable for many commonly occurring problem instances.

Both the dynamic programming and state-based search techniques described in Section 3 exploit structure of a different kind. Value functions that can be decomposed into state-dependent reward functions, or state-based goal functions, can be tackled by dynamic programming and regression search, respectively. These algorithms exploit the structure in decomposable value functions to prevent having to search explicitly through all possible policies. However, while these algorithms are polynomial in the size of the state space, the curse of dimensionality makes even these algorithms infeasible for practical problems. Though compact problem representations aid in the specification of large problems, it is clear that a large system can be specified compactly only if the representation exploits "regularities" found in the domain. Recent AI research on DTP has stressed using the regularities implicit in compact representations to speed up the planning process. These techniques focus on both optimal and approximately optimal policy construction.

In the following subsection we focus on *abstraction* and *aggregation* techniques, especially those that manipulate factored representations. Roughly, these techniques allow the explicit or implicit grouping of states that are indistinguishable with respect to certain characteristics (e.g., value or optimal action choice). We refer to a set of states grouped in this manner as an *aggregate* or *abstract state*, or sometimes as a *cluster*, and assume that the set of abstract states constitutes a *partition* of the state space; that is to say, every state is in exactly one abstract state and the union of all abstract states comprises the entire state space.[40] By grouping similar states, each abstract state can be treated as a single state, thus alleviating the need to perform computations for each state individually. These techniques can be used for approximation if the elements of an abstract state are only approximately indistinguishable (e.g., if the values of those states lie within some small interval).

We then look at the use of *problem decomposition* techniques in which an MDP is broken into various pieces, each of which is solved independently; the solutions are then pieced together or used to guide the search for a global solution. If subprocesses whose solutions interact minimally are treated as independent, we might expect an approximately optimal global solution. Furthermore, if the structure of the problem requires a solution to a particular subproblem only, then the solutions to other subproblems can be ignored altogether.

Related is the use of *reachability analysis* to restrict attention to "relevant" regions of state space. Indeed, reachability analysis and the *communicating structure* of an MDP can be used to form certain types of decompositions. Specifically, we distinguish *serial* decompositions from *parallel* decompositions.

The result of a serial decomposition can be viewed as a partitioning of the state space into *blocks*, each representing a (more or less) independent subprocess to be solved. In serial decomposition, the relationship between blocks is generally more complicated than in the case of abstraction or aggregation. In a partition resulting from decomposition, the

40. We might also group states into non-disjoint sets that cover the entire state space. We do not consider such *soft-state aggregation* here, but see (Singh, Jaakkola, & Jordan, 1994).

states within a particular block may behave quite differently with respect to (say) value or dynamics. The important consideration in choosing a decomposition is that it is possible to represent each block compactly and to compute efficiently the consequences of moving from one block to another and, further, that the subproblems corresponding to the subprocesses can themselves be solved efficiently.

A parallel decomposition is somewhat more closely related to an abstract MDP. An MDP is divided into "parallel sub-MDPs" such that each decision or action causes the state to change within each sub-MDP. Thus, the MDP is the cross product or join of the sub-MDPs (in contrast to the union, as in serial decomposition). We briefly discuss several methods that are based on parallel MDP decomposition.

### 5.1 Abstraction and Aggregation

One way problem structure can be exploited in policy construction relies on the notion of *aggregation*—grouping states that are indistinguishable with respect to certain problem characteristics. For example, we might group together all states that have the same optimal action, or that have the same value with respect to the $k$-stage-to-go value function. These aggregates can be constructed during the solution of the problem.

In AI, emphasis has generally been placed on a particular form of aggregation, namely *abstraction* methods, in which states are aggregated by ignoring certain problem features. The policy in Figure 25 illustrates this type of abstraction: those states in which *CR*, *RHC* and *Loc*(*O*) are true are grouped, and the same action is selected for each such state. Intuitively, when these three propositions hold, other problem features are ignored and abstracted away (i.e., they are deemed irrelevant). A decision-tree representation of a policy or a value function partitions the state space into a distinct cluster for each leaf of the tree. Other representations (e.g., Strips-like rules) abstract the state space similarly.

It is precisely this type of abstraction that is used in the compact, factored representations of actions and goals discussed in Section 4. In the 2TBN shown in Figure 16, the effect of the action *DelC* on the variable *CR* is given by the CPT for $CR^{t+1}$; however, this (stochastic) effect is the same at any state for which the parent variables have the same value. This representation abstracts away other variables, combining states that have distinct values for the irrelevant (non-parent) variables. Intensional representations often make it easy to decide which features to ignore at a certain stage of problem solving, and thus (implicitly) how to aggregate the state space.

There are at least three dimensions along which abstractions of this type can be compared. The first is *uniformity*: a uniform abstraction is one in which variables are deemed relevant or irrelevant uniformly across the state space, while a nonuniform abstraction allows certain variables to be ignored under certain conditions and not under others. The distinction is illustrated schematically in Figure 26. The tabular representation of a CPT can be viewed as a form of uniform abstraction—the effect of an action on a variable is distinguished for all clusters of states that differ on the value of a parent variable, and is not distinguished for states that agree on parent variables but disagree on others—while a decision tree representation of a CPT embodies a nonuniform abstraction.

A second dimension of comparison is *accuracy*. States are grouped together on the basis of certain characteristics, and the abstraction is called *exact* if all states within a cluster agree on this characteristic. A non-exact abstraction is called *approximate*. This is illustrated schematically in Figure 26: the exact abstraction groups together states that agree on the value assigned to them by a value function, while the approximate abstraction allows states to be grouped together that differ in value. The extent to which these states differ is often used as a measure of the quality of an approximate abstraction.

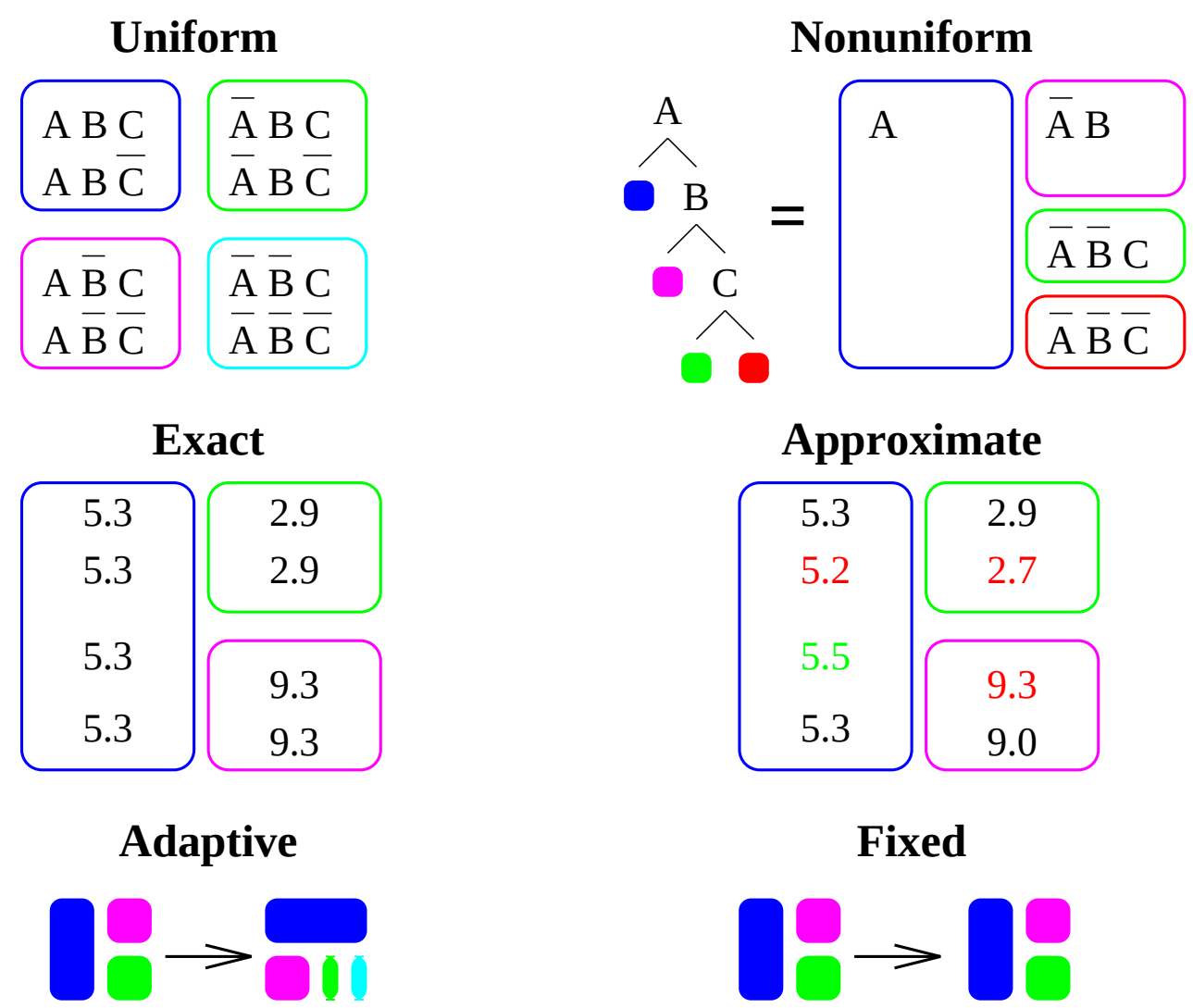

Figure 26: Different forms of state space abstraction.

A third dimension is *adaptivity*. Technically, this is a property not of an abstraction itself, but of how abstractions are used by a particular algorithm. An *adaptive* abstraction technique is one in which the abstraction can change during the course of computation, while a *fixed* abstraction scheme groups together states once and for all (again, see Figure 26). For example, one can imagine using an abstraction in the representation of a value function $V^k$, then revising this abstraction to represent $V^{k+1}$ more accurately.

Abstraction and aggregation techniques have been studied in the OR literature on MDPs. Bertsekas and Castanon (1989) develop an adaptive aggregation (as opposed to abstraction) technique. The proposed method operates on flat state spaces, however, and therefore does not exploit implicit structure in the state space itself. An adaptive, uniform abstraction method is proposed by Schweitzer et al. (1985) for solving stochastic queuing models. These methods, often referred to as *aggregation-disaggregation procedures*, are typically used to accelerate the calculation of the value function for a fixed policy. Value-function calculation requires computational effort at least quadratic in the size of the state space, which is impractical for very large state spaces. In aggregation-disaggregation procedures, the states are first aggregated into clusters. A system of equations is then solved, or a series of summations performed, requiring effort no more than cubic in the number of clusters. Next, a disaggregation step is performed for each cluster, requiring effort at least linear in the size of the cluster. The net result is that the total work, while at least linear in the total number of states, is at worst cubic in the size of the largest cluster.

In DTP it is generally assumed that computations even linear in the size of the full state space are infeasible. Therefore it is important to develop methods that perform work polynomial in the log of the size of the state space. Not all problems are amenable to such reductions without some (perhaps unacceptable) sacrifice in solution quality. In the following section, we review some recent techniques for DTP aimed at achieving such reductions.

#### 5.1.1 Goal Regression and Classical Planning

In Section 3.2 we introduced the general technique of regression (or backward) search through state space to solve classical planning problems, those involving deterministic actions and performance criteria specified in terms of reaching a goal-satisfying state. One difficulty is that such a search requires that any branch of the search tree lead to a *particular* goal state. This commitment to a goal state may have to be retracted (by backtracking the search process) if no sequence of actions can lead to that particular goal state from the initial state. However, a goal is usually specified as a set of literals $G$ representing a set of states, where reaching any state in $G$ is equally suitable—it may, therefore, be wasteful to restrict the search to finding a plan that reaches a *particular* element of $G$.

*Goal regression* is an abstraction technique that avoids the problem of choosing a particular goal state to pursue. A regression planner works by searching for a sequence of actions as follows: the *current set of subgoals* $SG_0$ is initialized as $G$. At each iteration an action $\alpha$ is selected that achieves one or more of the current subgoals of $SG_i$ without deleting the others, and whose preconditions do not conflict with the "unachieved subgoals." The subgoals so achieved are removed from the current subgoal set and replaced by a formula representing the context under which $\alpha$ will achieve the current subgoals, forming $SG_{i+1}$. This process is known as *regressing* $SG_i$ through $\alpha$. The process is repeated until one of two conditions holds: (a) the current subgoal set is satisfied by the initial state, in which case the current sequence of actions so selected is a successful plan; or (b) no action can be applied, in which case the current sequence cannot be extended into a successful plan and some earlier action choice must be reconsidered.

**Example 5.1** As an example, consider the simplified version of the robot planning example used in Section 3.1 to illustrate value iteration: the robot has only four actions *PUM*, *GetC*, *DelC* and *DelM*, which we make deterministic in the obvious way. The initial state $s_{init}$ is $\langle CR, M, \overline{RHC}, \overline{RHM} \rangle$ and the goal set $G$ is $\{\overline{CR}, \overline{M}\}$. Regressing $G$ through *DelM* results in $SG_1 = \{\overline{CR}, M, RHM\}$. Regressing $SG_1$ through *DelC* results in $SG_2 = \{RHC, M, RHM\}$. Regressing $SG_2$ through *PUM* results in $SG_3 = \{RHC, M\}$. Regressing $SG_3$ through *GetC* results in $SG_4 = \{M\}$. Note that $s_{init} \in SG_4$, so the sequence of actions *GetC*, *PUM*, *DelC*, *DelM* will successfully reach a goal state. □

To see how this algorithm implements a form of abstraction, first note that the goal itself provides an initial partition of the state space, dividing it into one set of states in which the goal is satisfied ($G$) and a second set in which it is not ($\overline{G}$). Viewed as a partition of a zero-stage-to-go value function, $G$ represents those states whose value is positive while $\overline{G}$ represents those states whose value is zero.

Every regression step can be thought of as revising this partition. When the planning algorithm attempts to satisfy the current subgoal set $SG_i$ by applying action $\alpha$, it uses regression to compute the (largest) set of states such that, after executing $\alpha$, all subgoals are satisfied. In particular, the state space is repartitioned into two abstract states: $SG_{i+1}$ and $\overline{SG_{i+1}}$. In this way, the abstraction mechanism implemented by goal regression should be considered *adaptive*. This can be viewed as an $(i+1)$-stage value function: any state satisfying $SG_{i+1}$ can reach a goal state in $i+1$ steps using the action sequence that produced $SG_{i+1}$.[41] The regression process can be stopped when the initial state is a member of the abstract state $SG_{i+1}$. Figure 27 illustrates the repartitioning of the state space into the different regions $SG_{i+1}$ for each of the steps in the example above.

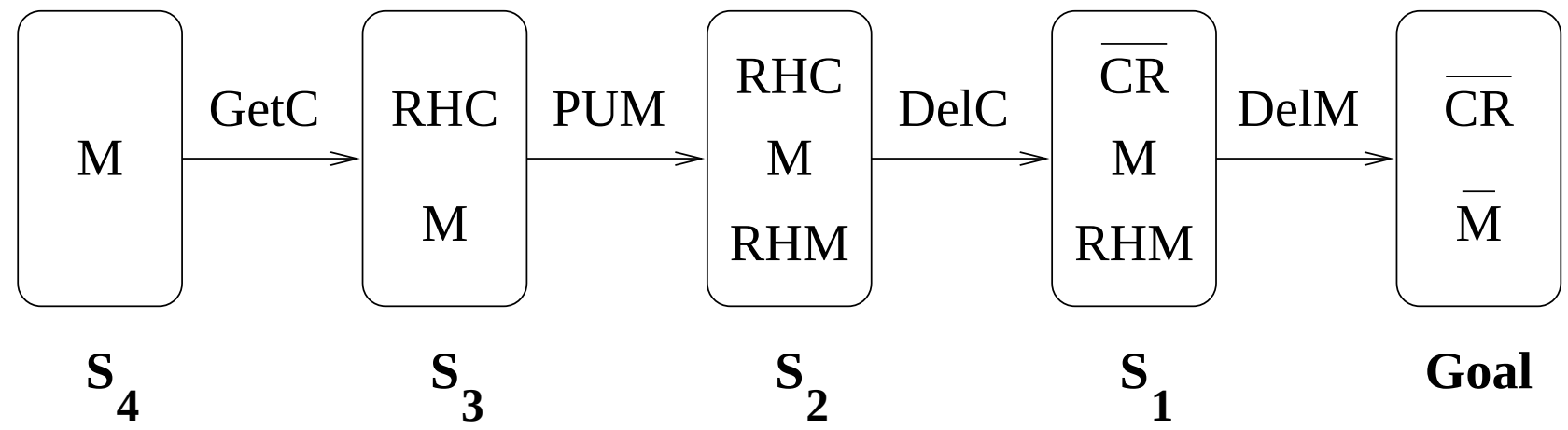

Figure 27: An example of goal regression.

While regression produces a compact representation of something like a value function (as in our discussion of deterministic, goal-based dynamic programming in Section 3.2), the analogy is not exact in that the regions produced by regression record only the property of goal reachability contingent on a *particular choice* of action or action sequence.

Standard dynamic programming methods can be implemented in a structured way by simply noticing that a number of different regions can be produced at the $i$th iteration by considering *all* actions that can be regressed at that stage. The union of all of these regressions form the states that have positive values in $V_i$, thus making the representation of the $i$-stage-to-go value function exact. Notice that each iteration is now more costly, since regression through all actions must be attempted, but this approach obviates the need for backtracking and can ensure that a shortest plan is found. Standard regression does not provide such guarantees without commitment to a particular search strategy (e.g., breadth-first). This use of dynamic programming using Strips action descriptions forms the basic idea of Schoppers's *universal planning* method (Schoppers, 1987).

Another general technique for solving classical planning problems is *partial order planning* (POP) (Chapman, 1987; Sacerdoti, 1975), embodied in such popular planning algorithms as SNLP (McAllester & Rosenblitt, 1991) and UCPOP (Penberthy & Weld, 1992).[42] The main motivation for the least-commitment approach comes from the realization that regression techniques are incrementally building a plan from the end to the beginning (in the temporal dimension). Thus, each iteration must commit to inserting a step *last* in the plan.

In many cases it can be determined that a particular step must appear *somewhere* in the plan, but not necessarily as the *last* step in the plan; and, indeed, in many cases the step

41. It is not the case, however, that states in $\overline{SG_{i+1}}$ cannot reach the goal region in $i+1$ steps. It is only the case that they cannot do so using the specific sequence of actions chosen so far.

42. This type of planning is also sometimes called nonlinear or least-commitment planning. See Weld's (1994) survey for a nice overview.

under consideration *cannot* appear last, but this fact cannot be recognized until later choices reveal an inconsistency. In these cases, a regression algorithm will prematurely commit to the incorrect ordering and will eventually have to backtrack over that choice. For example, suppose in the problem scenario above that the robot can hold only one item at a time, coffee or mail. Picking up mail causes the robot to spill any coffee in its possession, and similarly grasping the coffee makes it drop the mail. The plan generated by regression would no longer be valid: once the first two actions (*DelC* and *DelM*) have been inserted into the plan, no action can be added to achieve *RHC* or *RHM* without making the other one false; the search for a plan would have to backtrack. Ultimately it would be discovered that no successful plan can end with these two actions performed in sequence.

Partial-order planning algorithms proceed much like regression algorithms, choosing actions to achieve unachieved subgoals and using regression to determine new subgoals, but leaving actions unordered to whatever extent possible. Strictly speaking, subgoal *sets* aren't regressed; rather, each unachieved goal or action precondition is addressed separately, and actions are ordered relative to one another only if one action threatens to negate the desired effect of another. In the example above, the algorithm might first place actions *DelC* and *DelM* into the plan, but leave them unordered. *PUM* can be added to the plan to achieve the requirement *RHM* of *DelM*; it is ordered before *DelM* but is still unordered with respect to *DelC*. When *GetC* is finally added to the plan so as to achieve *RHC* for action *DelC*, two *threats* arise. First, *GetC* threatens the desired effect *RHM* of *PUM*. This can be resolved by ordering *GetC* before *PUM* or after *DelM*. Assume the former ordering is chosen. Second, *PUM* threatens the desired effect *RHC* of *GetC*. This threat can also be resolved by placing *PUM* before *GetC* or after *DelC*; since the first threat was resolved by ordering *GetC* before *PUM*, the latter ordering is the only consistent one. The result is the plan *GetC*, *DelC*, *PUM*, *DelM*. No backtracking was required to generate the plan, because the actions were initially unordered, and orderings were introduced only when the discovery of threats required them.

In terms of abstraction, any incomplete, partially ordered plan that is threat-free, but perhaps has certain "open conditions" (unachieved preconditions or subgoals), can be viewed in much the same way as a partially completed regression plan: any state satisfying the open conditions can reach a goal state by executing any total ordering of the plan's actions consistent with current set of ordering constraints. See (Kambhampati, 1997) for a framework that unifies various approaches to solving classical plan-generation problems.

While techniques relying on regression have been studied extensively in the deterministic setting, they have only recently been applied to probabilistic unobservable (Kushmerick et al., 1995) and partially observable (Draper, Hanks, & Weld, 1994b) domains. For the most part, these techniques assume a goal-based performance criterion and attempt to construct plans whose probability of reaching a goal state exceeds some threshold. These augment standard POP methods with techniques for evaluating a plan's probability of achieving the goal, and techniques for improving this probability by adding further structure to the plan. In the next section, we consider how to use regression-related techniques to solve MDPs with performance criteria more general than goals.

### 5.1.2 Stochastic Dynamic Programming with Structured Representations

A key idea underlying propositional goal regression—that one need only regress the relevant propositions through an action—can be extended to stochastic dynamic programming methods, like value iteration and policy iteration, and used to solve general MDPs. There are, however, two key difficulties to overcome: the lack of a specific goal region and the uncertainty associated with action effects.

Instead of viewing the state space as partitioned into goal and non-goal clusters, we consider grouping states according to their expected values. Ideally, we might want to group states according to their value with respect to the optimal policy. Here we consider a somewhat less difficult task, that of grouping states according to their value with respect to a fixed policy. This is essentially the task performed by the policy evaluation step in policy iteration, and the same insights can be used to construct optimal policies.

For a fixed policy, we want to group states that have the same value under that policy. Generalizing the goal versus non-goal distinction, we begin with a partition that groups states according their immediate rewards. Then, using an analogue of regression developed for the stochastic case, we reason backward to construct a new partition in which states are grouped according to their value with respect to the one-stage-to-go value function. We iterate in this manner so that on the $k$th iteration we produce a new partition that groups states according the $k$-stage-to-go value function.

On each iteration, we perform work polynomial in the number of abstract states (and the size of the MDP representation) and, if we are lucky, the total number of abstract states will be bounded by some logarithmic factor of the size of the state space. To implement this scheme effectively, we have to perform operations like regression without ever enumerating the set of all states, and this is where the structured representations for state-transition, value, and policy functions play a role.

For FOMDPs, approaches of this type are taken in (Boutilier, 1997; Boutilier & Dearden, 1996; Boutilier et al., 1995; Boutilier, Dearden, & Goldszmidt, 1999; Dietterich & Flann, 1995; Hoey et al., 1999). We illustrate the basic intuitions behind this approach by describing how value iteration for discounted infinite-horizon FOMDPs might work. We assume that the MDP is specified using a compact representation of the reward function (such as a decision tree) and actions (such as 2TBNs).

In value iteration, we produce a sequence of value functions $V_0, V_1, \cdots, V_n$, each $V_k$ representing the utility of the optimal $k$-stage policy. Our aim is to produce a compact representation of each value function and, using $V_n$ for some suitable $n$, produce a compact representation of the optimal stationary policy. Given a compact representation of the reward function $R$, it is clear that this constitutes a compact representation of $V_0$. As usual, we think of each leaf of the tree as a cluster of states having identical utility. To produce $V_1$ in compact form, we can proceed in two phases.

Each branch of the tree for $V_0$ provides an intensional description—namely, the conjunction of variable values labeling the branch—of an abstract state, or region, comprising states with identical value with respect to the initial value function $V_0$. For any deterministic action $\alpha$, we can perform a regression step using this description to determine the conditions under which, should we perform $\alpha$, we would end up in this cluster. This would, furthermore, determine a region of the state space containing states of identical future value with respect to the execution of $\alpha$ with one stage to go.[43] Unfortunately, nondeterministic actions cannot be handled in quite this way: at any given state, the action might lead to several different regions of $V_0$ with non-zero probability. However, for each leaf in the tree representing $V_0$ (i.e., for each region of $V_0$), we can regress the conjunction $X$ describing that region through action $\alpha$ to produce the conditions under which $X$ becomes true or false with a specified probability. In other words, instead of regressing in the standard fashion to determine the conditions under which $X$ becomes true, we produce a *set* of distinct conditions under which $X$ becomes true with different probabilities. By piecing together the regions produced for the different labels in the description of $V_0$, we can construct a set of regions such that each state in a given region: (a) transitions (under action $\alpha$) to a particular part of $V_0$ with identical probability; and hence (b) has identical expected future value (Boutilier et al., 1995). We can view this as a generalization of propositional goal regression suitable for decision-theoretic problems.

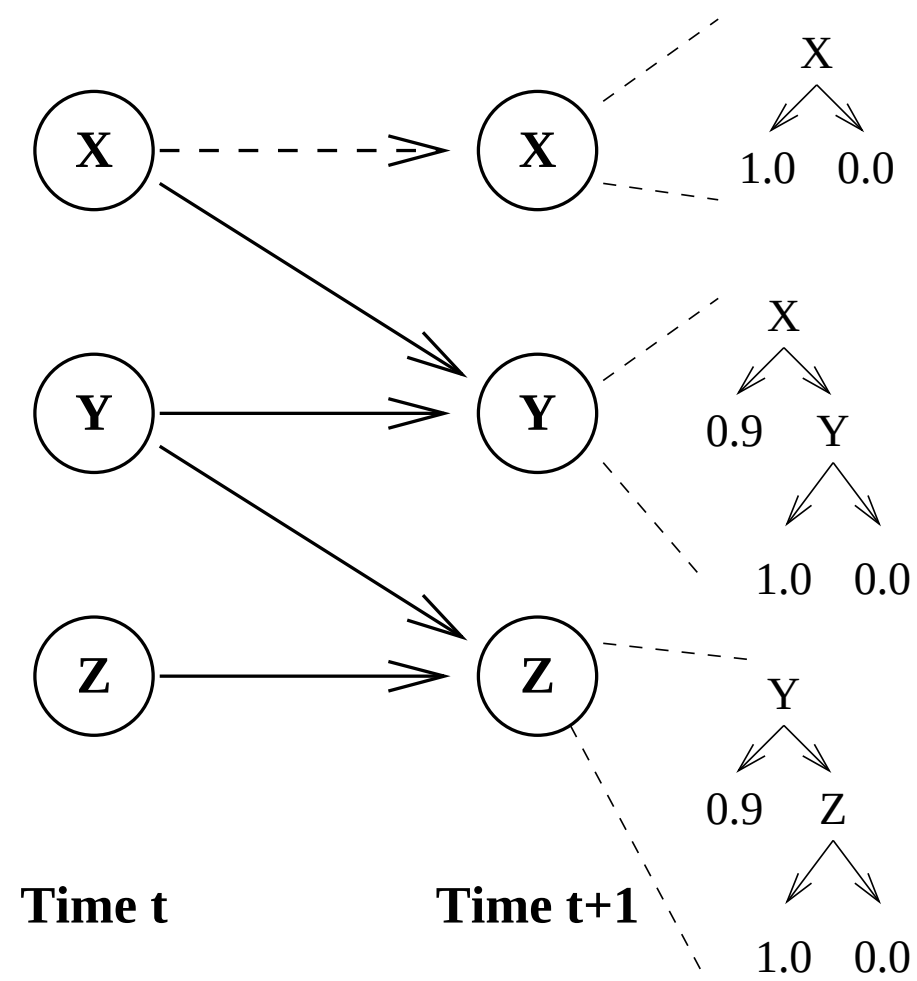

Figure 28: An example action.

**Example 5.2** To illustrate, consider the example action $a$ shown in Figure 28 and the value function $V^0$ shown to the left of Figure 29. In order to generate the set of regions consisting of states whose future value (w.r.t. $V^0$) under $a$ is identical, we proceed in two steps (see Figure 29). We first determine the conditions under which $a$ has a fixed probability of making $Y$ true (hence we have a fixed probability of moving to the left or right subtree of $V^0$). These conditions are given by the tree representing the CPT for node $Y$, which makes up the first portion of the tree representing $V^1$—see Step 1 of Figure 29. Notice that this tree has leaves labeled with the probability of making $Y$ true or (implicitly) false.

If $a$ makes $Y$ true, then we know that its future value (i.e., value with zero stages to go) is 8.1; but if $Y$ becomes false, we need to know whether $a$ makes $Z$ true (to

43. We ignore immediate reward and cost distinctions within the region so produced in our description; recall that the value of performing $\alpha$ at any state $s$ is given by $R(s)$, $C(\alpha, s)$ and expected future value. We simply focus on abstract states whose elements have identical future expected value. Differences in immediate reward and cost can be added after the fact.

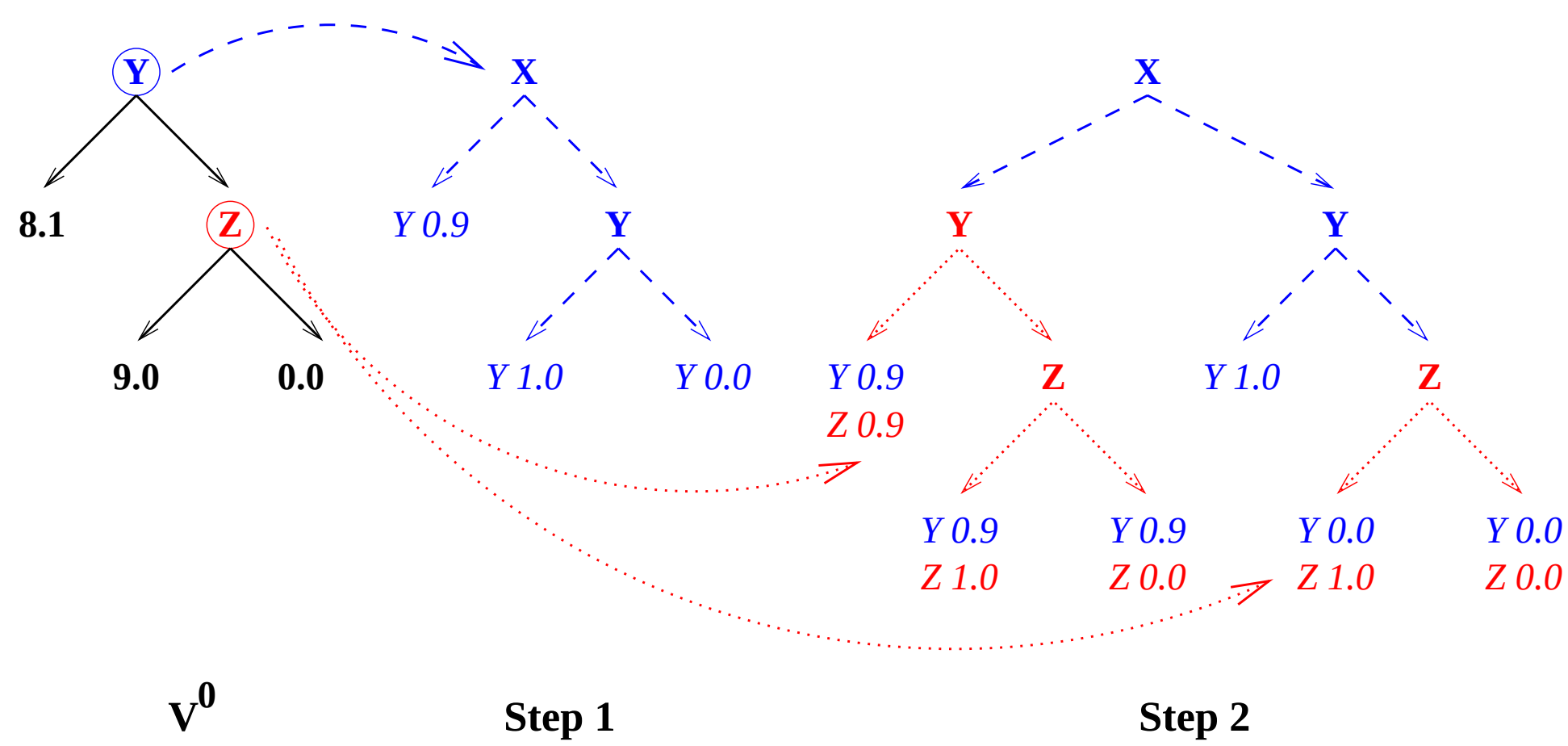

Figure 29: An iteration of decision-theoretic regression. Step 1 produces the portion of the tree with dashed lines, while Step 2 produces the portion with dotted lines.

determine whether the future value is 0 or 9.0). The probability with which $Z$ becomes true is given by the tree representing the CPT for node $Z$. In Step 2 in Figure 29, the conditions in that CPT are conjoined to the conditions required for predicting $Y$'s probability (by "grafting" the tree for $Z$ to the tree for $Y$ given by the first step). This grafting is slightly different at each of the three leaves of the tree for $Y$: (a) the full tree for $Z$ is attached to the leaf $X = t$; (b) the tree for $Z$ is simplified where it is attached to to the leaf $X = f \wedge Y = f$ by removal of the redundant test on variable $Y$; (c) notice that there is no need to attach the tree for $Z$ to the leaf $X = f \wedge Y = t$, since $a$ makes $Y$ true with probability 1 under those conditions (and $Z$ is relevant to the determination of $V^0$ only when $Y$ is false).

At each of the leaves of the newly formed tree we have both $\Pr(Y)$ and $\Pr(Z)$. Each of these joint distributions over $Y$ and $Z$ (the effect of $a$ and these variables is independent by the semantics of the network) tells us the probability of having $Y$ and $Z$ true with zero stages to go given that the conditions labeling the appropriate branch of the tree hold with one stage to go. In other words, the new tree uniquely determines, for any state with one stage remaining, the probability of making any of the conditions labeling the branches of $V^0$ true. The computation of expected future value obtained by performing $a$ with one stage to go can then be placed at the leaves of this tree by taking expectation over the values at the leaves of $V^0$. $\square$

The new set of regions produced this way describes the function $Q_1^\alpha$, where $Q_1^\alpha(s)$ is the value associated with performing $\alpha$ at state $s$ with one stage to go and acting optimally thereafter. These functions (for each action $\alpha$) can be pieced together (i.e., "maxed"—see Section 3.1) to determine $V_1$. Of course, the process can be repeated some number of times to produce $V_n$ for some suitable $n$, as well as the optimal policy with respect to $V_n$.

This basic technique can be used in a number of different ways. Dietterich and Flann (1995) propose ideas similar to these, but restrict attention to MDPs with goal regions and deterministic actions (represented using Strips operators), thus rendering true goal-regression techniques directly applicable.[44] Boutilier et al. (1995) develop a version of modified policy iteration to produce tree-structured policies and value functions, while Boutilier and Dearden (1996) develop the version of value iteration described above. These algorithms are extended to deal with correlations in action effects (i.e., synchronic arcs in the 2TBNs) in (Boutilier, 1997). These abstraction schemes can be categorized as nonuniform, exact and adaptive.

The utility of such exact abstraction techniques has not been tested on real-world problems to date. In (Boutilier et al., 1999), results on a series of abstract process-planning examples are reported, and the scheme is shown to be very useful, especially for larger problems. For example, in one specific problem with 1.7 million states, the tree representation of the value function has only 40,000 leaves, indicating a tremendous amount of regularity in the value function. Schemes like this exploit such regularity to solve problems more quickly (in this example, in much less than half the time required by modified policy iteration) and with much lower memory demands. However, these schemes do involve substantial overhead in tree construction, and for smaller problems with little regularity, the overhead is not repaid in time savings (simple vector-matrix representations methods are faster), though they still generally provide substantial memory savings. What might be viewed as best- and worst-case behavior is also described in (Boutilier et al., 1999). In a series of "linear" examples (i.e., problems with value functions that can be represented with trees whose size is linear in the number of problem variables), the tree-based scheme solves problems many orders of magnitude faster than classical state-based techniques. In contrast, problems with exponentially-many distinct values are also tested (i.e., with a distinct value at each state): here tree-construction methods are required to construct a complete decision tree in addition to performing the same number of expected value and maximization computations as classical methods. In this worst case, tree-construction overhead makes the algorithm run about 100 times slower than standard modified policy iteration.

In (Hoey et al., 1999), a similar algorithm is described that uses *algebraic decision diagrams (ADDs)* (Bahar, Frohm, Gaona, Hachtel, Macii, Pardo, & Somenzi, 1993) rather than trees. ADDs are a simple generalization of *boolean decision diagrams (BDDs)* (Bryant, 1986) that allow terminal nodes to be labeled with real values instead of just boolean values. Essentially, ADD-based algorithms are similar to the tree-based algorithms except that isomorphic subtrees can be shared. This lets ADDs provide more compact representations of certain types of value functions. Highly optimized ADD manipulation and evaluation software developed in the verification community can also be applied to solving MDPs. Initial results provided in (Hoey et al., 1999) are encouraging, showing considerable savings over tree-based algorithms on the same problems. For example, the ADD algorithm applied to the 1.7-million-state example described above revealed the value function to have only 178 distinct values (cf. the 40,000 tree leaves required) and produced an ADD description of the value function with less than 2200 internal nodes. It also solved the same problem in seven minutes, about 40 times faster than earlier reported timing results using decision trees (though some of this improvement was due to the use of optimized ADD software packages). Similar results obtain with other problems (problems of up to 268 million states

44. Dietterich and Flann (1995) also describe their work in the context of reinforcement learning rather than as a method for solving MDPs directly.

were solved in about four hours). Most encouraging is the fact that on the worst-case (exponential) examples, the overhead associated with using ADDs—compared to classical, vector-based methods—is much less than with trees (about a factor of 20 compared to "flat" modified policy iteration with 12 state variables), and lessens as problems become larger. Like tree-based algorithms, these methods have yet to be applied to real-world problems.

With these exact abstraction schemes it is clear that, while in some examples the resulting policies and value functions may be compact, in others the set of regions may get very large (even reaching the level of individual states Boutilier et al., 1995), thus precluding any computational savings. Boutilier and Dearden (1996) develop an approximation scheme that exploits the tree-structured nature of the value functions produced. At each stage $k$, the value function $V_k$ can be pruned to produce a smaller, less accurate tree that approximates $V_k$. Specifically, approximate value functions are represented using trees whose leaves are labeled with an upper and lower bound on the value function in that region; decision-theoretic regression is performed on these bounds. Certain subtrees of the value tree can be pruned when leaves of the subtree are very close in value or when the tree is too large given computational constraints. This scheme is nonuniform, approximate and adaptive. This approximation scheme can be tailored to provide (roughly) the most accurate value function of a given maximum tree size, or the smallest value function (with respect to tree size) of some given minimum accuracy. Results reported in (Boutilier & Dearden, 1996) show that approximation on a small set of examples (including the worst-case examples for tree-based algorithms) allows substantial reduction in computational cost. For instance, in a 10-variable worst-case example, a small amount of pruning introduced an average error of only 0.5% but reduced computation time by a factor of 50. More aggressive pruning tends to increase error and decrease computation time very rapidly; making appropriate tradeoffs in these two dimensions is still to be addressed. This method too remains to be tested and evaluated on realistic problems.

Structured representations and solution algorithms can be applied to problems other than FOMDPs. Methods for solving influence diagrams (Shachter, 1986) exploit structure in a natural way; Tatman and Shachter (1990) explore the connection between influence diagrams and FOMDPs and the relationship between influence diagram solution techniques and dynamic programming. Boutilier and Poole (1996) show how classic history-independent methods for solving POMDPs, based on conversion to a FOMDP with belief states, can exploit the types of structured representations described here. However, exploiting structured representations of POMDPs remains to be explored in depth.

### 5.1.3 Abstract Plans

One of the difficulties with the adaptive abstraction schemes suggested above is the fact that different abstractions must be constructed repeatedly, incurring substantial computational overhead. If this overhead is compensated by the savings obtained during policy construction—e.g., by reducing the number of backups—then it is not problematic. But in many cases the savings can be dominated by the time and space required to generate the abstractions, and thus motivates the development of cheaper but less accurate *approximate* clustering schemes.

Another way to reduce this overhead is to adopt a fixed abstraction scheme so that only one abstraction is ever produced. This approach has been adopted in classical planning in *hierarchical* or *abstraction-based* planners, pioneered by Sacerdoti's AbStrips system (Sacerdoti, 1974). A similar form of abstraction is studied by Knoblock (1993) (see also Knoblock, Tenenberg, & Yang, 1991). In this work, variables (in this case propositional) are ranked according to *criticality* (roughly, how important such variables are to the solution of the planning problem) and an abstraction is constructed by deleting from the problem description a set of propositions of low criticality. A solution to this abstract problem is a plan that achieves the elements of the original goal that have not been deleted. However, preconditions and effects of actions that have been deleted are not accounted for in this solution, so it might not be a solution to the original problem. Even so, the abstract solution can be used to restrict search for a solution in the underlying concrete space. Very often hierarchies of more and more refined abstractions are used and propositions are introduced back into the domain in stages.

This form of abstraction is uniform (propositions are deleted uniformly) and fixed. Since the abstract solution need not be a solution to the problem, we might be tempted to view it as an approximate abstraction method. However, it is best not to think of the abstract plan as a solution at all, rather as a form of heuristic information that can help solve the true problem more quickly.

The intuitions underlying Knoblock's scheme are applied to DTP by Boutilier and Dearden (1994, 1997): variables are ranked according to their degree of influence on the reward function and a subset of the most important variables is deemed relevant. Once this subset is determined, those variables that influence the relevant variables through the effects of actions (which can be determined easily using Strips or 2TBN action descriptions) are also deemed relevant, and so on. All remaining variables are deemed irrelevant and are deleted from the description of the problem (both action and reward descriptions). This leaves an abstract MDP with a smaller state space (i.e., fewer variables) that can be solved by standard methods. Recall that the state space reduction is exponential in the number of variables removed. We can view this method as an uniform fixed approximate abstraction scheme. Unlike the output of classical abstraction methods, the abstract policy produced can be implemented and has a value. The degree to which the optimal abstract policy and the true optimal policy differ in value can be bounded *a priori* once the abstraction is fixed.

**Example 5.3** As a simple illustration, suppose that the reward for satisfying coffee requests (or penalty for not satisfying them) is substantially greater than that for keeping the lab tidy or for delivering mail. Suppose that time pressure requires our agent to focus on a specific subset of objectives in order to produce a small abstract state space. In this case, of the four reward-laden variables in our problem (see Figure 24), only *CR* will be judged to be important. When the action descriptions are used to determine the variables that can (directly or indirectly) affect the probability of achieving *CR*, only *CR*, *RHC* and *Loc* will be deemed relevant, allowing $T$, $M$, and *RHM* to be ignored. The state space is thus reduced from size 400 to size 20. In addition, several of the action descriptions (e.g., *Tidy*) become trivial and can be deleted. □

The advantage of these abstractions is that they are easily computed and incur little overhead. The disadvantages are that the uniform nature of such abstractions is restrictive, and the relevant "reward variables" are determined before the policy is constructed and without knowledge of the agent's ability to control these variables. As a result, important variables—those that have a large impact on reward—but over which the agent has no control, may be taken into account, while less important variables that the agent can actually influence are ignored. However, a series of such abstractions can be used that take into account objectives of decreasing importance, and the *a posteriori* most valuable objectives can be dealt with once risk and controllability are taken into account (Boutilier et al., 1997). The policies generated at more abstract levels can also be used to "seed" value or policy iteration at less abstract levels, in certain cases reducing the time to convergence (Dearden & Boutilier, 1997). It has also been suggested (Dearden & Boutilier, 1994, 1997) that the abstract value function be used as a heuristic in an online search for policies that improve the abstract policy so constructed, as discussed in Section 3.2.2. Thus, the error in the approximate value function is overcome to some extent by search, and the heuristic function can be improved by asynchronous updates.

A different use of abstraction is adopted in the DRIPS planner (Haddawy & Suwandi, 1994; Haddawy & Doan, 1994). Actions can be abstracted by collapsing "branches," or possible outcomes, and maintaining probabilistic intervals over the abstract, disjunctive effects. Actions are also combined in an decomposition hierarchy, much like those in hierarchical task networks. Planning is done by evaluating abstract plans in the decomposition network, producing ranges of utility for the possible instantiations of those plans, and refining only those plans that are possibly optimal. The use of task networks means that search is restricted to finite-horizon, open-loop plans with action choice restricted to possible refinements of the network. Such task networks offer a useful way to encode *a priori* heuristic knowledge about the structure of good plans.

#### 5.1.4 Model Minimization and Reduction Methods

The abstraction techniques defined above can be recast in terms of minimizing a stochastic automaton, providing a unifying view of the different methods and offering new insights into the abstraction process (Dean & Givan, 1997). From automata theory we know that for any given finite-state machine $M$ recognizing a language $L$ there exists a unique minimal finite-state machine $M'$ that also recognizes $L$. It could be that $M = M'$, but it might also be that $M'$ is exponentially smaller than $M$. This minimal machine, called the *minimal model* for the language $L$, captures every relevant aspect of $M$ and so the machines are said to be equivalent. We can define similar notions of equivalence for MDPs. Since we are primarily concerned with planning, it is important that equivalent MDPs agree on the value functions for all policies. From a practical standpoint, it may not be necessary to find the minimal model if we can find a *reduced* model that is sufficiently small but still equivalent.

We apply the idea of model minimization (or model reduction) to planning as follows: we begin by using an algorithm that takes as input an implicit MDP model in factored form and produces (if we are lucky) an explicit, reduced model whose size is within a polynomial factor of the size of the factored representation. We then use our favorite state-based dynamic programming algorithms to solve the explicit model.

We can think of the dynamic programming techniques that rely on structured representations discussed earlier as operating on a reduced model without ever explicitly constructing that model. In some cases, building the reduced model once and for all may be appropriate; in other cases, one might save considerable effort by explicitly constructing only those parts of the reduced model that are absolutely necessary.

There are some potential computational problems with the model-minimization techniques sketched above. A small minimal model may exist, but it may be hard to find. Instead, we might look for a reduced model that is easier to find but not necessarily minimal. This too could fail, in which case we might look for a model small enough to be useful but only approximately equivalent to the original factored model. We have to be careful what we mean by "approximate," but intuitively two MDPs are approximately equivalent if the corresponding optimal value functions are within some small factor of one another.

In order to be practical, MDP model reduction schemes operate directly on the implicit or factored representation of the original MDP. Lee and Yannakakis (1992) call this *online model minimization*. Online model minimization starts with an initial partition of the states. Minimization then iteratively refines the partition by splitting clusters into smaller clusters. A cluster is split if and only if the states in the cluster behave differently with respect to transitions to states in the same or other clusters. If this local property is satisfied by all clusters in a given partition, then the model consisting of aggregate states that correspond to the clusters of this partition is equivalent to the original model. In addition, if the initial partition and the method of splitting clusters satisfy certain properties,[45] then we are guaranteed to find the minimal model. In the case of MDP reduction, the initial partition groups together states that have the same reward, or nearly the same reward in the case of approximation methods.

The clusters of the partitions manipulated by online model reduction methods are represented intensionally as formulas involving the state variables. For instance, the formula $RHC \wedge Loc(M)$ represents the set of all states such that the robot has coffee and is located in the mail room. The operations performed on these clusters require conjoining, complementing, simplifying, and checking for satisfiability. In the worst case, these operations are intractable, and so the successful application of these methods depends critically on the problem and the way in which it is represented. We illustrate the basic idea on a simple example.

**Example 5.4** Figure 30 depicts a simple version of our running example with a single action. There are three boolean state variables corresponding to $RHC$—the robot has coffee (or not, $\overline{RHC}$), $CR$—there is an outstanding request for coffee (or not, $\overline{CR}$), and, considering only two location possibilities, $Loc(C)$—the robot is in the coffee room (or not, $\overline{Loc(C)}$). Whether there is an outstanding coffee request depends on whether there was a request in the previous stage and whether the robot was in the coffee room. Location depends only on the location at the previous stage, and the reward depends only on whether or not there is an outstanding coffee request.

45. The property required of the initial partition is that, if two states are in the same cluster of the partition defining the minimal model (recall that the minimal model is unique), then they must be in same cluster in the initial partition.

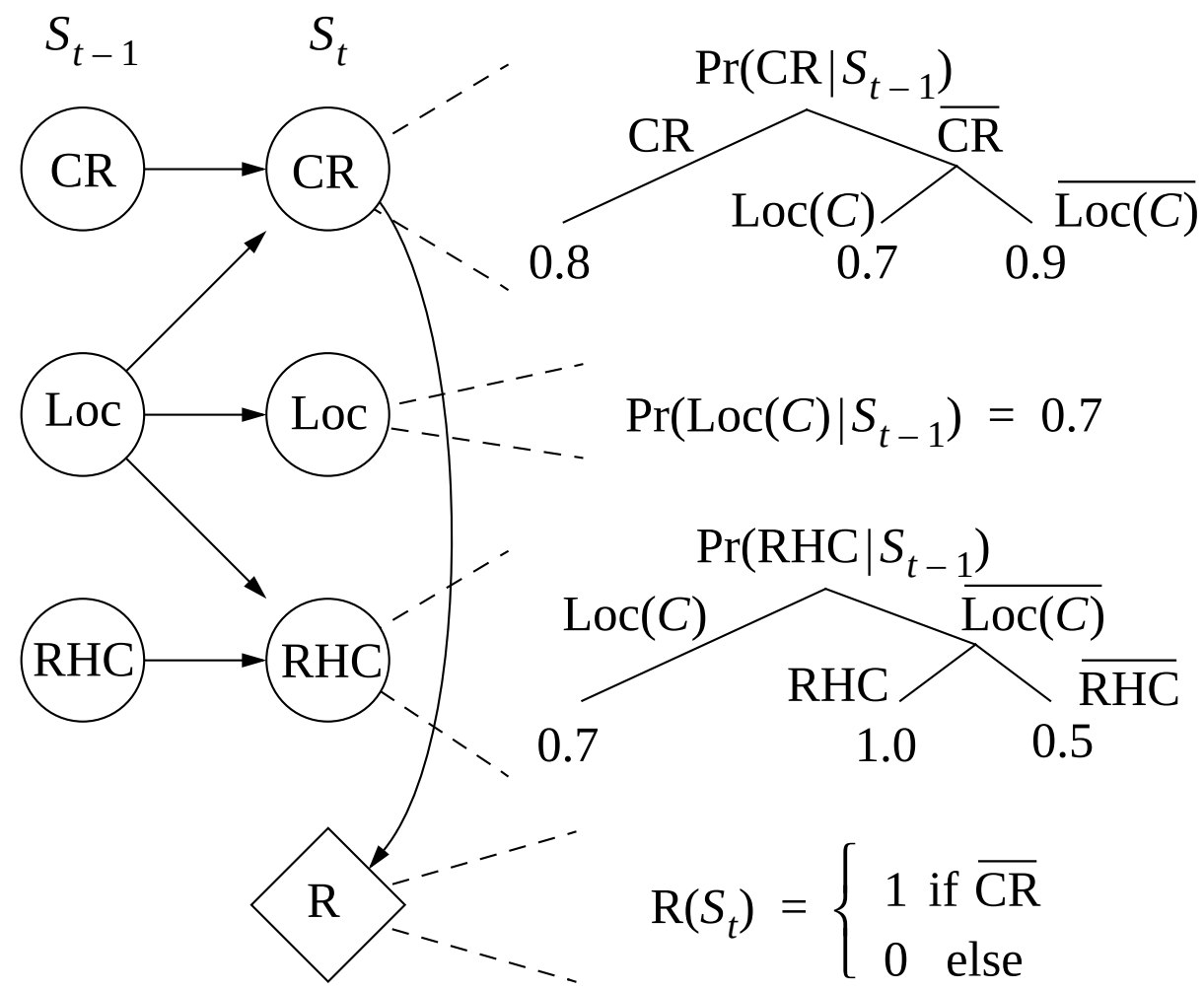

Figure 30: Factored model illustrating model-reduction techniques.

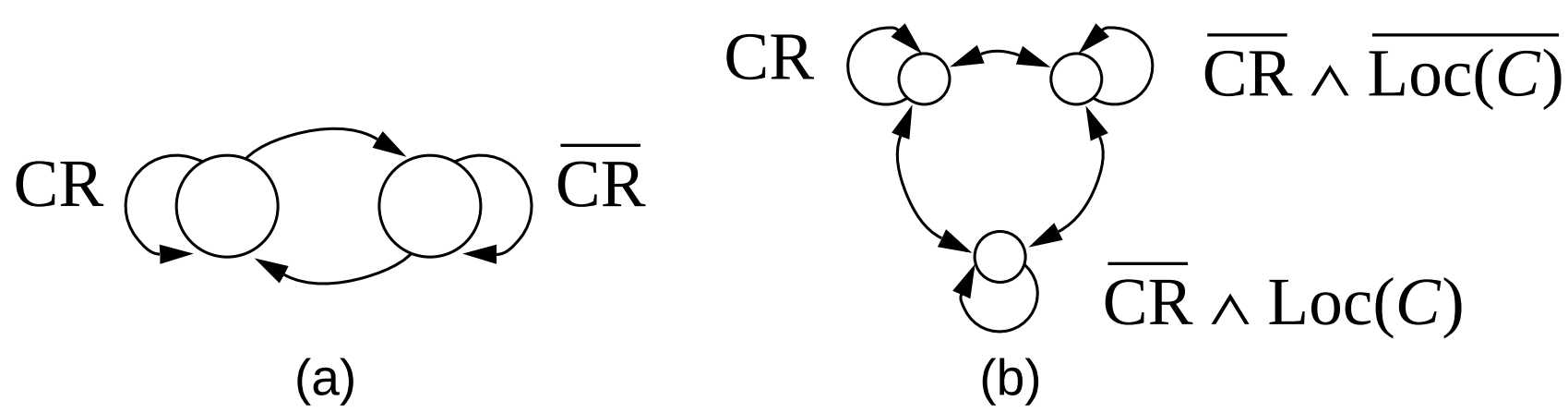

Figure 31: Models involving aggregate states: (a) the model corresponding to the initial partition and (b) the minimal model.

The initial partition shown in Figure 31(a) is defined in terms of immediate rewards. We say that all the states in a particular starting cluster behave the same with respect to a particular destination cluster if the probability of ending up in the destination cluster is the same for all states in the starting cluster. This property is not satisfied for starting cluster $\overline{CR}$ and destination cluster $CR$ in Figure 31(a), and so we split the cluster labeled $\overline{CR}$ to obtain the model in Figure 31(b). Now the property is satisfied for all pairs of clusters and the model in Figure 31(b) is the minimal model. □

The Lee and Yannakakis algorithm for non-deterministic finite-state machines has been extended by Givan and Dean to handle classical Strips planning problems (Givan & Dean, 1997) and MDPs (Dean & Givan, 1997). The basic step of splitting a cluster is closely related to goal regression, a relationship explored in (Givan & Dean, 1997). Variants of the model reduction approach apply when the action space is large and represented in a factored form (Dean, Givan, & Kim, 1998); for example, when each action is specified by a set of parameters such as those corresponding to the allocations of several different resources in an optimization problem. There also exist algorithms for computing approxi-mate models (Dean, Givan, & Leach, 1997) and efficient planning algorithms that use these approximate models (Givan, Leach, & Dean, 1997).

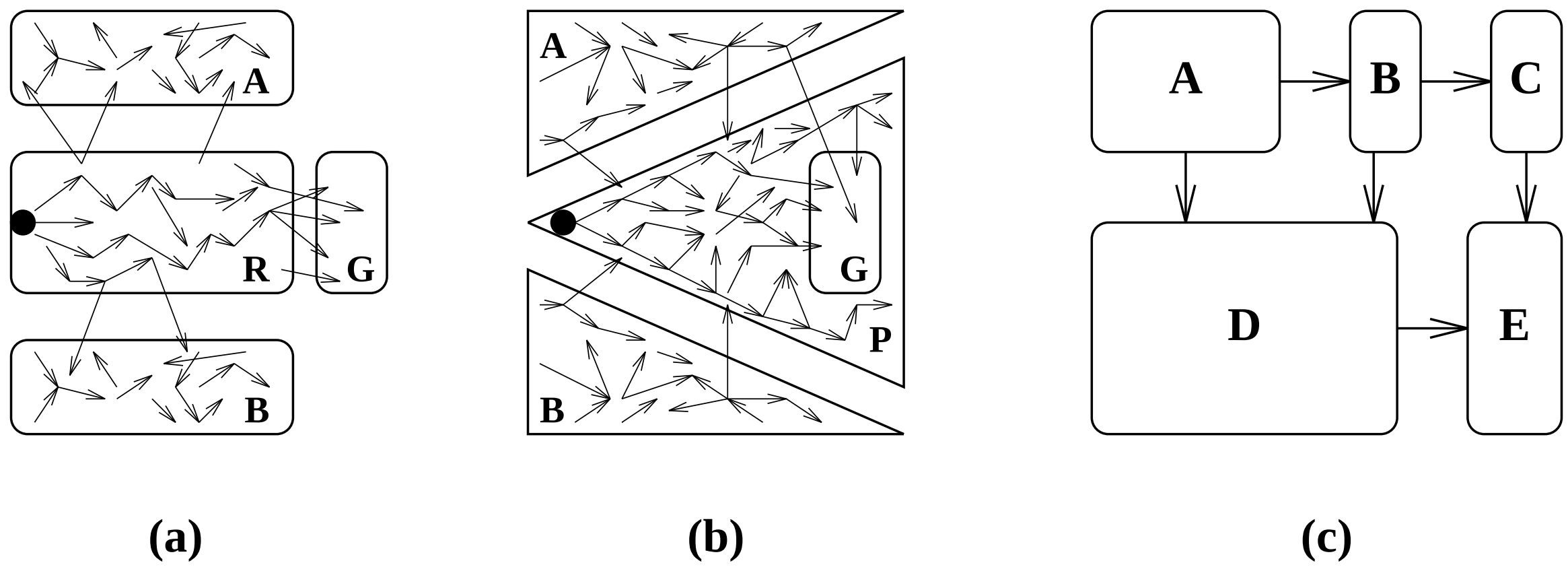

Figure 32: Reachability and serial problem decomposition.

## 5.2 Reachability Analysis and Serial Problem Decomposition

### 5.2.1 Reachability Analysis

The existence of goal states can be exploited in different settings. For instance, in deterministic classical planning problems, regression can be viewed as a form of directed dynamic programming. Without uncertainty, a certain policy either reaches a goal state or does not, and the dynamic programming backups need be performed only from goal states, not from all possible states. Regression, therefore, implicitly exploits certain *reachability* characteristics of the domain along with the special structure of the value function.

Reachability analysis applied much more broadly forms the basis for various types of *problem decomposition*. In decomposition problem solving, the MDP is broken into several subprocesses that can be solved independently, or roughly independently, and the solutions can be pieced together. If subprocesses whose solutions interact marginally are treated as independent, we might expect a good but nonoptimal global solution to result. Furthermore, if the structure of the problem requires that only a solution to a particular subproblem is needed, then the solutions to other subproblems can be ignored or need not be computed at all. For instance, in regression analysis, the optimal action for states that cannot reach a goal region is irrelevant to the solution of a classical AI planning problem. This is shown schematically in Figure 32(a), where regions $A$ and $B$ are never explored in the backward search through state space: only states that can reach the goal within the search horizon are ever deemed relevant. While regions $A$ and $B$ may be reachable from the start state, the fact that they do not reach the goal state means they are known to be irrelevant. Should the system dynamics be stochastic, such a scheme can form the basis of an approximately optimal solution method: regions $A$ and $B$ can be ignored if they are unlikely to transition to the regression of the goal region (region $R$). Similar remarks using progression or forward search from the start state apply, as illustrated in Figure 32(b).

Several schemes have been proposed in the AI literature for exploiting such reachability constraints, apart from the usual forward- or backward-search approaches. Peot and Smith (1993) introduce the *operator graph*, a structure computed prior to problem solving that caches reachability relationships among propositions. The graph can be consulted during the planning process in deciding which actions to insert into the plan and how to resolve threats.

The GraphPlan algorithm of Blum and Furst (1995) attempts to blend considerations of both forward and backward reachability in a deterministic planning context. One of the difficulties with regression is that we may regress the goal region through a sequence of operators only to find ourselves in a region that cannot be reached from the initial state. In Figure 32(a), for example, not all states in region $R$ may be reachable from the initial state. GraphPlan constructs a variant of the operator graph called the *planning graph*, in which certain forward reachability constraints are posted. Regression is then implemented as usual, but if the current subgoal set violates the forward reachability constraints at any point, this subgoal set is abandoned and the regression search backtracks.

Conceptually, one might think of GraphPlan as constructing a forward search tree through state space with the initial state as the root, then doing a backward search from the goal region backward through this tree. Of course, the process is not state-based: instead, constraints on the possible variable values that can hold simultaneously at different planning stages are recorded, and regression is used to search backward through the planning graph. In a sense, GraphPlan can be viewed as constructing an abstraction in which forward-reachable states are distinguished from unreachable states at each planning stage, and using this distinction among abstract states quickly to identify infeasible regression paths. Note, however, the GraphPlan *approximates* this distinction by overestimating the set of reachable states. Overestimation (as opposed to underestimation) ensures that the regression search space contains all legitimate plans.

Reachability has also been exploited in the solution of more general MDPs. Dean et al. (1995) propose an *envelope* method for solving "goal-based" MDPs approximately. Assuming *some* path can be generated quickly from a given start state to the goal region, an MDP consisting of the states on this path and perhaps neighboring states is solved. To deal with transitions that lead out of this *envelope*, a heuristic method estimates a value for these states.[46] As time permits, the set of neighboring states can be expanded, increasing solution quality by more accurately evaluating the quality of alternative actions.

Some of the ideas underlying GraphPlan have been applied to more general MDPs in (Boutilier, Brafman, & Geib, 1998), where the construction of a planning graph is generalized to deal with the stochastic, conditional action representation offered by 2TBNs. Given an initial state (or set of initial states), this algorithm discovers reachability constraints that have a form like those in GraphPlan — for instance, that two variable values $X = x_1$ and $Y = y_3$ cannot both obtain simultaneously; that is, no action sequence starting at the given initial state can lead to a state in which these values both hold.[47] The reachability constraints discovered by this process are then used to simplify the action and reward representation of an MDP so that it refers only to reachable states. In this case, any action that

46. The approximate abstraction techniques described in Section 5.1.3 might be used to generate such heuristic information.
47. General $k$-ary constraints of this type are considered in (Boutilier et al., 1998).

requires an unreachable set of values to hold is effectively deleted. In some cases, certain variables are discovered to be immutable given the initial conditions and can themselves be deleted, leading to much smaller MDPs. This simplified representation retains the original propositional structure so standard abstraction methods can be applied to the reachable MDP. It is also suggested that a strong synergy exists between abstraction and reachability analysis such that together these techniques reduce the size of the "effective" MDP to be solved much more dramatically than either does in isolation. Just as reachability constraints can be used to prune regression paths in deterministic domains, they can be used to prune value function and policy estimates generated by decision-theoretic regression and abstraction algorithms (Boutilier et al., 1998).

The results reported in (Boutilier et al., 1998) are limited to a single process-planning domain, but show that reachability analysis together with abstraction can provide substantial reductions in the size of the effective MDP that must be solved, at least in some domains. In a domain with 31 binary variables, reachability considerations generally eliminated on the order of 10 to 15 variables (depending on the initial state and the arity—binary or ternary—of the constraints considered), reducing the state space from size $2^{31}$ to anywhere from $2^{22}$ to $2^{15}$. Incorporating abstraction on the reachable MDP provided considerably more reduction, reducing the MDP to sizes ranging from $2^8$ to effectively zero states. The latter case would occur if it is discovered that no values of variables that impact reward can be altered—in which case every course of action has the same expected utility and the MDP needn't be solved (or can be solved by applying null actions with zero cost).

#### 5.2.2 Serial Problem Decomposition and Communicating Structure

The *communicating* or reachability structure of an MDP provides a way to formalize different types of problem decomposition. We can classify an MDP according to the Markov chains induced by the stationary policies it admits. For a fixed Markov chain, we can group states into maximal recurrent classes and transient states, as described in Section 2.1. An MDP is *recurrent* if each policy induces a Markov chain with a single recurrent class. An MDP is *unichain* if each policy induces a single recurrent class with (possibly) some transient states. An MDP is *communicating* if for any pair of states $s, t$, there is *some* policy under which $s$ can reach $t$. An MDP is *weakly communicating* if there exists a closed set of states that is communicating plus (possibly) a set of states transient under *every* policy. We call other MDPs *noncommunicating.*

These notions are crucial in the construction of optimal average-reward policies, but can also be exploited in problem decomposition. Suppose an MDP is discovered to consist of a set of recurrent classes $\mathcal{C}_1, \cdots \mathcal{C}_n$ (i.e., no matter what policy is adopted, the agent cannot leave any such class once it enters that class) and a set of transient states.[48] It is clear that optimal policy restricted to any class $\mathcal{C}_i$ can be constructed without reference to the policy decisions made at any states outside of $\mathcal{C}_i$ or even their values. Essentially, each $\mathcal{C}_i$ can be viewed as an independent subprocess.

48. A simple way to view these classes is to think of the agent adopting a randomized policy where each action is adopted at any state with positive probability. The classes of the induced Markov chain correspond to the classes of the MDP.

This observation leads to the following suggestion for optimal policy construction:[49] we solve the subprocesses consisting of the recurrent classes for the MDPs; we then remove these states from the MDP, forming a reduced MDP consisting only of the transient states. We then break the reduced MDP into its recurrent classes and solve these independently. The key to doing this effectively is to use the value function for the original recurrent states (computed in solving the independent subproblems in Step 1) to take into account transitions out of the recurrent classes in the reduced MDP. Figure 32(c) shows an MDP broken into the classes that might be constructed this way. In the original MDP, classes $C$ and $E$ are recurrent and can be solved independently. Once removed from the MDP, class $D$ is recurrent in the reduced MDP. It can, of course, be solved without reference to classes $A$ and $B$, but does rely on the value of the states that it transitions to in class $E$. However, the value function for $E$ is available for this purpose, and can be used to solve for $D$ as if $D$ consisted only of $|D|$ states. With this in hand, $B$ can then be solved, and finally $A$ can be solved. Lin and Dean (1995) provide a version of this type of decomposition that also employs a factored representation. The factored representation allows dimensionality reduction in different state subspaces by aggregating states that differ only in the values of the irrelevant variables in their subspaces.

A key to such a decomposition is the discovery of the recurrent classes of an MDP. Puterman (1994) suggests an adaptation of the Fox-Landi algorithm (Fox & Landi, 1968) for discovering the structure of Markov chains that is $O(N^2)$ (recall $N = |\mathcal{S}|$).[50] To alleviate the difficulties of algorithms that work with an explicit state-based representation, Boutilier and Puterman (1995) propose a variant of the algorithm that works with a factored 2TBN representation.

One difficulty with this form of decomposition is its reliance on strongly independent subproblems (i.e., recurrent classes) within the MDP. Others have explored exact and approximate techniques that work under less restrictive assumptions. One simple method of approximation is to construct "approximately recurrent classes." In Figure 32(c) we might imagine that $C$ and $E$ are nearly independent in the sense that all transitions between them are very low-probability or high-cost. Treating them as independent might lead to approximately optimal policies whose error can be bounded. If the solutions to $C$ and $E$ interact strongly enough that the solutions should not be constructed completely independently, a different approach to solving the decomposed problem can be taken.

If we have the optimal value function for $E$ then, as pointed out, we can calculate the optimal value function for $D$. The first thing to note is that we don't need to know the value function for all of the states in $E$, just the value of every state in $E$ that is reachable from some state in $D$ in a single step. The set of all states outside $D$ reachable in a single step from a state inside $D$ is referred to as the *states in the periphery of* $D$. The values of the states in the intersection of $E$ and the periphery of $D$ summarize the value of exiting $D$ and ending up in $E$. We refer to the set of all states that are in the periphery of some block as the *kernel* of the MDP. All of the different blocks interact with one another through states in the kernel.

49. Ross and Varadarajan (1991) make a related suggestion for solving average-reward problems.
50. A slight correction is made to the suggested algorithm in (Boutilier & Puterman, 1995).

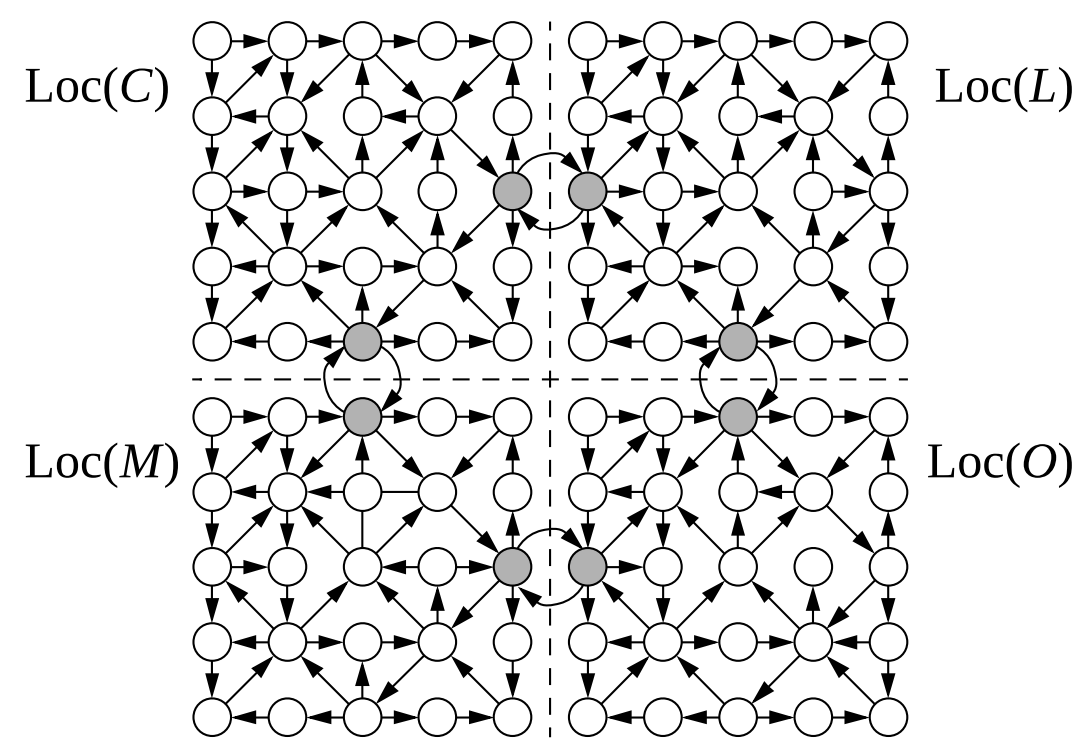

Figure 33: Decomposition based on location.

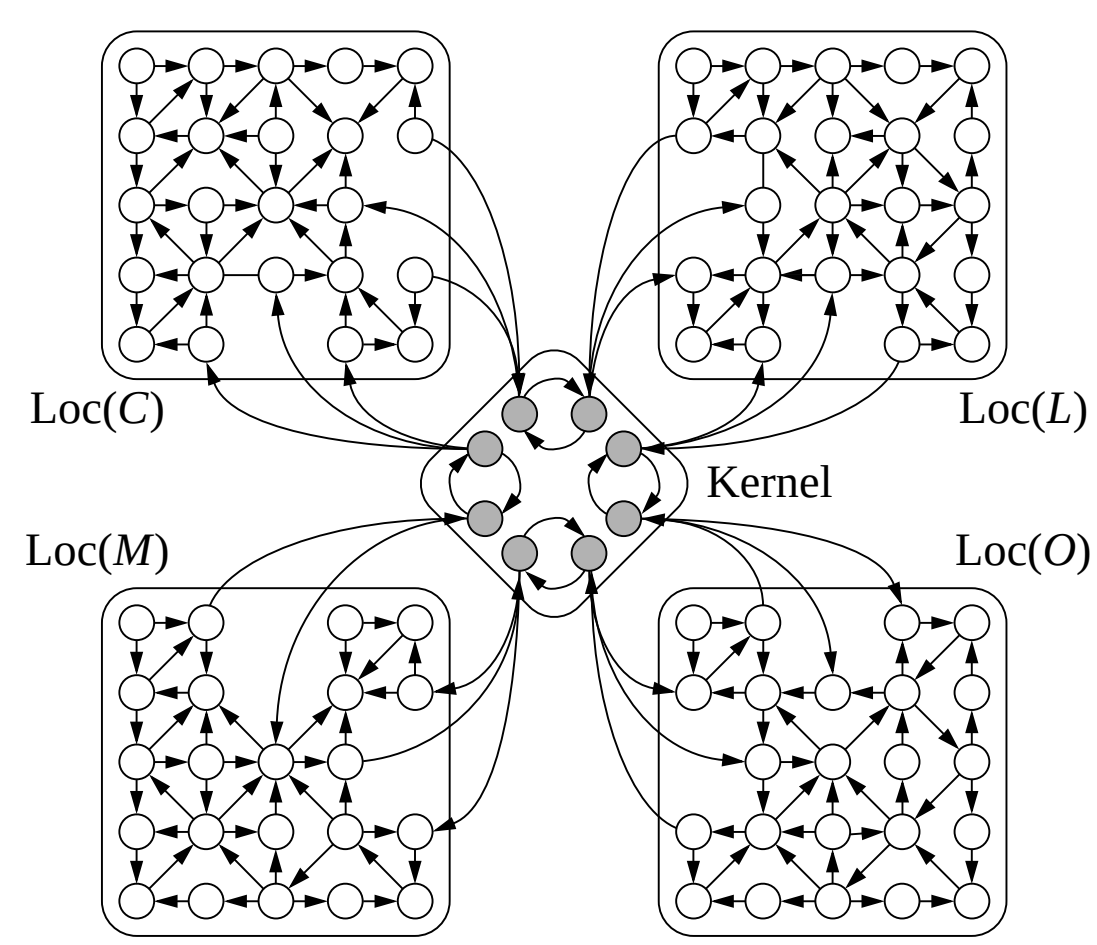

Figure 34: Kernel-based decomposition depicting the kernel states.

**Example 5.5** Spatial features often provide a natural dimension along which to decompose a domain. In our running example, the location of the robot might be used to decompose the state space into blocks of states, one block for each of the possible locations. Figure 33 shows such a decomposition superimposed over the state-transition diagram for the MDP. States in the kernel are shaded and might correspond to the entrances and exits of locations. The star-shaped topology, induced by the kernel decomposition used in (Kushner & Chen, 1974) and (Dean & Lin, 1995), is illustrated in Figure 34. In Figure 33, the hallway location is not explicitly represented. This simplification may be reasonable if the hallway is only a conduit for moving from one room to another; in this case the function of the hallway is accounted for in the dynamics governing states in the kernel. Figures 33 and 34 are idealized in that, given the full set of features in our running example, the kernel would contain many more states. □

One technique for computing the optimal policy for the entire MDP involves repeatedly solving the MDPs corresponding to the individual blocks. The techniques works as follows: initially, we guess the value of every state in the kernel.[51] Given a current estimate for the values of the kernel states, we solve the component MDPs; this solution produces a new estimate for the states in the kernel. We adjust the values of the states in the kernel by considering the difference between the current and the new estimates and iterate until this difference is negligible.

This iterative method for solving a decomposed MDP is a special case of the Lagrangian method for finding the extrema of a function. The OR literature is replete with such methods for both linear and nonlinear systems of equations (Winston, 1992). It is possible to formulate an MDP as a linear program (D'Epenoux, 1963; Puterman, 1994). Dantzig and Wolfe (1960) developed a method of decomposing a system of equations involving a very large number of variables into a set of smaller systems of equations interacting through a set of coupling variables (variables shared by more two or more blocks). In the Dantzig-Wolfe decomposition method, the original, very large system of equations is solved by iteratively solving the smaller systems and adjusting the coupling variables on each iteration until no further adjustment is required. In the linear programming formulation of an MDP, the values of the states are encoded as variables.

Kushner and Chen (1974) exploit the fact that MDPs can be modeled as linear programs by using the Dantzig-Wolfe decomposition method to solve MDPs involving a large number of states. Dean and Lin (1995) describe a general framework for solving decomposed MDPs pointing to the work of Kushner and Chen as a special case, but neither work addresses the issue of where the decompositions come from. Dean et al. (1995) investigate methods for decomposing the state space into two blocks: those reachable in $k$ steps or fewer and those not reachable in $k$ steps (see the discussion of reachability above). The set of states reachable in $k$ or fewer steps is used to construct an MDP that is the basis for a policy that approximates the optimal policy. As $k$ increases, the size of the block of states reachable in $k$ steps increases, ensuring a better solution; but the amount of time required to compute a

51. Ideally we would aggregate kernel states with the same value so as to provide a compact representation. In the remainder of this section, however, we won't consider this or any other opportunities for combining aggregation and decomposition methods.

solution also increases. Dean et al. (1995) discuss methods for solving MDPs in time-critical problems by trading off quality against time.

We have ignored the issue of how to obtain decompositions that expedite our calculations. Ideally, each component of the decomposition would yield to simplification via aggregation and abstraction, reducing the dimensionality in each component and thereby avoiding explicit enumeration of all the states. Lin (1997) presents methods for exploiting structure for certain special cases in which the communicating structure is revealed by a domain expert. In general, however, finding a decomposition so as to minimize the effort spent in solving the component MDPs is quite hard (at least as hard as finding the smallest circuit consistent with a given input-output behavior) and so the best we can hope for are good heuristic methods. Unfortunately, we are not aware of any particularly useful heuristics for finding serial decompositions for Markov decision processes. Developing such heuristics is clearly an area for investigation.

Related to this form of decomposition is the development of *macro operators* for MDPs (Sutton, 1995). Macros have a long history in classical planning and problem solving (Fikes, Hart, & Nilsson, 1972; Korf, 1985), but only recently have they been generalized to MDPs (Hauskrecht, Meuleau, Kaelbling, Dean, & Boutilier, 1998; Parr, 1998; Parr & Russell, 1998; Precup, Sutton, & Singh, 1998; Stone & Veloso, 1999; Sutton, 1995; Thrun & Schwartz, 1995). In most of this work, a macro is taken to be a local policy over a region of state space (or block in the above terminology). Given an MDP comprising these blocks and a set of macros defined for each block, the MDP can be solved by selecting a macro action for each block such that the global policy induced by the set of macros so picked is close to optimal, or at the very least is the best combination of macros from the set available. In (Sutton, 1995; Precup et al., 1998), macros are treated as temporally-abstract actions and models are defined by which a macro can be treated as if it were a single action and used in policy or value iteration (along with concrete actions). In (Hauskrecht et al., 1998; Parr, 1998; Parr & Russell, 1998), these models are exploited in a hierarchical fashion, with a high-level MDP consisting only of states lying on the boundaries of blocks, and macros the only "actions" that can be chosen at these states. The issue of macro generation—constructing a set of macros guaranteed to provide the flexibility to select close to optimal global behavior—is addressed in (Hauskrecht et al., 1998; Parr, 1998). The relationship to serial decomposition techniques is quite close; thus, the problems of discovering good decompositions, constructing good sets of macros, and exploiting intensional representations are areas in which clearer, compelling solutions are required. To date, work in this area has not provided much computational utility in the solution of MDPs—except in cases where good, hand-crafted, region-based decompositions and macros can be provided—and little of this work has taken into account the factored nature of many MDPs. For this reason, we do not discuss it in detail. However, the general notion of serial decomposition continues to develop and shows great promise.

### 5.3 Multiattribute Reward and Parallel Decomposition

Another form of decomposition is *parallel decomposition*, in which an MDP is broken into a set of sub-MDPs that are "run in parallel." Specifically, at each stage of the (global) decision process, the state of each subprocess in affected. For instance, in Figure 35, action $a$ affects the state of each subprocess. Intuitively, an action is suitable for execution in the original MDP at some state if it is reasonably good in each of the sub-MDPs.

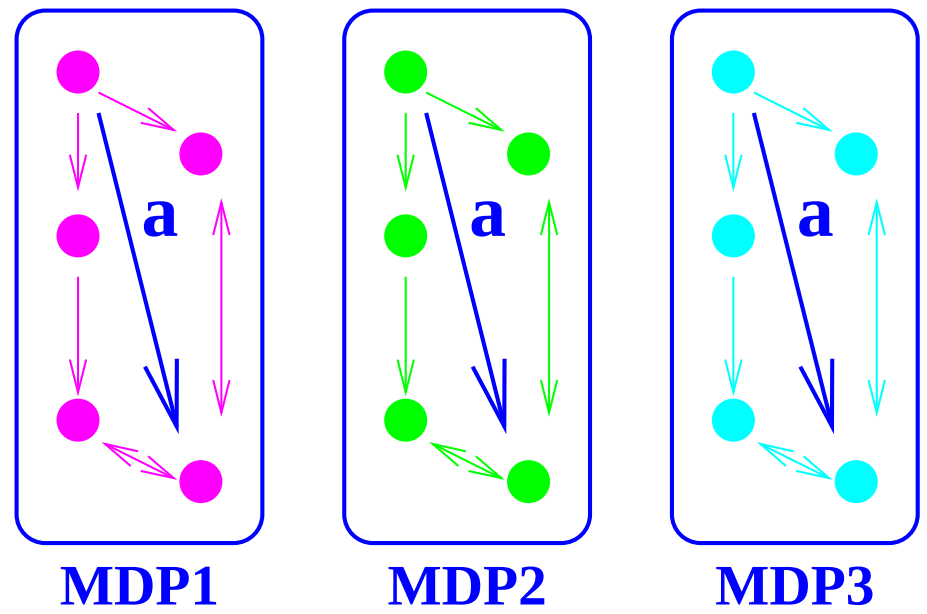

Figure 35: Parallel problem decomposition.

Generally, the sub-MDPs form either a product or join decomposition of the original state space (contrast this with the union decompositions of state space determined by serial decompositions): the state space is formed by taking the cross product of the sub-MDP state spaces, or the join if certain states in the subprocesses cannot be linked. The subprocesses may have identical action spaces (as in Figure 35), or each may have its own action space, with the global action choice being factored into a choice for each subprocess. In the latter case, the sub-MDPs may be completely independent, in which case the (global) MDP can be solved exponentially faster. A more challenging problem arises when there are constraints on the legal action combinations. For example, if the actions in the subprocesses each require certain shared resources, interactions in the global choice may arise.

In a parallel MDP decomposition, we wish to solve the sub-MDPs and use the policies or value functions generated to help construct an optimal or approximately optimal solution to the original MDP, highlighting the need to find appropriate decompositions for MDPs and to develop suitable merging techniques. Recent parallel decomposition methods have all involved decomposing an MDP into subprocesses suitable for distinct objectives. Since reward functions often deal with multiple objectives, each associated with an independent reward, and whose rewards can be summed to determine a global reward, this is often a very natural way to decompose MDPs. Thus, ideas from multiattribute utility theory can be seen to play a role in the solution of MDPs.

Boutilier et al. (1997) decompose an MDP specified using 2TBNs and an additive reward function using the abstraction technique described in Section 5.1.3. For each component of the reward function, abstraction is used to generate an MDP referring only to variables relevant to that component.[52] Since certain state variables may be present in multiple sub-MDPs (i.e., relevant to more than one objective), the original state space in the join of the subspaces. Thus, decomposition is tackled automatically. Merging is tackled in several ways. One involves using the sum of the value functions obtained by solving the sub-MDPs as a heuristic estimate of the true value function. This heuristic is then used to guide online, state-based search (see Section 3.2.1). If the sub-MDPs do not interact, then this heuristic is perfect and leads to backtrack-free optimal action selection; if they interact, search is

52. Note that the existence of a factored MDP representation is crucial for this abstraction method.

required to detect conflicts. Note that each sub-MDP has identical sets of actions. If the action space is large, the branching factor of the search process may be prohibitive.

Singh and Cohn (1998) also deal with parallel decomposition, though they assume the global MDP is specified explicitly as a set of parallel MDPs, thus generating decompositions of a global MDP is not at issue. The global MDP is given by the cross product of the state and action spaces of these sub-MDPs and the reward functions are summed. However, constraints on the feasible action combinations couple the solutions of these sub-MDPs. To solve the global MDP, the sum of the sub-MDP value functions is used as an upper bound on the optimal global value function, while the maximum of these (at any global state) is used as a lower bound. These bounds then form the basis of an action-elimination procedure in a value-iteration algorithm for solving the global MDP.[53] Unfortunately, value iteration is run over the explicit state space of the global MDP. Since the action space is also a cross product, this is a potential computational bottleneck for value iteration, as well.

Meuleau et al. (1998) use parallel decomposition to approximate the solution of stochastic resource allocation problems with very large state and action spaces. Much like Singh and Cohn (1998), an MDP is specified in terms of a number of independent MDPs, each involving a distinct objective, whose action choices are linked through shared resource constraints. The value functions for the individual MDPs are constructed offline and then used in set of online action-selection procedures. Unlike many of the approximation procedures we have discussed, this approach makes no attempt to construct a policy explicitly (and is similar to real-time search or RTDP in this respect) nor to construct the value function explicitly. This method has been applied to very large MDPs, with state spaces of size $2^{1000}$ and actions spaces that are even larger, and can solve such problems in roughly half an hour. The solutions produced are approximate, but the size of the problem precludes exact solution; so good estimates of solution quality are hard to derive. However, when the same method is applied to smaller problems of the same nature whose exact solution can be computed, the approximations have very high quality (Meuleau et al., 1998). While able to solve very large MDPs (with large, but factored, state and action spaces), the model relies on somewhat restrictive assumptions about the nature of the local value functions that ensure good solution quality. However, the basic approach appears to be generalizable, and offers great promise for solving very large factored MDPs.

The algorithms in both (Singh & Cohn, 1998) and (Meuleau et al., 1998) can be seen to rely at least implicitly on structured MDP representations involving almost independent subprocesses. It seems likely that such approaches could take further advantage of automatic MDP decomposition algorithms such as that of (Boutilier et al., 1997), where factored representations explicitly play a part.

### 5.4 Summary

We have seen a number of ways in which intensional representations can be exploited to solve MDPs effectively without enumeration of the state space. These include techniques for abstraction of MDPs, including those based on relevance analysis, goal regression and decision-theoretic regression; techniques relying on reachability analysis and serial decomposition; and methods for parallel MDP decomposition exploiting the multiattribute nature

53. Singh and Cohn (1998) also incorporate methods for removing unreachable states during value iteration.

of reward functions. Many of these methods can, in fortunate circumstances, offer exponential reduction is solution time and space required to represent a policy and value function; but none come with guarantees of such reductions except in certain special cases. While most of the methods described provide approximate solutions (often with error bounds provided), some of them offer optimality guarantees in general, and most can provide optimal solutions under suitable assumptions.

One avenue that has not been explored in detail is the relationship between the structured solution methods developed for MDPs described above and techniques used for solving Bayesian networks. Since many of the algorithms discussed in this section rely on the structure inherent in the 2TBN representation of the MDP, it is natural to ask whether they embody some of the intuitions that underlie solution algorithms for Bayes nets, and thus whether the solution techniques for Bayes nets can be (directly or indirectly) applied to MDPs in ways that give rise to algorithms similar to those discussed here. This remains an open question at this point, but undoubtedly some strong ties exist. Tatman and Shachter (1990) have explored the connections between influence diagrams and MDPs. Kjaerulff (1992) has investigated computational considerations involved in applying join tree methods for reasoning tasks such as monitoring and prediction in temporal Bayes nets. The abstraction methods discussed in Section 5.1.2 can be interpreted as a form of variable elimination (Dechter, 1996; Zhang & Poole, 1996). Elimination of variables occurs in temporal order, but good orderings within a time slice must also exploit the tree or graph structure of the CPTs. Approximation schemes based on variable elimination (Dechter, 1997; Poole, 1998) may also be related to certain of the approximation methods developed for MDPs. The independence-based decompositions of MDPs discussed in Section 5.3 can clearly be viewed as exploiting the independence relations made explicit by "unrolling" a 2TBN. The development of these and other connections to Bayes net inference algorithms will no doubt prove very useful in enhancing our understanding of existing methods, increasing their range of applicability and pointing to new algorithms.

## 6. Concluding Remarks

The search for effective algorithms for controlling automated agents has a long and important history, and the problem will only continue to grow in importance as more decision-making functionality is automated. Work in several disciplines, among them AI, decision analysis, and OR, has addressed the problem, but each has carried with it different problem definitions, different sets of simplifying assumptions, different viewpoints, and hence different representations and algorithms for problem solving. More often than not, the assumptions seem to have been made for historical reasons or reasons of convenience, and it is often difficult to separate the essential assumptions from the accidental. It is important to clarify the relationships among problem definitions, crucial assumptions, and solution techniques, because only then can a meaningful synthesis take place.

In this paper we analyzed various approaches to a particular class of sequential decision problems that have been studied in the OR, decision analysis, and AI literature. We started with a general, reasonably neutral statement of the problem, couched, for convenience, in the language of Markov decision processes. From there we demonstrated how various disciplines define the problem (i.e., what assumptions they make), and the effect of these assumptions on the worst-case time complexity of solving the problem so defined. Assumptions regarding two main factors seem to distinguish the most commonly studied classes of decision problems:

- observation or sensing: does sensing tend to be fast, cheap, and accurate or laborious, costly and noisy?
- the incentive structure for the agent: is its behavior evaluated on its ability to perform a particular task, or on its ability to control a system over an interval of time?

Moving beyond the worst-case analysis, it is generally assumed that, although pathological cases are inevitably difficult, the agent should be able to solve "typical" or "easy" cases effectively. To do so, the agent needs to be able to identify structure in the problem and to exploit that structure algorithmically.

We identified three ways in which structural regularities can be recognized, represented, and exploited computationally. The first is structure induced by domain-level simplifying assumptions like full observability, goal satisfaction or time-separable value functions, and so on. The second is structure exploited by compact domain-specific encodings of states, actions, and rewards. The designer can use these techniques to make structure explicit, and decision-making algorithms can then exploit the structural regularities as they apply to the particular problem at hand. The third involves aggregation, abstraction and decomposition techniques, whereby structural regularities can be discovered and exploited during the problem-solving process itself. In developing this framework—one that allows comparison of domains, assumptions, problems, and techniques drawn from different disciplines—we discover the essential problem structure required for specific representations and algorithms to prove effective; and we do so in such a way that the insights and techniques developed for certain problems, or within certain disciplines, can be evaluated and potentially applied to new problems, or within other disciplines.

A main focus of this work has been the elucidation of various forms of structure in decision problems and of how each can be exploited representationally or computationally. For the most part, we have focused on propositional structure, which is most commonly associated with planning in AI circles. A more complete treatment would also have included other compact representations of dynamics, rewards, policies, and value functions often considered in continuous, real-valued domains. For instance, we have not discussed linear dynamics and quadratic cost functions, often used in control theory (Caines, 1988), or the use of neural-network representations of value functions, as frequently adopted within the reinforcement learning community (Bertsekas & Tsitsiklis, 1996; Tesauro, 1994),[54] nor have we discussed the partitioning of continuous state spaces often addressed in reinforcement learning (Moore & Atkeson, 1995). Neither have we addressed the relational or quantificational structure used in first-order planning representations. However, even these techniques can be cast within the framework described here; for example, the use of piecewise-linear value functions can be seen as a form of abstraction in which different linear components are applied to different regions or clusters of state space.

54. Bertsekas and Tsitsiklis (1996) provide an in-depth treatment of neural network and linear function approximators for MDPs and reinforcement learning.

Although in certain cases we have indicated how to devise methods that exploit several types of structure at once, research along these lines has been limited. To some extent, many of the representations and algorithms described in this paper are complementary and should pose few obstacles to combination. It remains to be seen how they interact with techniques developed for other forms of structure, such as those used for continuous state and action spaces.

So our analysis raises opportunities and challenges: by understanding the assumptions, the techniques, and their relationships, a designer of decision-making agents has many more tools with which to build effective problem solvers; and the challenges lie in the development of additional tools and the integration of existing ones.

## Acknowledgments

Many thanks to the careful comments of the referees. Thanks to Ron Parr and Robert St-Aubin for their comments on an earlier draft of this paper. The students taking CS3710 (Spring 1999) taught by Martha Pollack at the University of Pittsburgh and CPSC522 (Winter 1999) at the University of British Columbia also deserve thanks for their detailed comments.

Boutilier was supported by NSERC Research Grant OGP0121843, and the NCE IRIS-II program Project IC-7. Dean was supported in part by a National Science Foundation Presidential Young Investigator Award IRI-8957601 and by the Air Force and the Advanced Research Projects Agency of the Department of Defense under Contract No. F30602-91-C-0041. Hanks was supported in part by ARPA / Rome Labs Grant F30602–95–1–0024 and in part by NSF grant IRI–9523649.